\documentclass[11pt,onecolumn]{IEEEtran}
\usepackage{afterpage, amsfonts, amsmath, amssymb, amsxtra,tabularx, arydshln, boxedminipage, changepage,  comment, diagbox, epsfig, epstopdf, float, graphics, graphicx, lastpage, lscape, mathtools, mdwmath,nicefrac, multirow, pifont, psfig, url, xspace, gensymb} 

\usepackage{lineno}
\usepackage{setspace} 

\usepackage{xcolor}
\usepackage{subcaption}

\newcommand{\vb}{{\bf b}}
\newcommand{\vc}{{\bf c}}
\newcommand{\vd}{{\bf d}}
\newcommand{\vm}{{\bf m}}
\newcommand{\vn}{{\bf n}}
\newcommand{\vp}{{\bf p}}
\newcommand{\vr}{{\bf r}}
\newcommand{\vt}{{\bf t}}

\newcommand{\ma}{{\bf A}}
\newcommand{\mc}{{\bf C}}

\newcommand{\mmp}{{\bf P}}
\newcommand{\mpi}{{\bf \Pi}}
\newcommand{\mr}{{\bf R}}
\newcommand{\mv}{{\bf V}}
\newcommand{\mw}{{\bf W}}
\newcommand{\hvn}{\hat{\vn}}
\newcommand{\hvt}{\hat{\vt}}

\newcommand{\zr}{{\bf 0}}
\newcommand{\TI}{{\widetilde{I}}}
\newcommand{\TS}{{\widetilde{S}}}

\newcommand{\om}{{\omega}}
\newcommand{\tet}{{\theta}}

\newcommand{\dtet}{{\delta\theta}}
\newcommand{\dR}{{\delta\Re}}

\newcommand\abs[1]{|#1|}

\newcommand{\tb}{\widetilde{b}}
\newcommand{\tc}{\widetilde{c}}
\newcommand{\tB}{\widetilde{B}}
\newcommand{\tn}{\widetilde{N}}
\newcommand{\tS}{\widetilde{S}}
\newcommand{\tx}{\widetilde{x}}
\newcommand{\ty}{\widetilde{y}}
\newcommand{\tX}{\widetilde{X}}
\newcommand{\tY}{\widetilde{Y}}
\newcommand{\tZ}{\widetilde{Z}}

\begin{document}
\doublespacing

\begin{titlepage}
\title{Object Modeling from Underwater Forward-Scan
Sonar Imagery with Sea-Surface Multipath}
\author{Yuhan Liu and Shahriar Negaharipour\\
Electrical and Computer Engineering Department\\
University of Miami\\
Coral Gables, FL 33146\\
{\tt\small nshahriar@miami.edu}}
\end{titlepage}
\maketitle


\begin{abstract}  
\noindent 
We propose an optimization technique for 3-D underwater object modeling from 2-D forward-scan sonar images at known poses. A key contribution, for objects imaged in the proximity of the sea surface, is to resolve the multipath artifacts due to the air-water interface. Here, the object image formed by the direct target backscatter is almost always corrupted by the ghost and sometimes by the mirror components (generated by the multipath propagation). Assuming a planar air-water interface, we model, localize, and discard the corrupted object region within each view, thus avoiding the distortion of recovered 3-D shape. Additionally, complementary visual cues from the boundary of the mirror component, distinct at suitable sonar poses, are employed to enhance the 3-D modeling accuracy. 

The optimization is implemented as iterative shape adjustment by displacing the vertices of triangular patches in the 3-D surface mesh model, in order to minimize the discrepancy between the data and synthesized views of the 3-D object model. To this end, we first determine 2-D motion fields that align the object regions in the data and synthesized views, then calculate the 3-D motion of triangular patch centers, and finally the model vertices. The 3-D model is initialized with the solution of an earlier space carving method applied to the same data. The same parameters are applied in various experiments with 2 real data sets, mixed real-synthetic data set, and computer-generated data guided by general findings from a real experiment, to explore the impact of non-flat air-water interface. The results confirm the generation of a refined 3-D model in about half-dozen iterations. 

\end{abstract}


\section{Introduction}
Poor visibility in turbid waters severely degrades the performance in automated underwater scene 
analysis and interpretation from 2-D optical images. Sonar is the preferred imaging modality because acoustic 
signals can penetrate through silt, mud and other similar sources of turbidity. This has led to growing efforts in 
developing techniques that utilize the images from high-frequency (MHz) forward-scan sonar (FSS) systems
\cite{DIDSON}, \cite{oculus}, \cite{blueview}, \cite{gemeni} and the likes.

The 3-D scene to 2-D image projection leads to loss of information about one coordinate of
points on the scene surfaces: depth $Z$ in the Cartesian coordinates $(X,\ Y,\ Z)$  and elevation 
angle $\phi$ in the spherical coordinates $(\Re,\tet,\phi)$ in optical and FSS systems, respectively. The reconstruction 
of the geometric properties of the scene requires the estimation of this unknown coordinate. For either 
modality, we can exploit the visual cues in motion sequences [3], [4], [5], [6] or stereo pairs [7], [8], namely, 
image feature motions or disparities, respectively. For FFS images, among serious drawbacks are the sparsity 
of reliable features and complexities posed by outliers due to high levels of speckle noise.

Alternatively, we may employ the backscatter cues within object region in a single image in accordance to the 
shape-from-shading (SfS) paradigm [9]; first introduced in Horn’s work for optical images [10], and applied 
to side-scan sonar (SSS) images [11], [12]. The key bottleneck in applying to FSS imagery is the many-to-one projection 
ambiguity: echos producing each pixel's intensity value are received from an unknown number of surface patches within the 
elevation arc of each sonar beam/azimuth direction. The ambiguity can be overcome in restrictive cases, 
primarily for scenes with monotonically varying surface-to-sonar distances, and visible cast shadows on a 
(relatively) flat background \cite{aykin2013forward}.

It has been shown that the FSS video recorded by an AUV navigating over the seabed encodes the missing elevation information for objects resting on the bottom [14], [15]. In [14], the sector-scan sonar images of an object encircled by the sonar platform are segmented into echo, shadow and background regions. The reconstruction involves merging the 
2-D elevation and reflection maps, estimated from the shadow and highlight (echo) regions, respectively. In [15],
the distinct boundaries between shadowed (dark) and illuminated (brighter) image regions during a fly-over trajectory, due to    
the relatively-narrow vertical beam-width of the FSS sonar. 
The boundary points are at the same range over all sonar beams in the absence of any protruding object on the flat background. In contrast, highlight due to backscatter from a protruding object extends across the
boundaries into near-range shadow region due to the shorter round-trip travel distance of reflected
acoustic beams. The extent of penetration yields the missing elevation information. These approaches
have limitations for natural objects with complex shape details. 

 {The space carving (SC) method in [2] starts with the discretized 3-D space (voxels) visible in all known sonar poses  (where data is captured). Based on the binary object/background classification, a voxel is assigned the “object” (“non-background”) label if it projects onto the feasible regions (FR) in all views. The FR of each image is defined by the target highlight and cast shadow regions; corresponding to visible frontal and occluded surfaces in that view, respectively.  Eliminating the (non-object) background voxels, the collection of all “object voxels” yields the estimated 3-D volumetric model. How the reconstruction accuracy depends on the sonar poses, number of images, and target shape complexity are yet to be determined. Additionally, the binary classification is solely based on the FR membership; the intensity values that potentially encode shape information are not utilized. This is addressed by the 3-D model refinement scheme in [1] where the space carving solution is adjusted by minimizing the discrepancy between the data and synthesized object views (generated by a sonar image formation model [18]). Unfortunately the proposed iterative gradient-descent scheme generally converges to the nearby local minima of initial solution with noisy data. Hence, the 3-D shape improvement is generally marginal.}
In this and some other recent approaches [16], [17] , \cite{wang2021b}, accurate generation of 3-D models relies on a suitable sonar image formation model  \cite{aykin2016modeling, westman2020volumetric, wang2021}.

In [16], two different techniques are proposed. In one, equivalent to the online implementation of SC, the 
sonar is moved constantly to update the 3-D representation until the whole scene is
covered. Afterwards, an occlusion resolution is performed to retain only the frontal surfaces. In the second
method, images are captured during a rectilinear motion of a wide-aperture sonar along the elevation axis;
approximating the elevation arcs of sonar beams by vertical line segments. The
3-D reconstruction from image intensities is treated as a blind deconvolution with a spatially-varying kernel.
Inspired by  the non-line-of-sight (NLOS) paradigm and treating the scene as a non-directional albedo field, 
the interpretation of the image intensities for 3-D object modeling has been formulated as a discrete 
convex linear optimization problem [17]. 

The fundamental assumption in all earlier work is that FSS images are generated solely by the object echos, namely, 
the transmitted sound energy that is directly received by the target surfaces and reflected in the sonar direction. 
This generally applies to objects floating within water column, and sufficiently far from the seabed, sea surface, and other 
acoustic reflectors. 
For objects floating near a relatively flat water surface, we have to account for the multipath propagation. It has been shown that the water–air interface can transmit the low-frequency sound from underwater to the atmosphere \cite{ref33} to \cite{ref35}, thus not acting as a perfect mirror (as previously assumed). However, the interface still acts as a perfect reflector of high-frequency signals transmitted by the 2-D FL imaging sonars, leading to multipath propagation that can generate a mirror image (corresponding to a virtual object on the opposite side of the interface). The same mirror image can be formed for objects resting on (or near) a hard seafloor, where some portion of object echos incident upon the seabed can reflect towards the sonar receivers. The multi-path propagation in the opposite directions (of paths forming the mirror image) generate delayed object echos, (traveling longer paths along the same beams forming the object image) that form a ghost image.

Fig.~1a depicts the scene in a shallow pool, imaged with a rotated Oculus sonar to achieve a (relatively-large) 60\degree\,vertical field of view (FoV), but only 12\degree \,horizontally \cite{oculus}. This allows capturing the multipath due to both the hard bottom and water surface: Here,  we observe the adjacent mirror image on the opposite side of the bottom surface (due to the bottom reflection of object echoes), the mirror image on the opposite side relative to the air-water interface, as well as the ghost object formed behind the object region. 

{The multipath scenario also arises in deploying high-resolution near-range radar on autonomous road vehicles for environmental perception \cite{Kraus_2020} to \cite{Kraus_2021}, induced by the mirror-like reflecting surfaces within the complex road environments. In processing of the data, the ghost detection, identification, and removal is the main objective, in order to localize the real targets and surrounding reflection surfaces (e.g., other vehicles). In this work, we additionally aim to improve the 3-D target modeling by exploiting the visual cues from the multipath-induced ghost and mirror regions.}

 \begin{figure*}[t!]
    \centering
\begin{tabular}{cc}
  \hskip -.2in
  {\begin{tabular}{c}
    {\includegraphics[width=1.6in] {fig1_a-top_.pdf}}\\
    \hskip .08in
   \subcaptionbox{}[.255\linewidth]{\raisebox{-2ex}{\includegraphics[width=1.7in] {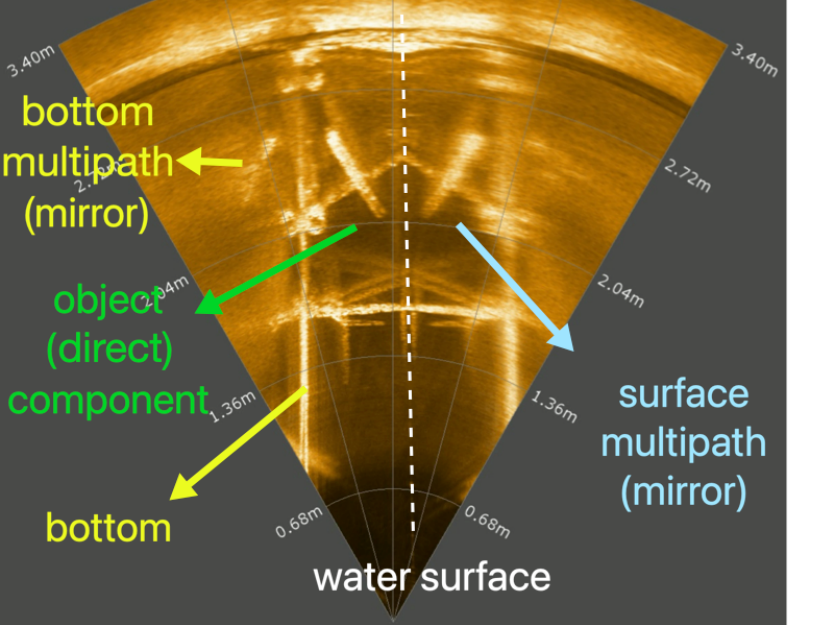}}}
   \end{tabular}}
&   \hskip .2in
  {\begin{tabular}{ccc}
    \subcaptionbox{}[.17\linewidth]{\raisebox{-2ex}{\includegraphics[width=1\linewidth] {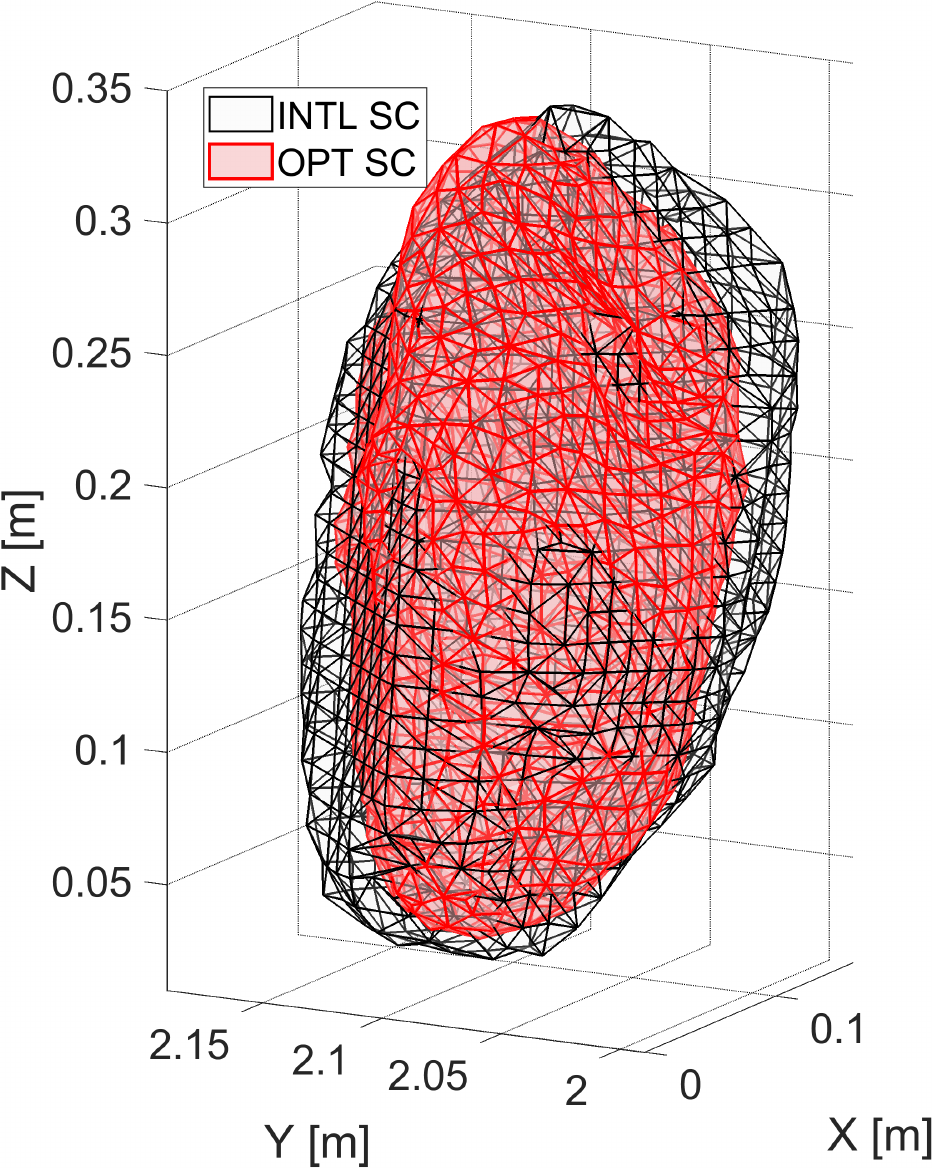}}} &
    \subcaptionbox{}[.2\linewidth]{\includegraphics[width=1\linewidth] {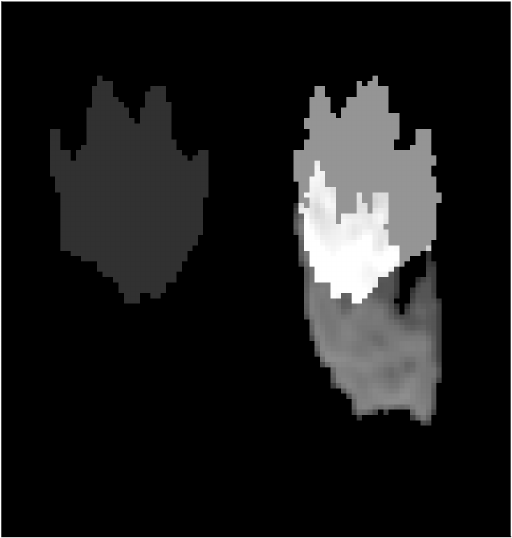}} &
    \subcaptionbox{}[.2\linewidth]{\includegraphics[width=1\linewidth] {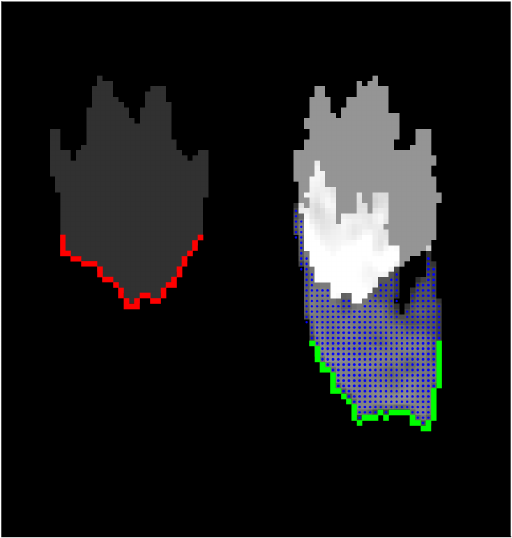}} \\
    \subcaptionbox{}[.2\linewidth]{\includegraphics[width=1\linewidth] {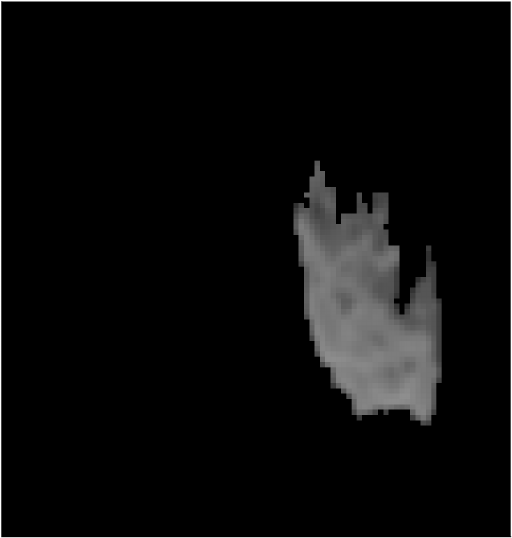}} &
    \subcaptionbox{}[.2\linewidth]{\includegraphics[width=1\linewidth] {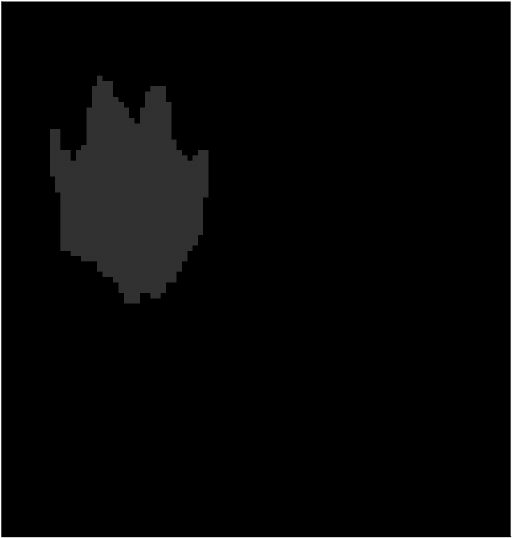}} &
    \subcaptionbox{\label{fig:syn_gho}}[.2\linewidth]{\includegraphics[width=1\linewidth] {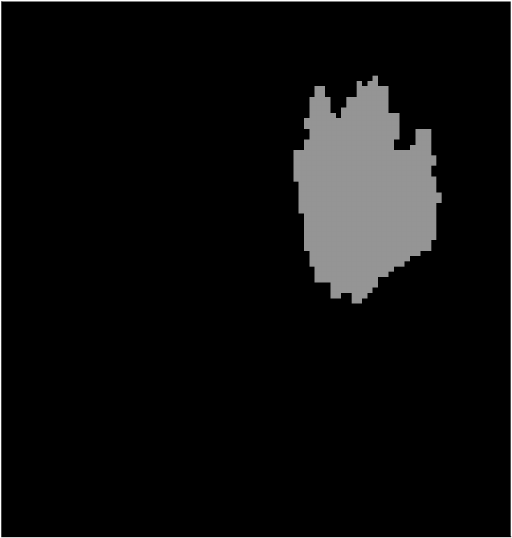}}\\   

\end{tabular}}
 \end{tabular}
\vskip -.1in
  \caption{(a) Sonar image of a scene in shallow pool, comprising of object, mirror and ghost components due to both surface and bottom multi path contributions. Black mesh in (b) is the initial 3-D target model to generate the synthetic sonar image (c), comprising of (e) object, (f) mirror and (g) ghost components. Optimizing the 3-D model by our method yields the red mesh in (b). (d)  lower parts of object (green) and mirror (red) contours and the non-overlapping region of object and ghost components (blue) used in the optimization method.}
  \label{fig:intro_to_syn_img}
  \vskip -.15in
  \end{figure*}

Summarizing our contributions, we adopt the iterative 3-D model enhancement paradigm in [1], but our optimization scheme rests on a completely different computational process: the iterative model refinement is by displacing the vertices in the 3-D surface mesh model, in accordance with the 2-D image motion fields that align the real and synthesized images in various views. {Fig.~1b to Fig.~1g highlights the process, starting with the black surface mesh in (b) as an initial 3-D model (of a coral rock), generated from roughly a dozen images at known sonar poses by space carving [2]. Among 
proposed methods for generating synthetic images of a 3-D object model \cite{aykin2016modeling, wang2021}, we require agreement with real data from a DIDSON unit, employed in this work \cite{aykin2016modeling}.
Moreover, we localize the ghost and mirror objects by modeling the multipath due to air-sea interface. 
A typical synthetic image for one sonar pose in (c) comprises of the target (shaded), mirror (dark gray), and ghost components (light gray), shown separately in (e-g), respectively. By the modeling of ground echo \cite{wang2023}, it is feasible to synthesize the ghost image, thus removed in applying our method. Similarly, air-water surface reflection modeling enables exploiting the intensity values within the mirror region. In practice, these requires accurate knowledge of multiple bounces among object, seabed, and the surface in shallow waters (as in example of Fig.~11a1), as well as reflectance characteristics of the seabed. The ghost and mirror regions are depicted uniformly-colored to emphasize the absence of  such information to reliably generate their images. Referring to (d), we remove the corrupted object region (white), utilizing collectively: 1) the intensity values within the uncorrupted object region (blue dotted); 2) lower object contour (green), and 3) lower part of the mirror-image contour (red). The iterative minimization of discrepancy between the data and synthesized views yields the red surface mesh model in (b).}
 
 The remaining sections are organized as follows. Section II covers the relevant technical background, including the projection geometry of imaging sonars, 
beam-bin data and transformation to range-azimuth coordinates to form the 2-D sonar image, and representation of poses at which the collected data is used to produce the initial 3-D model by SC [2]. In section III, we describe the 
modeling of multipath components. Section IV provides an overview of our optimization framework. In section V, we present the key steps from 2-D alignment of object regions in the data and synthetic views to the 3-D model 
refinement by 3-D vertex displacement, as well as define various error metrics. Section VI covers experimental results, based on two real and one mixed-real-synthetic data sets, and real and computer-generated data for initial insight into the impact of non-flat air-water interface. We summarize our contributions and describe future work in section VII.

 \begin{figure*}[t!]
 \begin{tabular}{cccc}
  \subcaptionbox {\label{fig:many_to_one_ambiguity}}      {\includegraphics[height=1.2in]  {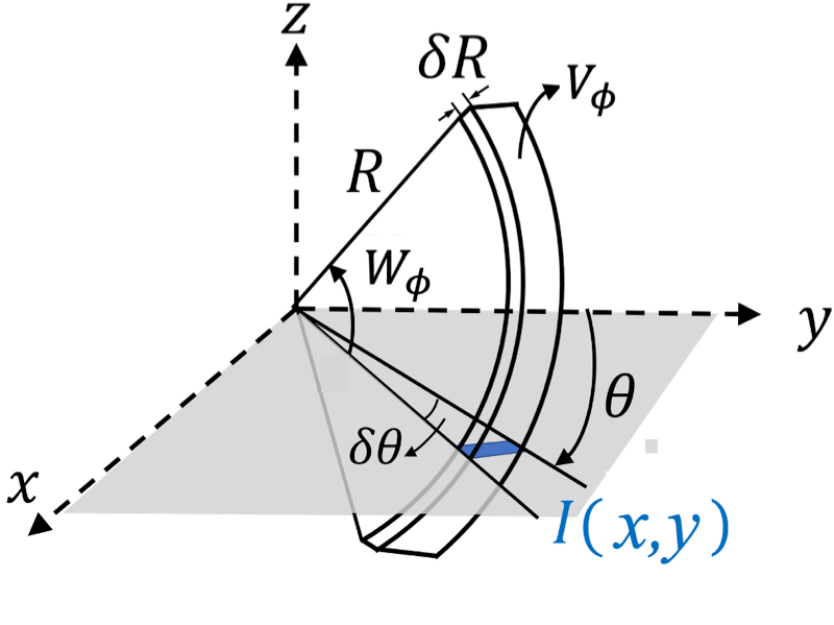}} &
  \hskip -.13in
        \subcaptionbox{\label{fig:sample_bin_beam_image}} {\includegraphics[height=1.2in]  {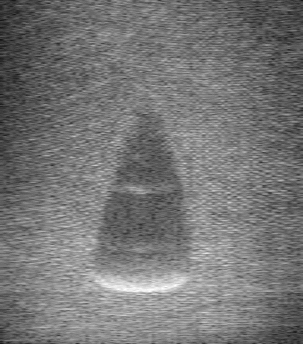}} &
    \hskip -.13in
    \subcaptionbox{\label{fig:sample_polar_image}}           {\includegraphics[height=1.2in]  {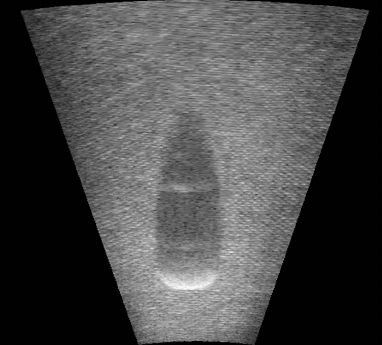}} &
    \hskip -0.1in
  \subcaptionbox {\label{fig:def_boundary}}                       {\includegraphics[height=1.2in]  {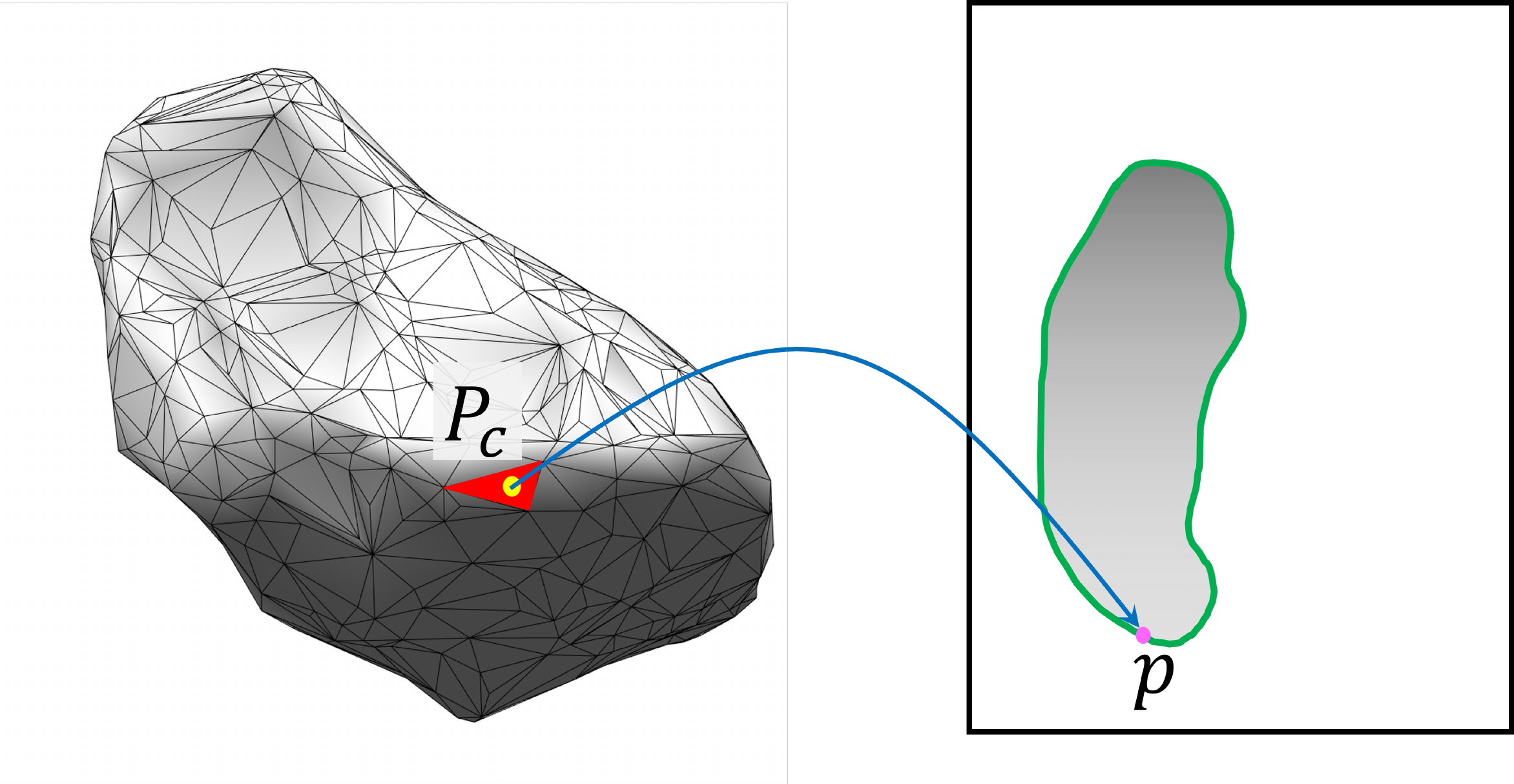}} \\
\end{tabular}
  \caption{(a) A beam in $\theta$ direction covers azimuthal interval [$\tet, \tet\!+\!\dtet$]. Image intensity $\!I\!$ of pixel $(x,y)$ depends on cumulative echos from unknown number of surface patches within volume $\!V_{\phi}\!$, arriving at sonar receiver simultaneously.
Along this beam, $V_{\phi}$ covers intervals  $[-W_{\phi},W_{\phi}]$ in elevation angle and  [$\Re$,\ $\Re\!+\dR]$ in range. DIDSON beam-bin data (b) and corresponding polar image (c) of a hemispherical rock on a shallow-pool bottom is corrupted by two near/far highlight bands within shadow region due to
bottom and surface multipath. (d) Boundary pixels (green), and a boundary patch (red) with its center (yellow) projecting onto a boundary pixel (magenta).}
 \vskip -.13in
   \end{figure*}

\section{Technical Background}
\label{techbackground}   
A 3-D point is denoted $\mmp= (X,Y,Z)$ in Cartesian and $\mpi=(\Re, \theta, \phi)$ in spherical coordinates, transforming according to 

 \begin{equation}
  \label{eq:cartesian_to_spherical}
\hskip -.15in
     \begin{bmatrix}
     \!    X \!\\
     \!    Y\!\\
     \!    Z\!
     \end{bmatrix} \!\!=\! \!\Re\!\begin{bmatrix}
         \!\cos\phi\sin\theta\\
         \cos\phi\cos\theta\\
         \sin\phi\!
     \end{bmatrix}\,
\, 
     \begin{bmatrix}
         \Re\\
         \theta\\
         \phi
     \end{bmatrix}\!\!=\!\!\begin{bmatrix}
         \sqrt{X^2 + Y^2 + Z^2}\\
         tan^{-1}(X/Y)\\
         tan^{-1}(Z/\sqrt{X^2+Y^2})
     \end{bmatrix}
 \end{equation}


Fig.~2a depicts one of $N_b$ acoustic beams, each with a narrow horizontal width $\delta\theta$ and elevation arc $2W_{\phi}$, transmitted by the sonar in a known azimuth direction $\theta$. Collectively, the beams cover $-W_{\tet} \le \tet \le W_{\tet} $ in horizontal field of view (6 [deg] $\leq W_{\phi}\leq$ 10 [deg] and 15 [deg] $\leq W_{\tet} \leq$ 65 [deg] for most existing FSS). 

At various sample times, the round-trip times of flight $T_{tof}$ of target echos in the same $\theta$ directions yield the range measurements  $\Re=\frac{1}{2}T_{tof}\upsilon$, where $\upsilon$ is the sound speed within the medium. We can calculate $(\Re,\tet)$ from the so-called ``beam-bin'' $(b,B)$ coordinates:
  \begin{equation}\label{BbtoRt}
\begin{array}{rl}
 \Re&\hskip-.1in =\Re_{\mbox{min}}+(B-1)\dR \;\; (B=\ 1,\ 2,\ \ldots,\ N_B) \\
 \theta&\hskip-.1in= \tet_{\mbox{min}}+(b-1)\,\delta\tet\quad(b=\ 1,\ 2,\ \ldots,\ N_b)
\end{array}
 \end{equation}
where $N_B$ is the number of range bins covering the range window $\{\Re_{\mbox{min}},\Re_{\mbox{max}}\}$ with range resolution $\dR=(\Re_{\mbox{max}}-\Re_{\mbox{min}})/(N_B-1)$. We can confirm or adjust these models by calibration, e.g., the transformation of beam number $b$ to azimuth angle $\tet$ in \eqref{BbtoRt} for a lens-based DIDSON follows a mild cubic due to lens distortion \cite{DIDSON}.
 
The intensity array $I(b,B)$ encodes the strength of (collective) echoes from potentially several scene surface patches within the elevation arc $[-W_{\phi},W_{\phi}]$, leading to the many-to-one projection ambiguity of FSS imaging. Referring to Fig.~2b and Fig.~2c, the 2-D polar image $I(x,y)$ is constructed from the bin-beam image using the transformation
\begin{equation}
  \begin{bmatrix}
      x\\
      y\\
  \end{bmatrix} = \Re\begin{bmatrix}
      \sin\theta\\
      \cos\theta\\
  \end{bmatrix}
  \label{eq:polar_img_coordinate}
\end{equation} 
The approximation $0.99\!<\!\cos\phi\!\approx\! 1$ ($\abs{\phi}\le 7\degree$) is used in one step of our iterative optimization algorithm 
as follows:  
 \begin{equation}
     \mmp =  \Re\begin{bmatrix}
         \cos\phi~\sin\theta\\
         \cos\phi~\cos\theta\\
         \sin\phi
     \end{bmatrix}
\approx \begin{bmatrix}
       \Re\sin\theta\\
       \Re\cos\theta\\
         \Re\sin\phi \\
           \end{bmatrix} =
\begin{bmatrix}
         x\\
         y\\
         \Re~sin\phi
     \end{bmatrix}
     \label{eq:cartesian_to_spherical_approx}
 \end{equation} 
 
 \noindent 
{We utilize the image formation model in [18] to synthesize the target views at known sonar poses. The SC solution is our initial 3-D model $S^o$, generated from the data captured with a lens-based \textit{D}ual-Frequency \textit{ID}entification \textit{SON}ar (DIDSON) [19]. 
Our real data is collected according to the guidelines in [2] for improving the carving process and generating a more accurate 3-D model. To this end, the sonar is rotated around the viewing direction into $M_r$ angles in equal increments
$d\om_r=\pi/M_r$ [rad], from $-\pi$ to $-\pi+(M_r-1)d\om_r$ at each of $M_p$ positions around the object (for a total of $M = M_pM_r$ images).
Both $M_p$ and $M_r$ depend on target shape complexity, as demonstrated by experiments in [2]. Selecting a reference pose, the pose for each image $I_{i,j}$ is parametrized by $\vm_{i,j}=[\vr_j^T,  \vt_i^T]^T$, with relative 3-D rotation and translational components $\vr_j = [r_1, r_2, r_3]_j^T$ and $\vt_i=[t_1, t_2, t_3]_i^T$, respectively. For simplicity, $I_k$ denotes the $k$-th image at pose $\vm_k$; $\vm_o=\zr$ denotes the reference pose.} 
 
\begin{figure*}[t!]
 \vskip -.3in
  \begin{tabular}{cc}
  \raisebox{2ex}{\includegraphics[width=.385\linewidth] {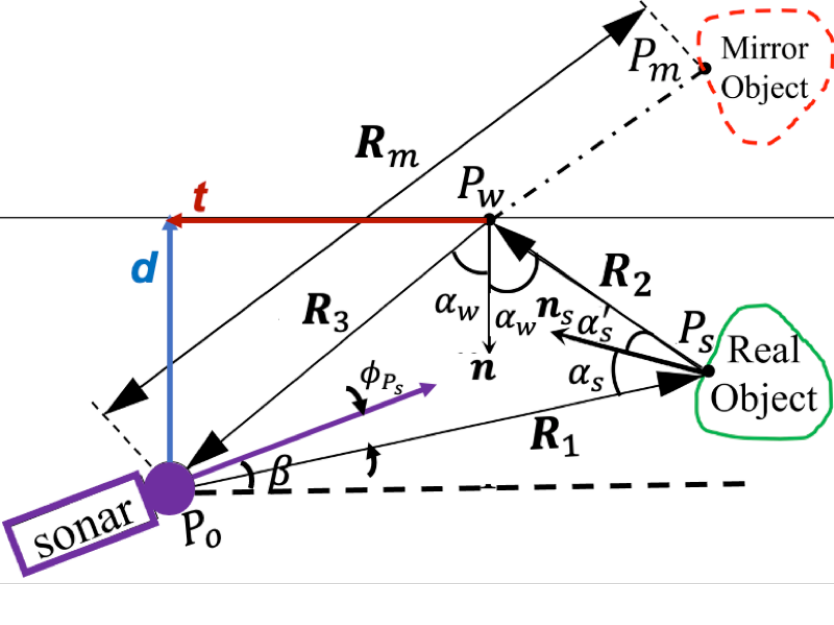}} &
   \raisebox{-8ex} {\hskip .3in\includegraphics[width=.5\linewidth] {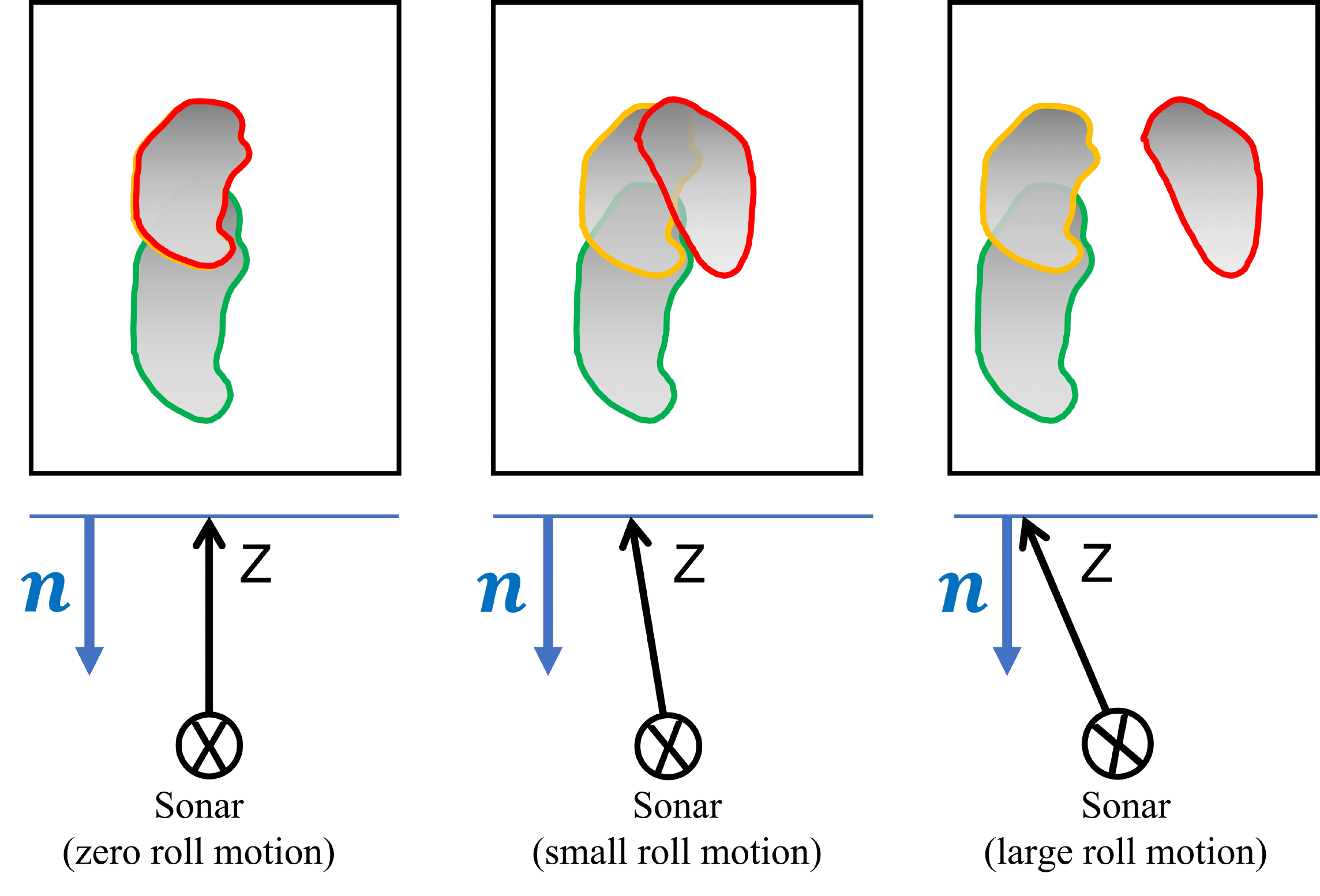}}\\
  \end{tabular} 
 \vskip -1.in
   \begin{tabular}{cc}
 \raisebox{0.5ex}{\hskip 1.3in (a)} & \raisebox{-2ex}{\hskip 2in} \\
  \end{tabular} 
  \vskip -.02in
  \begin{tabular}{cc}
  \raisebox{16ex}{
  {\includegraphics[width=.44\linewidth] {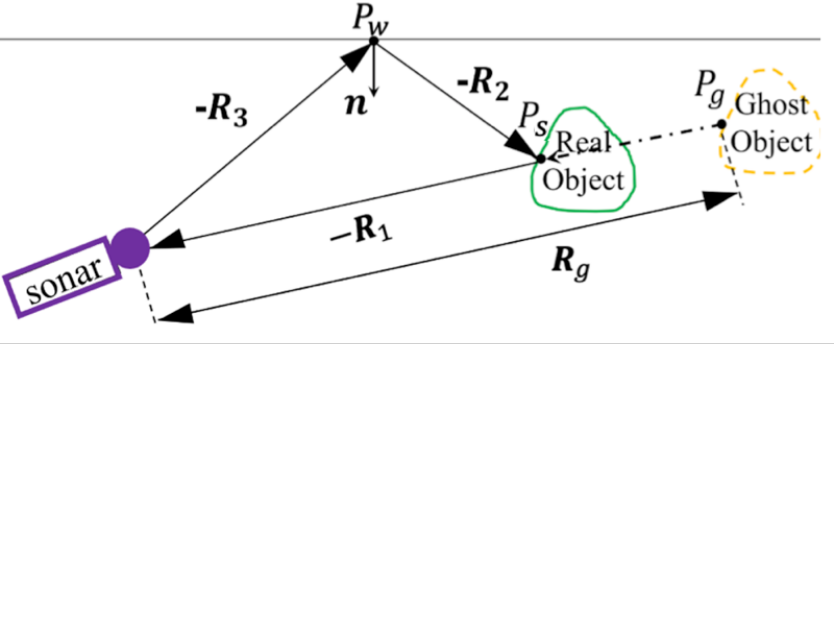}}} &
   \\
  \end{tabular}
    \begin{tabular}{cc}
  \raisebox{13ex}{\hskip -3.9in (b)}  &  \raisebox{13ex}{\hskip 1in (c)} \\
  \end{tabular} 
  \vskip -2.1in
  \caption{(a) Mirror image geometry: transmitted sound waves traveling along $\mr_1$, are scattered at object point $P_s$. Reflected component, propagating along ``unique direction'' $\mr_2$ to point $P_W$ at water surface with normal $\vn$, is specularly reflected towards the sonar along $\mr_3$.  (b) Ghost image geometry: traveling along mirror-image pathway in reverse direction, sound waves along $-\mr_3$ are specularly reflected towards the object along $-\mr_2$. Scattered at $P_s$, component along $-\mr_1$ is captured by the sonar.  (c) Relative locations of object (green), mirror (red) and ghost (yellow) components vary as sonar rolls. Object region overlap with ghost (and possibly mirror) component(s) leads to contour distortion and intensity corruption.}
 \label{fig:formation_mirror_ghost}
\end{figure*}

\section{Multipath Modeling}

Referring to Fig.~3a, the sea surface, assumed to be relatively flat, acts as an ideal specular reflector due to the large mismatch in acoustical impedances of air and water. A portion of object echo, arriving at the unique water surface point $P_W$, specularly reflects in the direction of sonar receivers to generate the mirror image (of a virtual object on the opposite side of the interface). According to Fig.~3b, the air-water interface also acts as a secondary source, specularly reflecting the incident sonar beams in the target direction. The ghost image is formed by the delayed echos (with longer travel paths) reaching the sonar along the same beams forming the object image.

In Fig. 3a,  $\beta$ denotes the pitch angle of the sonar relative to the flat water-surface in the base (unrotated) orientation. For a selected target point $P_s$, the water surface normal $\vn$ is (primarily) in the azimuthal plane of the sonar beam in the direction of $\mr_1$.  The air-water interface point $P_W$, where the echo in the direction $\mr_2$  is specularly reflected in the direction $\mr_3$ towards sonar receivers, is fixed by $\vt=\abs{\vt}\hvt$:\footnote{We can treat other sonar poses based on the known sonar rotation.}
\begin{equation}\label{multipath}
\hskip-.05in \hvt\!=\!\!\frac{\displaystyle 1}{\displaystyle \cos\!\beta'}\!(\hvn\times(\hvn\times\mr_1))\;\,
\abs{\vt}\!\!=\!\!\frac{\displaystyle  \abs{\vd}\, \abs{\mr_1} \cos\!\beta'}
                   {\displaystyle 2\abs{\vd} -\abs{\mr_1} \sin\!\beta'} \;\, \beta'\!\!=\!\beta\!-\phi_{P_s}
\end{equation}
Subsequently, we can compute 
\begin{equation}
\mr_3=\! \vt-\vd \quad\; \mr_2=\!-(\mr_1+\mr_3)
\end{equation}
To form  a ``mirror image'' (associated with a virtual ``mirror object'' on the opposite side of water surface), the direction of $\mr_3$ is constrained by the (vertical) field of view $\abs{\angle{\mr_3}} \le W_{\phi}$. 
The location of virtual mirror-object point $P_m$ can be derived for each surface point $P_s$:
 \begin{equation} 
   \label{eq:mir_img_geo}
   \hskip -.07in
   \mmp_m\!\!=\!\abs{\mr_m} (-\widehat{\mr}_3\!) \,\, \abs{\mr_m} \!=\! \frac{1}{2} (\abs{\mr_1} + \abs{\mr_2}+ \abs{\mr_3}) \!\ge\!\!\abs{\mr_1} 
 \end{equation}
The shapes of object and mirror regions somewhat differ because the corresponding acoustic returns are received from different directions, becoming more dissimilar for rotated sonar poses with larger roll angles;  see Fig. 3c.

Referring to Fig. 3b, the sonar beams propagating in the reverse directions (of mirror image pathways) are collected by the same sonar receivers that record the direct echos\footnote {In practice, these directions may vary slightly due to the discrete nature of sonar beams.}. The longer path $\abs{\mr_g}=\abs{\mr_m}$ of these delayed echoes generates a ghost image (of the virtual ``ghost object'') at longer ranges. The multi-path components can be significant for strong surface echo along $\mr_2$ direction; e.g., for $\alpha_s\approx \alpha_s'$ for surfaces with rough specular reflectance [20]. 

As depicted in Fig. 3c, the separation of mirror image from the other two components is controlled by the sonar pose, increasing with the sonar rolling motion; i.e. to acquire $M_r$ rotated views in our dataset. The ghost image generally overlaps with the more distant part of object image, corrupting both the contour and intensity values within the overlapping region. However, the frontal object region closer to the sonar remains undistorted.

\section{Computational Framework}\label{sec:optimization}
The initial 3-D surface model $S^o$, generated by SC, is a closed mesh of $N_T$ triangles formed by $N_p$ vertices $\mmp^o_{P_l}$ ($l=1,\  2,\ldots,\  N_p$), as in Fig. 2d. Moreover, a triangulation connectivity list $T_s$ defines triplet point sets $\{\mmp^o_{P_{j1}},\mmp^o_{P_{j2}},\mmp^o_{P_{j3}}\}$ forming the triangles in $S^o$. 

Given a 3-D object model $\TS$, we can generate a synthesized image $\TI_m$ at each sonar pose $\vm_m$ by applying the image formation model in [18]. In particular, we can determine the number and contribution of the surface patches of  $\tS$ to the intensity of each pixel. In [1],  a gradient descent updating scheme is applied to seek the optimum 3-D model $S^*$ that minimizes the discrepancy between the data and synthesized images constructed:
\begin{equation}\label{objfunction}
S^*= \arg\min_{\TS} E_{S}=\sum_{m} \rho({I_{m}-\TI_{m}(\TS))})
\end{equation}
where  $\rho$ is a suitable norm. Due to high {\it non-convexity of energy function} $E_S$ and  {\it  sensitivity to data noise},   convergence to one of multitude of local minima, typically in the vicinity of initial SC solution, leads to negligible model enhancement. Here, a more efficient and effective method based on a completely different computational process is proposed.

 \begin{figure*}[tb!]
  \centering
  \vskip -.1in
  \begin{tabular}{cc}
  \includegraphics[height=3.6in] {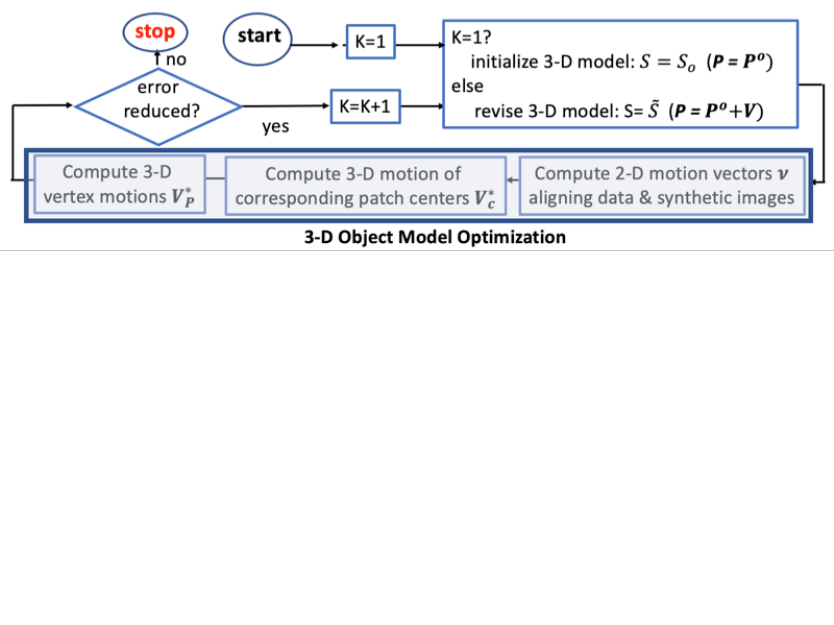} &
  \raisebox{33ex}{\includegraphics[height=1.55in] {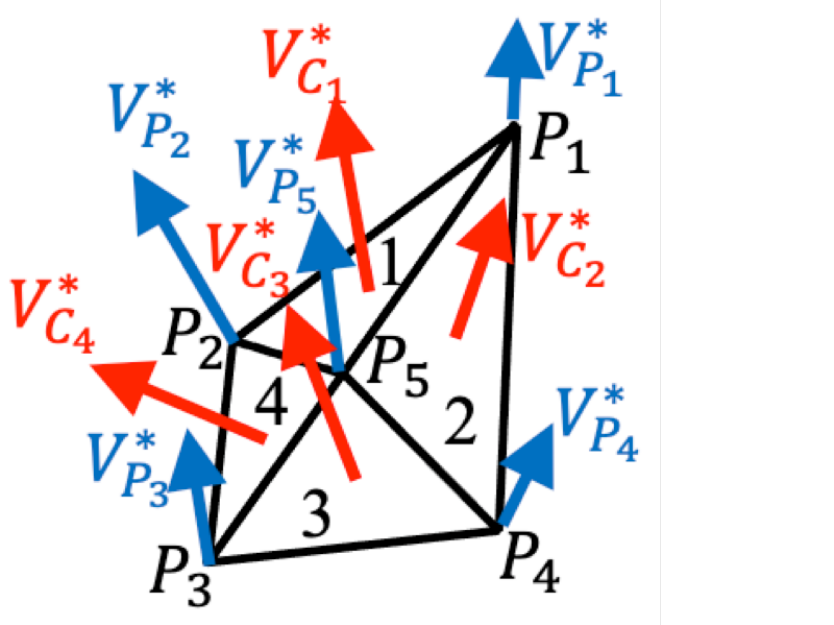}}  \\
  \end{tabular}
\vskip -2.15in
  \begin{tabular}{cc}
  \hskip 1.3in (a)  & \hskip 3.1in (b) \\
  \end{tabular}
\vskip -.015in
  \caption{(a) Block diagram of entire algorithm; (b) 3-D motions of vertices and patch centers; sample example
shows that each vertex $\mmp_k$ appear in two or more patches, and its motion $\mv^*_{P_k}$
have to be consistent with motions  $\mv^*_{c_i}$ ($i=1,\ 2,\ \ldots,\  p$) of $p$ neighboring patch centers.\label{fig:optimization_overview}}
\vskip -.15in
\end{figure*}

We refine the 3-D Model $\tS$ by displacing all 3-D vertices  $\mmp_{P_l}=\mmp_{P_l}^o+\mv_{P_l}$ ($l=1, 2,\;\ldots, N_p$) of the triangular mesh simultaneously. To be precise, we maintain the initial number ($N_p$) of vertices and triangulation connectivity $T_s$ of $S^o$, but vary the size and orientation of triangular patches through suitable displacement  $\mv_P=\, [\mv_{P_1},\, \mv_{P_2}, \, \ldots, \, \mv_{P_{N_p}}]$. These are derived iteratively by matching the data and synthesized images of the estimated 3-D model over available views. To this end, each iteration $k$ involves: 1) computing 2-D motion fields $ \pmb{v}_m^k$ to register the data and synthesized images at ``relevant'' sonar poses $\vm_m$ ($m=1, 2, \ldots M^{-}\le M$)\footnote{As we elaborate, some $(M-M^{-})$ views may not be suitable to use, thus we term $M^{-}$ poses as ``relevant views.''}; 2) transforming the 2-D motion fields $\pmb{v}_m^k$ to optimum 3-D vertex displacement field $\mv_P$ in two steps, as depicted in the block diagram of the complete process in Fig. 4a; 3) revising the 3-D model $\TS^k$. 

\section{From 2-D Image Alignment to 3-D Model refinement}\label{sec:method}
Referring to Fig. 2d, points on the object image boundary will be referred to as “boundary pixels” (green).
Moreover, “boundary patch” (red) will refer to any 3-D triangular surface patch that projects onto a particular
boundary pixel $\vp$ (magenta), represented by triangle center $\mmp_c$ (yellow).

The lower uncorrupted part of object boundary conveys useful information for the alignment of synthetic
image with the data, consequently providing guidance on how to refine the 3-D model. We utilize the term
``frontal contour,'' for the lower one-third of the detected object boundary $\Re\le \Re_{\mbox{min}}+W_{\Re}/3$ (
$W_{\Re}= \Re_{\mbox{max}}-\Re_{\mbox{min}}$). This is determined experimentally to be a good trade-off between
true contour integrity (avoiding overlap with ghost region) and containing sufficient data points to achieve robust alignment. 
We perform contour alignment by the Iterative Closest Point (ICP) algorithm [21], [22], namely, the IRLS-
ICP variant [23], [24], [25]. Applicability for registering two unequal-size unmatched point sets offers a key
advantage in our application.

Fig. 5 depicts one view from our experimental data set, depicting the ghost-corrupted image of a coral rock, 
the mirror component, and the detected contours. After contour alignment, the uncorrupted frontal parts 
match well, and outliers mainly due to corrupted region are discarded.  We employ correlation-based matching within the uncorrupted object interior, and apply a weighted linear combination of estimates from the two schemes to achieve smooth transition; weight is set based on proximity to the frontal boundary, as explained next.

 \subsection{Image Contour Alignment}
\label{sec:determining_2D_motions_by_image_contour}

\begin{figure*}[t!]
  \centering
  \begin{tabular}{cccc}
  \subcaptionbox{\label{fig:original_sonar_img}} {\includegraphics[height=1.4in] {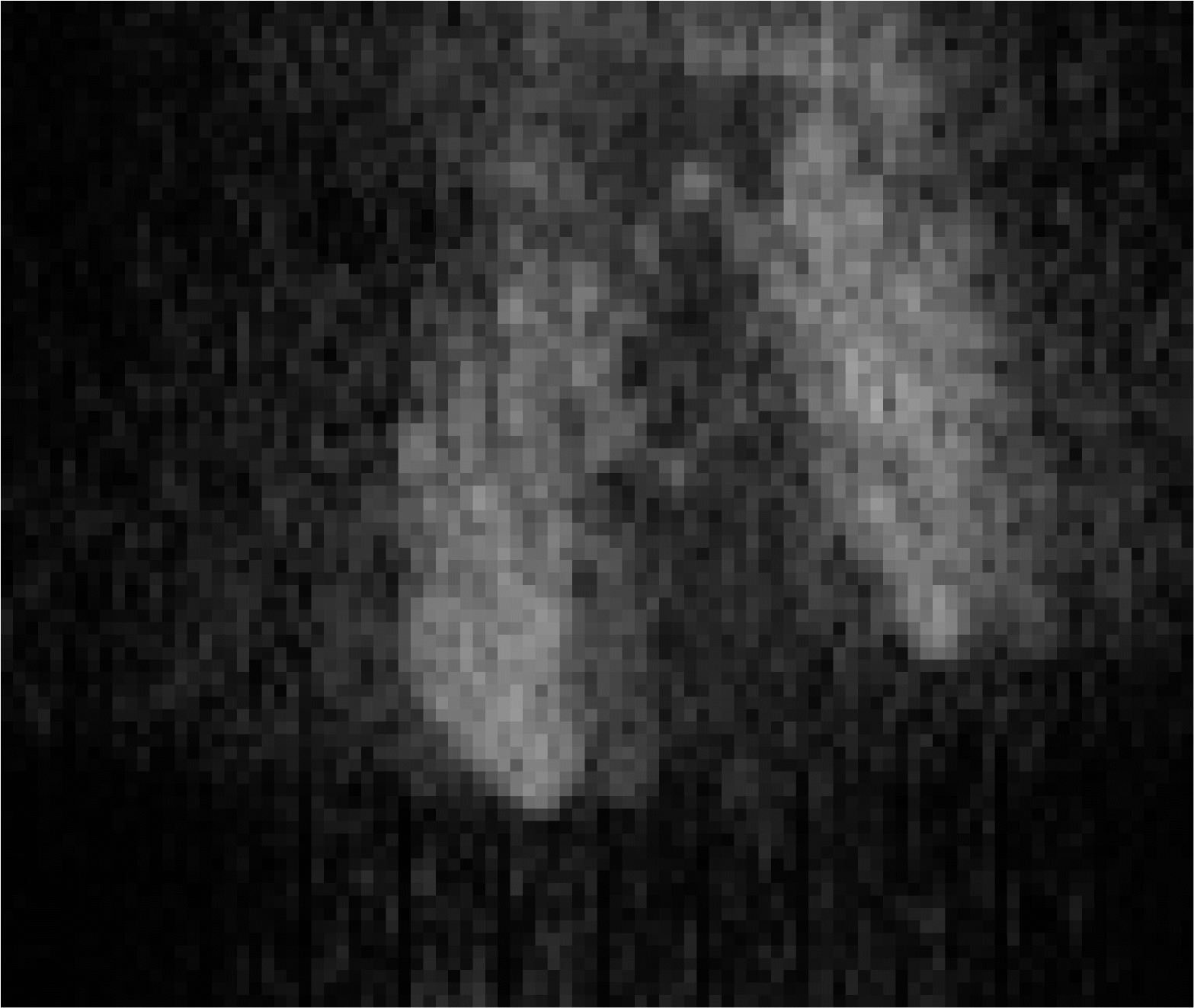}} &
    \hskip-.15in
  \subcaptionbox{\label{fig:binarized_sonar_img}}{\includegraphics[height=1.4in] {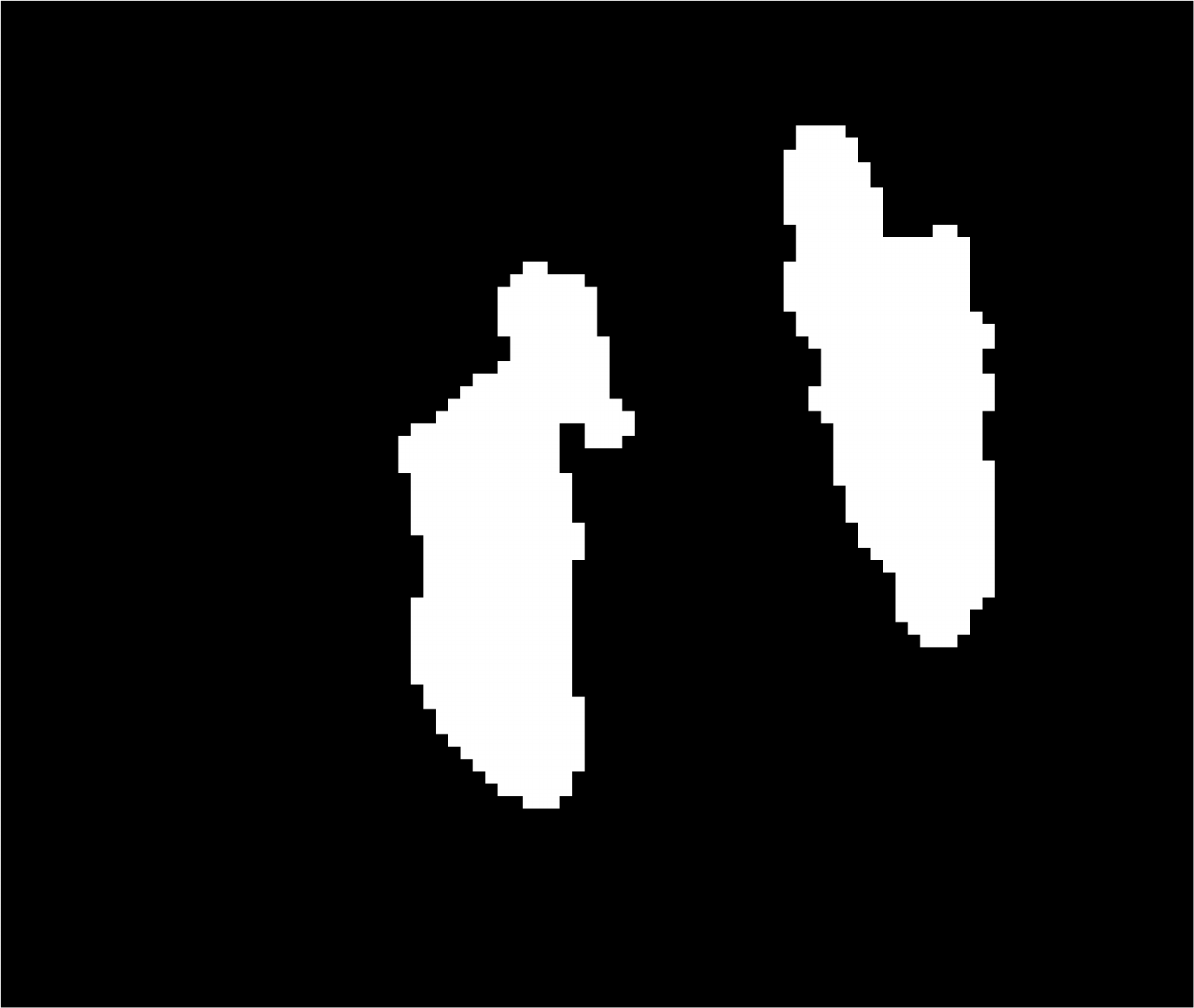}} &
   \hskip-.15in
  \subcaptionbox{\label{fig:contour_tracing}}         {\includegraphics[height=1.4in] {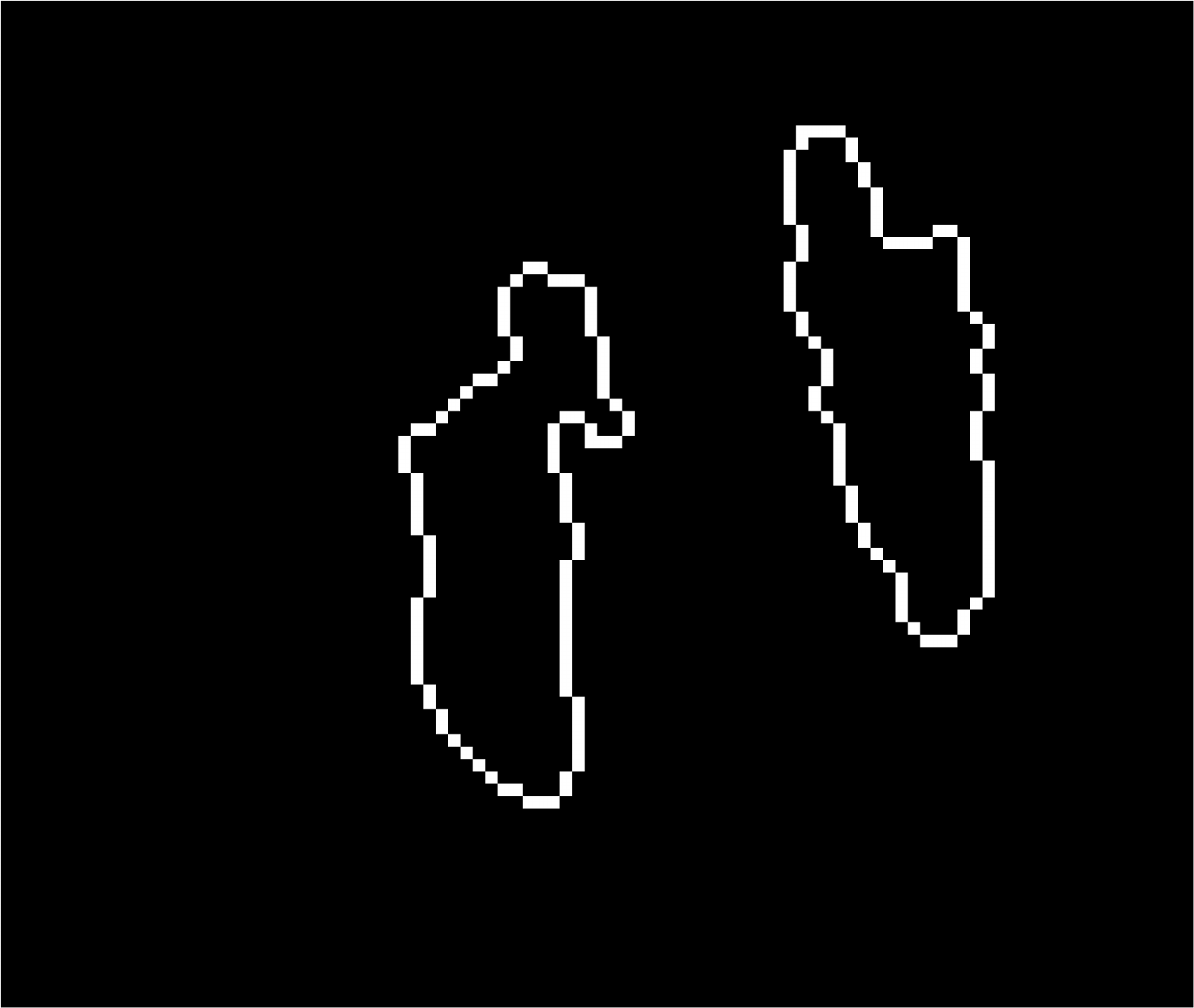}} &
   \hskip-.15in
  \subcaptionbox{\label{fig:sample_contour}}       {\includegraphics[height=1.4in] {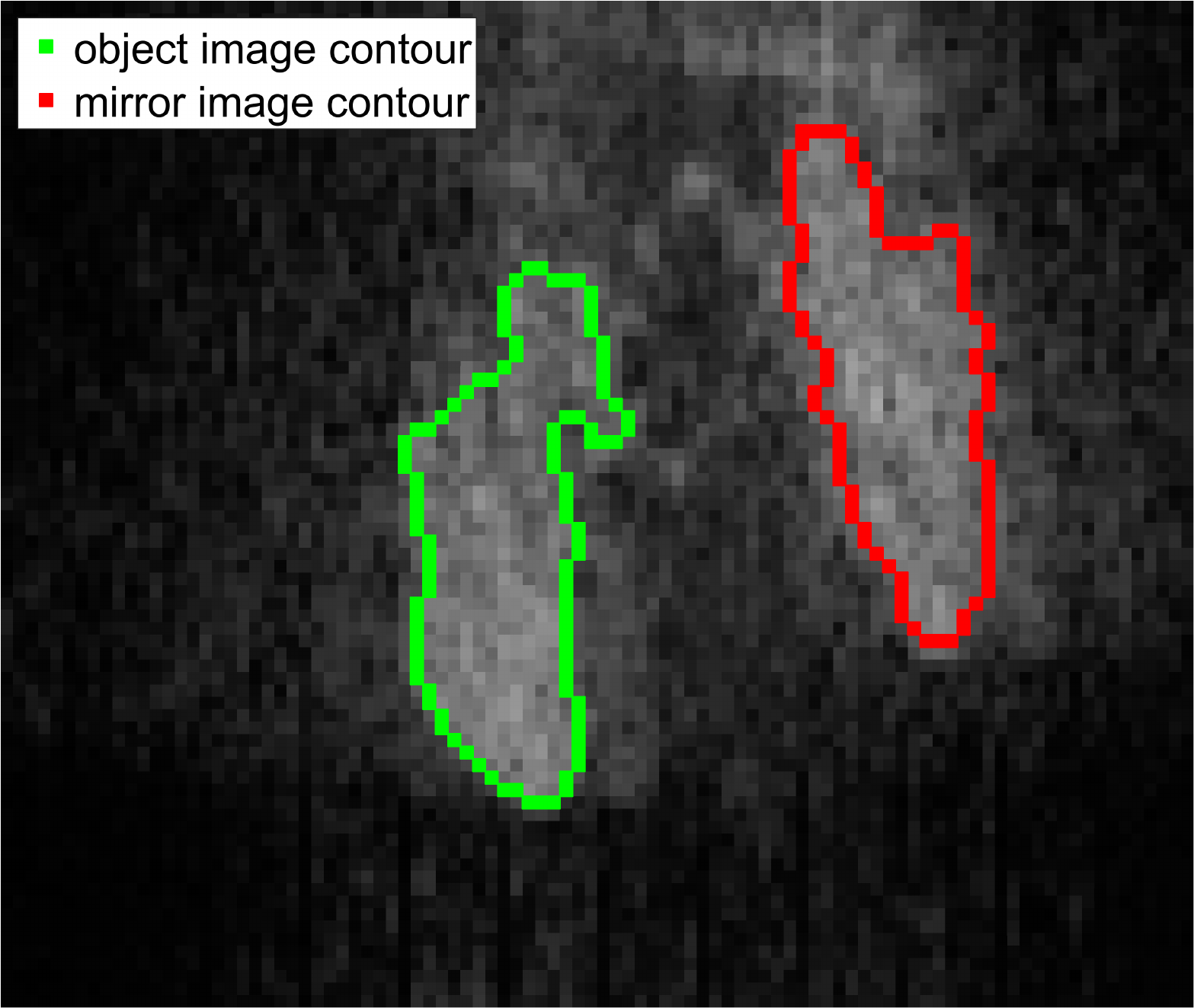}}\\
  \end{tabular}
  \vskip-.1in
  \caption{(a) Beam-bin image of coral rock near water surface with mirror and ghost components; (b) binary
image; (c) extracted contours; (d) contours superimposed on image (green: object and ghost regions; red:
mirror).
 \label{fig:contour_aligment}}
  \vskip0in

  \centering
  \vskip 0in
  \begin{tabular}{cc}
   \hskip -.05in\includegraphics[height=2.75in] {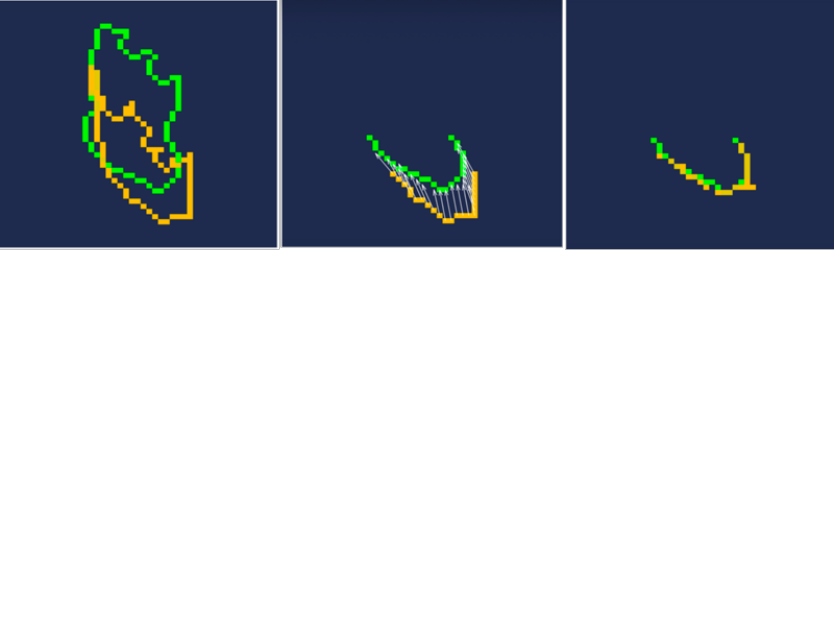} &
  \hskip -.2in \raisebox{10ex} {\includegraphics[height=2.35in] {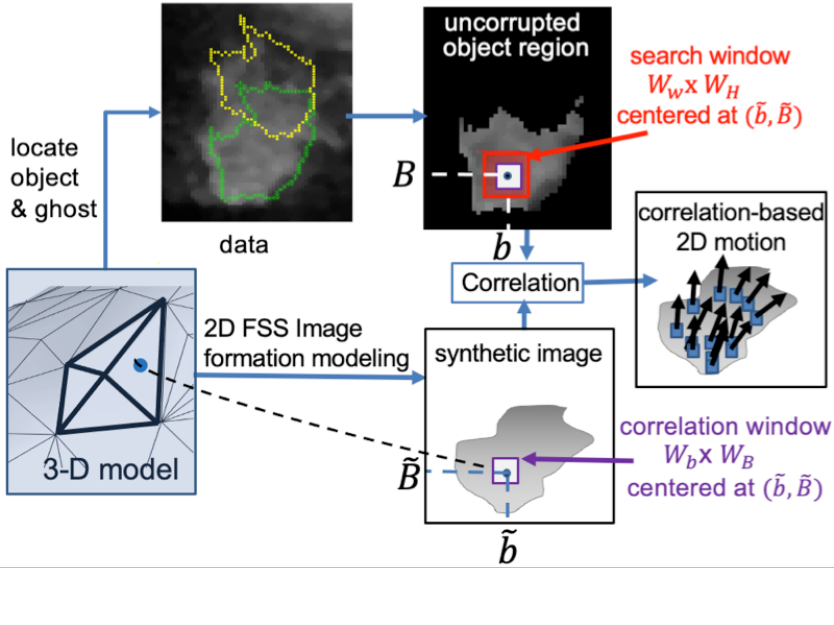}}\\
    \end{tabular}
      \vskip-1.35in
      \begin{tabular}{cc}
      \raisebox{6ex}{\hskip 0.5in (a)}  & \hskip 2.7in (b) \\
      \end{tabular}
   \vskip 0.in
  \caption{(a): object contours in original real (green) and synthetic (orange) images, with motion vectors (white)
to align frontal contour, and final alignment (data from Fig.~\ref{fig:contour_aligment}); (b) correlation-based 
  matching.}  \label{fig:align_frontal_cntr_icp}
   \vskip -.15in 
  \end{figure*}
 
Let $\widetilde{\mathcal{C}}_m = \{\tc_{mj}~|~j=1,2,..,\tn_m\}$ and $\mathcal{C}_m = \{c_{mi}~|~i=1,2,...,N_m\}$ ($\tn_m\not=N_m$, generally) be the frontal contour pixels $\tc_{mj}=(\tx_{mj},\ty_{mj})$ and 
${c}_{mi}=({x}_{mi},{y}_{mi})$ in pose $\vm_m$ of synthetic image $\TI_m$ and real data $I_m$, respectively. Registering $\widetilde{\mathcal{C}}_m$ and ${\mathcal{C}}_m$ yields: 1) 2-D Euclidean transformation $\tilde{c}'_{mj} = T(\tc_{mj})$ mapping $\tc_{mj}$ to $\tilde{c}'_{mj}$; 2) $\tn'_m \leq \tn_m$ matching pixel pairs $\{\tilde{c}'_{mj},c_{mi}\}$ in the registered synthetic contours $\mathcal{\widetilde{C}}^{\prime}_m=\{\tilde{c}'_{mj}~|~j=1,2,...\tn'_m\}$ and ${\mathcal{C}}_m$, denoted $i(j)$. The 2-D motion of $\tc_{mj}$ is given by $\pmb{v}^C_{mj}=\tc^{\prime}_{mj} - \tc_{mj}$; shown in Fig.~\ref{fig:align_frontal_cntr_icp}a for the corresponding beam-bin images. The matching quality can be assessed by
\begin{equation}
  \lambda (\mathcal{\widetilde{C}}^{\prime}_m, {\mathcal{C}}_m) \!=\! 
    \frac{1}{N'_m}\!\sum _{j=1}^{N'_m} 
    \!\sqrt {(\tx'_{mj}\!-\!{x}_{mi(j)} )^2\!+\! (\ty'_{mj}\!-\!{y}_{mi(j)})^2}
    \label{eq:contour_matching_error}
\end{equation}
The beam-bin coordinates $(b,B)$ of contour points $\mathcal{C}$ in $I(b,B)$ yield the range $\Re$ and azimuth angle $\tet$, and subsequently the polar coordinates $(x,y)= \Re(\sin{\tet},\cos{\tet})$.  Any pose for which the matching error $\lambda$ exceeds a threshold (determined empirically and fixed at 1 [cm] in all of the experiments) is deemed inaccurate, and excluded from the 2-D motion computations. As we show, a small subset of all views still provides redundancy.

\subsubsection{Correlation-Based Registration}
\label{sec:find_2D_move_by_correlation}
Referring to Fig. 6b, a surface patch center projects to some pixel  $(\tb,\tB)$ in the synthetic beam-bin array $\TI_m$ for pose $\vm_m$. We seek the matching pixel $(b,B)$ within a $W_{\small W} \times W_{\small H}$ region centered at $(\tb,\tB)$ in data array $I_m$.
As applied for image registration and disparity computation [26], [27], we employ the Pearson correlation
coefficient (PCC) $C_{pc}$ as the measure of pattern similarity, over a $W_{b}\times W_{B}$ window, centered at $(\tb, \tB)$ and $(b, B)$:\footnote{fixed to $W_b=5$ beams and $W_B=7$ range bins, in all experiments.} 

\begin{equation}
  \label{eq:pearson_correlation}
  C_{pc} = \frac{\sum_{i=1}\left( I_i-\overline{I}\right) \left( \TI_i-\overline{\TI}\right) }{\sqrt{\sum_{i}\left( I_i-\overline{I}\right) ^{2}\sum_{i}\left( \TI_i-\overline{\TI}\right) ^{2}}}
\end{equation}
\noindent where $\overline{I}$ and $\overline{\TI}$ denote the averages of paired datasets $\{I_i, \TI_i\}$ with unequal means and variances. Each pixel $(b,B)$ as potential match of $(\tb,\tB)$ within the search window is assigned the PCC value $C_{pc}(b,B)$. 

\begin{figure*}[t]
\vskip -0.3in
\begin{tabular}{ccc}
\hskip -.1in
  \subcaptionbox{\label{fig:r_curve}}        {\includegraphics[width=2.12in] {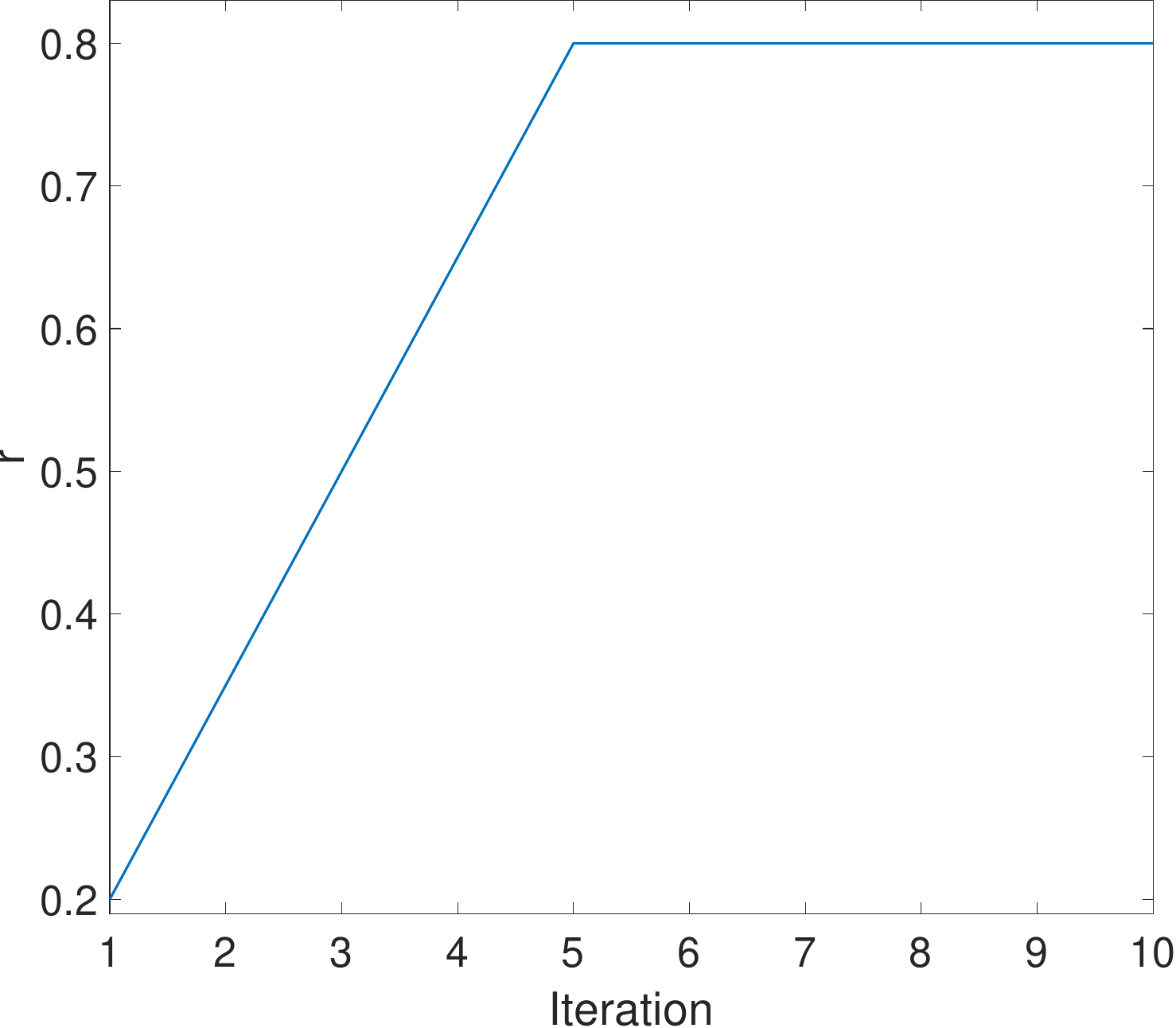}} &
\hskip -0.1in
  \subcaptionbox{\label{fig:score_curve}} {\includegraphics[width=2.12in] {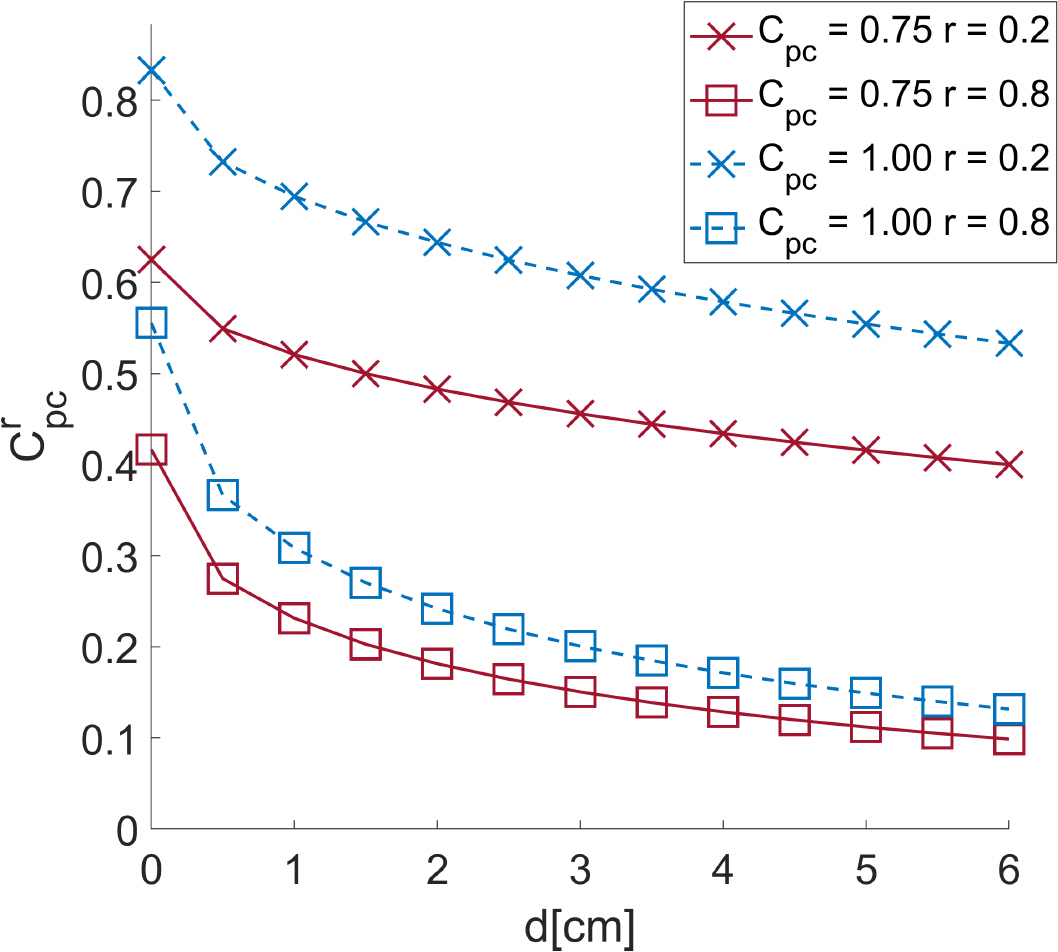}} &
  \raisebox{-1ex}{\hskip .05in
  \subcaptionbox{\label{fig:alpha_curve}} {\includegraphics[height=1.9in] {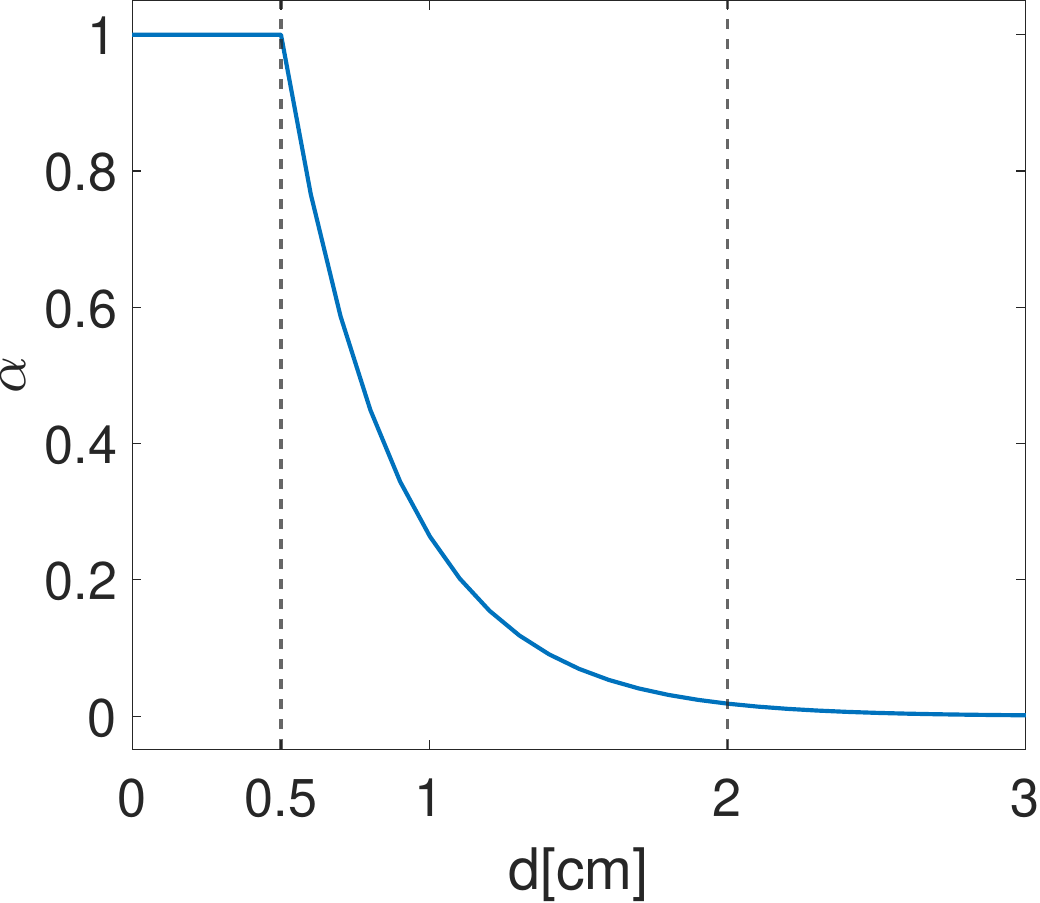}}} \\
\end{tabular}
\vskip -.05in
 \caption{(a) Variation of regularization term $r$ with iteration number 
  (b) similarity score $C_{pc}^r$ discounts PCC based on motion size $d$ and $r$. Initially (with $r=0.2$), larger motions $d>1$ [cm] with PCC value $C_{pc}=1$ outscore lower $C_{pc}$ values for smaller motion sizes $d\le1$ [cm].  This is reversed after few iterations, e.g., for $d\ge 2.5$ [cm] with $r=0.8$ (at 5 or more iterations). (c) exponentially-decaying scale $0\leq \alpha\le 1$ for 2-D motion estimation based on contour and correlation-based matching.}
  \label{fig:score_and_r_alpha}
  \vskip -.1in
   \end{figure*}

To reduce the PCC sensitivity to data noise within regions with negligible gradient, we introduce the scoring function $C_{pc}^r$ as a 
“regularized” PCC  based on the 2-D motion size and iteration number.  To elaborate, we note that a reasonably accurate 3-D 
model (as assessed by the error criteria introduced later) requires small adjustments by 3-D vertex displacements, 
and thus to align the 2-D data with synthetic images. Hence, object region matching can be improved by favoring 
smaller 2-D motions with sub-optimum PCC over larger (potentially erroneous) displacements with optimum PCC. Moreover,  
this strategy is to be enforced more strongly as iterations progress (towards a more accurate 3-D model). Accordingly, we 
define
\begin{equation}
  \label{eq:motion_socre}
  C_{pc}^r= \frac{C_{pc}}{(1 + r)^{1+\sqrt{d}}}
\end{equation}
\noindent Here, $d$ is the magnitude of  correlation-based 2-D motion $\pmb{v}^I=(\tx-x,\ty-y)$ in $\TI(\tx,\ty)$ (calculated from the beam-bin coordinates $(b,B)$ and $(\tb,\tB)$), and $r$ is a regulator which increases linearly with iteration, capped at 0.8 after 5 iterations; see Fig.~\ref{fig:r_curve}.
The similarity score  $C_{pc}^r$ prioritizes motions with higher PCC primarily in the early stages of optimization, and favors smaller motion magnitudes at later iterations; see Fig.~\ref{fig:score_curve}.  Accordingly, we pick 2-D motions with the highest similarity score.

\subsubsection{2-D Motion Estimation}
The correlation-based registration is suitable within the object region, but not reliable at (or near) 
the frontal contour points (where the correlation window overlaps with the background). Accordingly, we 
define a weighted-average of the 2-D contour and intensity-based motions ($\pmb{v}^C_{m}$  and 
$\pmb{v}^I_{m}$, respectively) for the object-region alignment:
\begin{equation}
  \label{eq:combine_two_motions}
  \pmb{v}_{m} = \alpha\cdot \pmb{v}^C_{m} + (1 - \alpha) \pmb{v}^I_{m}
\end{equation}
\noindent The weight $\alpha$, plotted in Fig.~\ref{fig:alpha_curve}, varies based on distance $d$ from the nearest point of the frontal contour\footnote{with abuse of notation, where $d$ is also used for 2-D correlation-based motion size.}, for which the 2-D motion is calculated by the ICP method. It is set as follows:
\begin{itemize}
  \item $\alpha=1$ for $d \leq d_1=0.5$ [cm]: point moves in harmony with nearest contour pixel;
  \item $\alpha\approx 0$ for $d \ge d_2= 2 [cm]$: object points is far enough from the frontal contour, thus moving according to the correlation-based matching;
  \item $0<\alpha <1$ for $d_1< d < d_2$: decaying exponential weight for the impact of frontal boundary on interior region.
\end{itemize}

\subsection{3-D Patch Motion Estimation}
\label{sec:converting_patch_motions_from_2D_to_3D}
We next present how the registration of synthesized images with the data for various views guides the 3-D motion
of (triangular) patch centers, and consequently the 3-D vertex displacements.

Consider the 2-D image $(\tx_m=\widetilde{\Re}_m\sin\widetilde{\tet}_m,\ty_m=\widetilde{\Re}_m\cos\widetilde{\tet}_m)$ of the patch center $\widetilde{\mmp}^c_{m}=(\tX^c_m, \tY^c_m, \tZ^c_m)$ in $\TI_m$ at pose $\vm_m$, with the 2-D displacement vector $ \pmb{v}_m=(\tx_m-x_m,\ty_m-y_m)$. This yields the match $(x_m,y_m)$ in  $I_m$. 
Applying $\cos\phi\approx 1$ (for $\abs{\phi}\le 7^\circ$), we estimate the 3-D motion $\mv$ of the patch center as follows:

\begin{equation}
\hskip -.12in
  \begin{bmatrix}
  \!    X_{m} \\
  \!  Y_{m} \\
  \!  Z_{m} \!
  \end{bmatrix}\! \!=\! \!
  \begin{bmatrix}
    x_m\cos\phi_m\\
   y_m\cos\phi_m \\
    \Re\sin\phi_m
  \end{bmatrix}\!\!\approx \!\!
  \begin{bmatrix}
   \! x_{m} \\
    \!y_{m} \\
   \! Z_{m}\!
  \end{bmatrix}\,
  \mv_m\!\!=\!\!\begin{bmatrix}
   \! U_{m} \\
   \! V_{m}  \\
   \! W_{m} \!
     \end{bmatrix}\!\!=\!\!
\begin{bmatrix}
   x_{m} \!-\! \tX^c_{m} \\
   y_{m} \!-\! \tY^c_{m} \\
   Z_{m} \!-\! \tZ^c_{m} \\
     \end{bmatrix}
  \label{eq:find_UV}
\end{equation}
\noindent Here, $W_m$ is the only unknown. Transforming $\mv_{m}=[U_{m}, V_{m}, W_{m}]^T$ to the reference pose 
yields $\widehat{\mv}_{c_m}(W_{m}) = [\widehat{U}_{c_m}(W_{m}), \widehat{V}_{c_m}(W_{m}), \widehat{W}_{c_m}(W_{m})]^T$, where dependency on unknown $W_{m}$ is explicitly emphasized. 

Assuming that the patch is visible at all $M$ sonar poses, we collect estimates $\widehat{\pmb{V}}_{c_m}$ for all $M$ views:
\begin{equation}
  \begin{bmatrix}
   \! \widehat{\pmb{V}}^T_{c_1}(W_{1}) \!\\
   \! \widehat{\pmb{V}}^T_{c_2}(W_{2}) \!\\
    \vdots\\
   \! \widehat{\pmb{V}}^T_{c_m}(W_{M}) \!\\
  \end{bmatrix}
  \!\!= \!\!\begin{bmatrix}
    {U}_{1}(W_{1}) & {V}_{1}(W_{1}) & {W}_{1}(W_{1}) \\
    {U}_{2}(W_{2}) & {V}_{2}(W_{2}) & {W}_{2}(W_{2}) \\
    \vdots                & \vdots                & \vdots                \\
    {U}_{M}(W_{M}) & {V}_{M}(W_{M}) & {W}_{M}(W_{M})
  \end{bmatrix}
  \label{eq:motion_matrix}
\end{equation}

Let $\mw = [W_{1},W_{2},...,W_{M}]^T$, and denoting $\mv^{\ast}_c$ as the optimal 3-D motion of the patch-center at the reference pose $\vm_o$ -- for which \eqref{eq:motion_matrix} provides $M$ estimates-- we formulate a least-square eatimation problem:
\begin{equation}
 \mbox{minimize } E_W(\mw) = \sum^{M }_{m=1} \abs{\mv^{\ast}_c-\mv_{c_m}(W_{m})}^2
  \label{eq:sum_square_err}
\end{equation}
\noindent The necessary optimality condition  $\partial E_W/\partial W_m$ yields the linear system 
$\ma_{3M \times M}\mw -\vb_{3M\times 1} = \zr$, and the optimal solution $\mw^{\ast}$
by singular-value decomposition. Finally, inserting into \eqref{eq:motion_matrix} and (median) averaging each column, we derive the 3-D motions $\mv^{\ast}_c$ of the patch center.

Next, recall that we discard some views with large 2-D motion errors. Combined with non-visibility of the patch in 
some views, only $M^r\le M^{-}\le M$ relevant views are available. Maintaining only the corresponding $M^r$ rows in \eqref{eq:motion_matrix}, we solve the remaining over-determined set of equations, using the reduced-size 
$3M^r\times M^r$ matrix $\ma$ and $3M^r\times 1$ vector $\vb$. In the (unlikely) degenerate scenario, where
$M^r\le 1$, we assign zero motion to the patch center. 

\subsection{Determining Vertex Displacements}
\label{sec:determining_vertex_motions}

Having calculated the 3-D displacements of $N_T$ patch centers, we seek the motions of $N_P$ vertices. 
As depicted in example of Fig. 4b, each vertex generally lies on two or more patches with different motions,
calling for a least-square estimation formulation, as described next.

A 3-D model with $N_T$ triangles and $N_P$ vertices has an $N_T\times N_P$ connectivity matrix $\mc$ with rows $\vc_i$ 
and entries $c_{ij} = 1$ if patch $i$ contains vertex $j$, and $c_{ij} = 0$, otherwise. For the example in Fig.~4b, 
the connectivity matrix is given by

\begin{equation}
  \mc = \begin{bmatrix}
    &1& 1& 0& 0& 1\\
    &1& 0& 0& 1& 1\\
    &0& 0& 1& 1& 1\\
    &0& 1& 1& 0& 1\\
  \end{bmatrix}
\end{equation}

\noindent Defining the $N_P\times 3$ and $N_T\times 3$ matrices of vertex and optimal patch center motions $\mv_P$ and
 $\mv^{\ast}_c$, respectively, we pose the following least-square problem for optimum vertex motion  $\mv^{\ast}_P$:
\begin{equation}
\hskip -.12in
   \mv^{\ast}_P \!\!=\!\! \arg\min_{\mv_P} \!\sum_{i = 1}^{N_T}  \abs {\vc_i\mv_P - \mv^{\ast}_{c_i}}^2 \;
  \! \mv_P\!\!=\!\!\begin{bmatrix}
   \!  \mv^T_{P_1} \!\\
   \!  \mv^T_{P_2} \!\\
    \vdots \\
   \! \mv^T_{P_{N_P}} \!\\
             \end{bmatrix}\,
  \mv_c^{\ast}\!\!=\!\!\begin{bmatrix}
  \! \mv_{c_1}^{\ast T}\!\\
  \! \mv_{c_2}^{\ast T}\!\\
    \vdots \\
  \!  \mv_{c_{N_T}}^{\ast T}\!
              \end{bmatrix}
\end{equation}
The solution $ \mv^{\ast}_P=(\mc^T\mc)^{-1}\mc^T\mv^{\ast}_c$ involves the pseudo-inverse of $\mc$. 
To avoid undesired 3-D model distortions by outliers, we replace motions with magnitudes exceeding three median-absolute-deviations (MAD) above the median by the weighted average of neighboring vertices.
For example, if $\mv_{\!\!P_2}$ in Fig.~4b 
is identified as an outlier, we  replaced by $(d_{21}\mv_{\!\! P_1} + d_{25}\mv_{\!\! P_5} + d_{23}\mv_{\!\! P_3})/(d_{21} +d_{25}+d_{23})$; $d_{ij}$ is the distance from $\mmp_i$ to $\mmp_j$. Finally, the 3-D model at iteration $k$ is updated by calculating the new vertex positions 
$(\mmp_{P_l})^{k+1}=(\mmp_{P_l})^{k}+(\mv^\ast_{P_l})^{k}$. 

\subsection{Error Metric}
\label{sec:volumetric_error}

Let $\mathcal{\TI}^t = \{\TI^t_m~|~m=1,2,...,M^r\}$ denote the set of relevant synthetic beam-bin images generated at iteration $t$, comprising of pixel set $\varOmega^t_m$ within the uncorrupted object region of $\TI^t_m$. We define the intensity error (IE) of $\TI^t_m$, and
average intensity error (AIE) over the whole synthetic image set:
\begin{equation}
\begin{array}{rl}
     \textrm{IE}(\TI^t_m) &\hskip-.1in = \frac{\displaystyle 1}{\displaystyle \abs{\varOmega_m^t}}\sum_{(b,B)\in \varOmega_m^t} \abs{\TI^t_m(b,B) - I^t(b,B)} \\
   \textrm{AIE}(\mathcal{\TI}^t)&\hskip-.1in  = \frac{\displaystyle 1}{\displaystyle M^r}\sum_{m=1}^{M^r} \textrm{IE}(\TI^t_m)
   \end{array} \\
\end{equation}
\noindent where $\abs{\varOmega^t_m}$ denotes the cardinality of set  $\varOmega^t_m$. Similarly, we define the contour error (CE) according to (9), and average contour alignment error (ACE) as follows:
\begin{equation}
  \textrm{CE}(\TI^t_m) =  \lambda (\mathcal{\widetilde{C}}^{\prime}_m, {\mathcal{C}}_m)\,\quad
  \textrm{ACE}(\mathcal{\TI}^t) = \frac{\displaystyle 1}{\displaystyle M^r}\sum_{m=1}^{M^r} \textrm{CE}(\TI^t_m)
\end{equation}

To assess the iterative 3-D model adjustment, we normalize each error measure by the value for the initial 3-D model $S^o$ and calculate the image error $E_I(t)$:
\begin{equation}\label{tot-error}
\hskip -.07in
\small{
    \textrm{NAIE}(t) \!\!=\!\! \frac{\displaystyle \textrm{AIE}(\mathcal{\TI}^t)}{\displaystyle \textrm{AIE}(\mathcal{\TI}^0)} \,
    \textrm{NACE}(t) \!\!=\!\! \frac{\displaystyle\textrm{AIE}(\mathcal{\TI}^t)}{\displaystyle \textrm{AIE}(\mathcal{\TI}^0)}\, 
  E_I(t) \!\!=\!\!  \frac{\displaystyle 1}{\displaystyle 2}(\textrm{NAIE}(t) \!\!+\!\! \textrm{NACE}(t))
  }
\end{equation}
{It is informative to validate that reducing $E_I$ is directly correlated with improving the 3-D object model accuracy.  To assess model accuracy, we define the normalized volumetric error (NVE) [2]:
\begin{equation}
0\le  E_{\Sigma} = \frac{\widetilde{\mathbf{\Sigma}} + \mathbf{\Sigma}^c - 2\mathbf{\Sigma}_{\tS\cap S^c } }{\widetilde{\mathbf{\Sigma}} +\mathbf{\Sigma}^c -\mathbf{\Sigma}_{\tS\cap S^c }} \quad \mathbf{\Sigma}_{\tS\cap S^c }= \mathbf{\Sigma}(\tS\cap S^c)
  \label{eq:volume_error}
\end{equation}
where $\mathbf{\Sigma}^c=\mathbf{\Sigma}(S^c)$ and $\widetilde{\mathbf{\Sigma}}=\mathbf{\Sigma}(\tS)$ denote the volumes of true and estimated models $S^c$ and $\tS$, respectively. The NVE represents the total non-common volume as a fraction of the total volume. The lower and upper bounds correspond to two identical and disjoint volumes, respectively. To reiterate, we measure $E_{\Sigma}$ primarily to examine the correlation with $E_I$, i.e. whether $E_I$ (which we can calculate) reflects the accuracy of the estimated 3-D model.}

{In the absence of true 3-D model $\mathbf{S}^c$ for each coral object in our experiments, we optimize (as we explain next) the 3-D model ${S}^K$ from a Kinect camera, generated in air at a distance of about 1 [m]. We utilize the same model for the experiments with synthetic data.}

\section{Experiments}\label{experiments}
{Aside from impact of noise, the inaccuracies of either the image formation model or 3-D Kinect model $S^K$  can lead to the deviations between the synthesized images (of Kinect model) and the real data.  Having verified the image formation model with the real images of dominantly convex objects \cite{aykin2013forward}, we explore if minimizing the image error over all views of real data requires distorting the Kinect model into an ``optimized model'' $S^{\tilde{K}}$. Here, quantifying the degree/percentage by which $S^{\tilde{K}}$ deviates from $S^K$ also becomes informative. Moreover, we can assess the optimization performance based on SC model initialization: whether the minimum error is achieved, and any discrepancy between the optimized SC and Kinect models. Thus, the normalization/comparison of the volumetric error relative to the optimized Kinect model $S^{\tilde{K}}$, instead of initial Kinect model $S^{K}$, is deemed to be more informative.}

We have introduced a number of parameters, settings, and thresholds in various sections, all of which are set to fixed values as given.  In particular, we have determined their suitable values and their role in the performance of different components of the optimization scheme through extensive simulations with synthetic data \cite{thesis}. Moreover, we have verified consistent performance with the results from experiments with real data. 

The real experiments in this work involve two data sets, comprised of eight rotated views at each of two opposite (N and S) sonar positions in water depth of about 43 [cm] with sonar tilt of $\beta\approx 10\degree$. We also use the second set in an experiment with mixed (real and synthetic) data to assess the improved reconstruction using larger object coverage. Here, the computer-generated data comprise of the missing rotated views at E and W positions, by utilizing the optimized Kinect models of coral objects (under the same conditions as for the real data), while initializing the 3-D model with the (less-accurate) SC solution. {A final experiment is aimed at exploring the impact of non-flat air-water interface. Here, we analyze variations in mirror image contour induced by surface waves in a video sequence. Accordingly, we generate synthetic data with water surface fluctuations yielding mirror contour variations at the same scale as in the real images, allowing the assessment of impact on model reconstruction accuracy based on ground-truth data.}

\begin{figure*}[th!]
  \centering
    \vskip -.15in
      \begin{tabular}{ccccc}
         \hskip -.18in   \includegraphics[height=1.46in]  {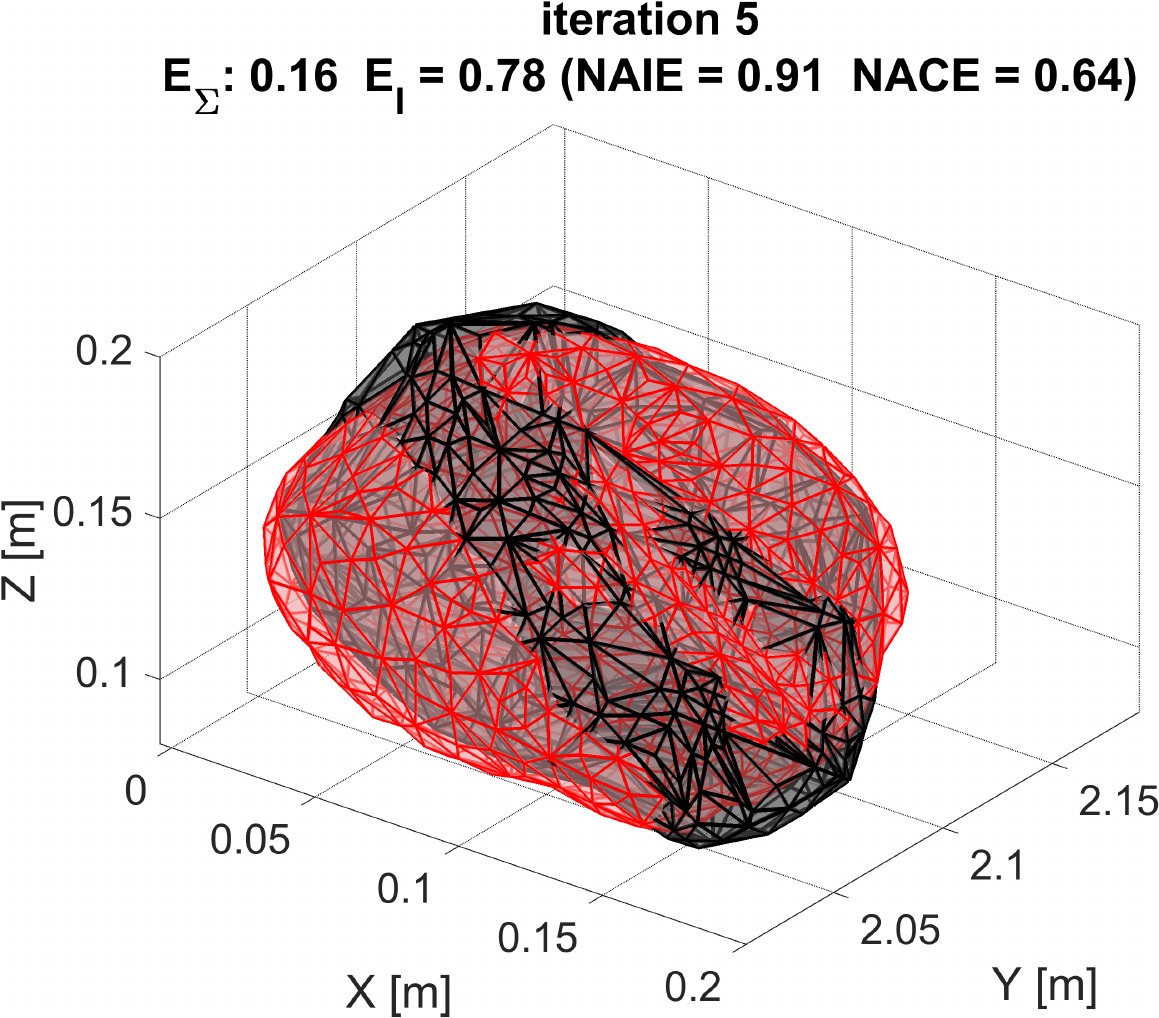} &
      \raisebox{18ex}{\hskip -3.25in(a)} &
         \hskip -.18in   \includegraphics[height=1.24in]    {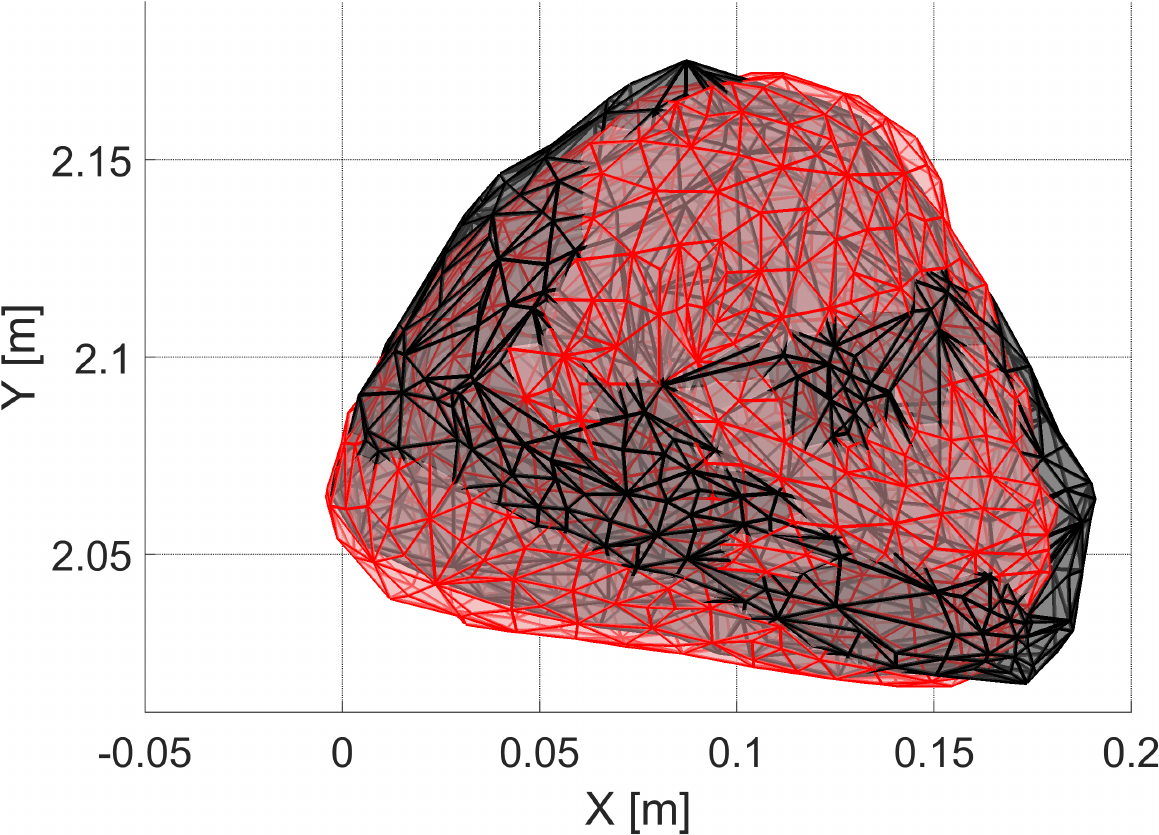} & 
         \hskip -1.68in  \includegraphics[height=1.2in]     {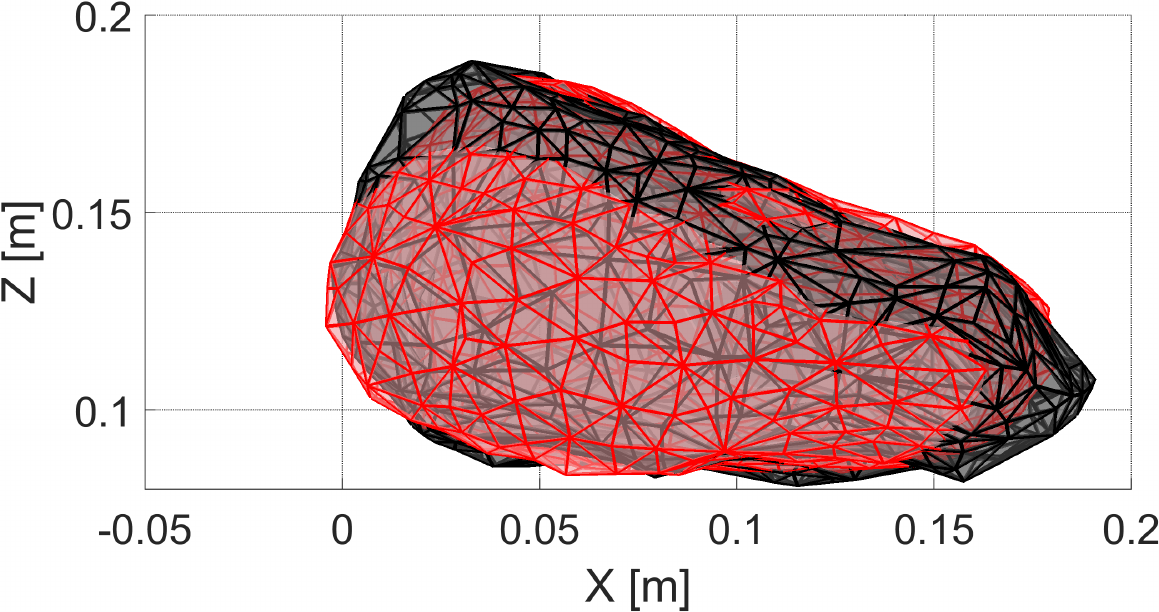} & 
          \hskip -1.78in  \includegraphics[height=1.18in]  {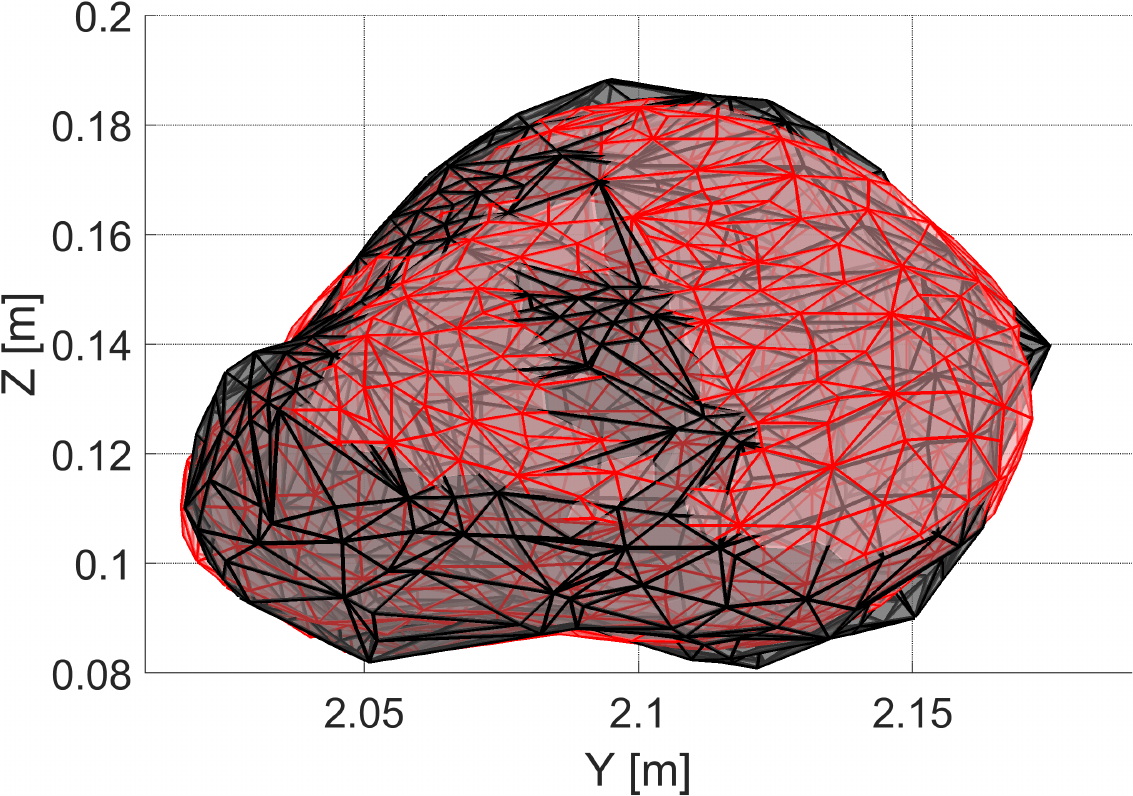} \\
         
          \hskip -.18in  \includegraphics[height=1.5in]  {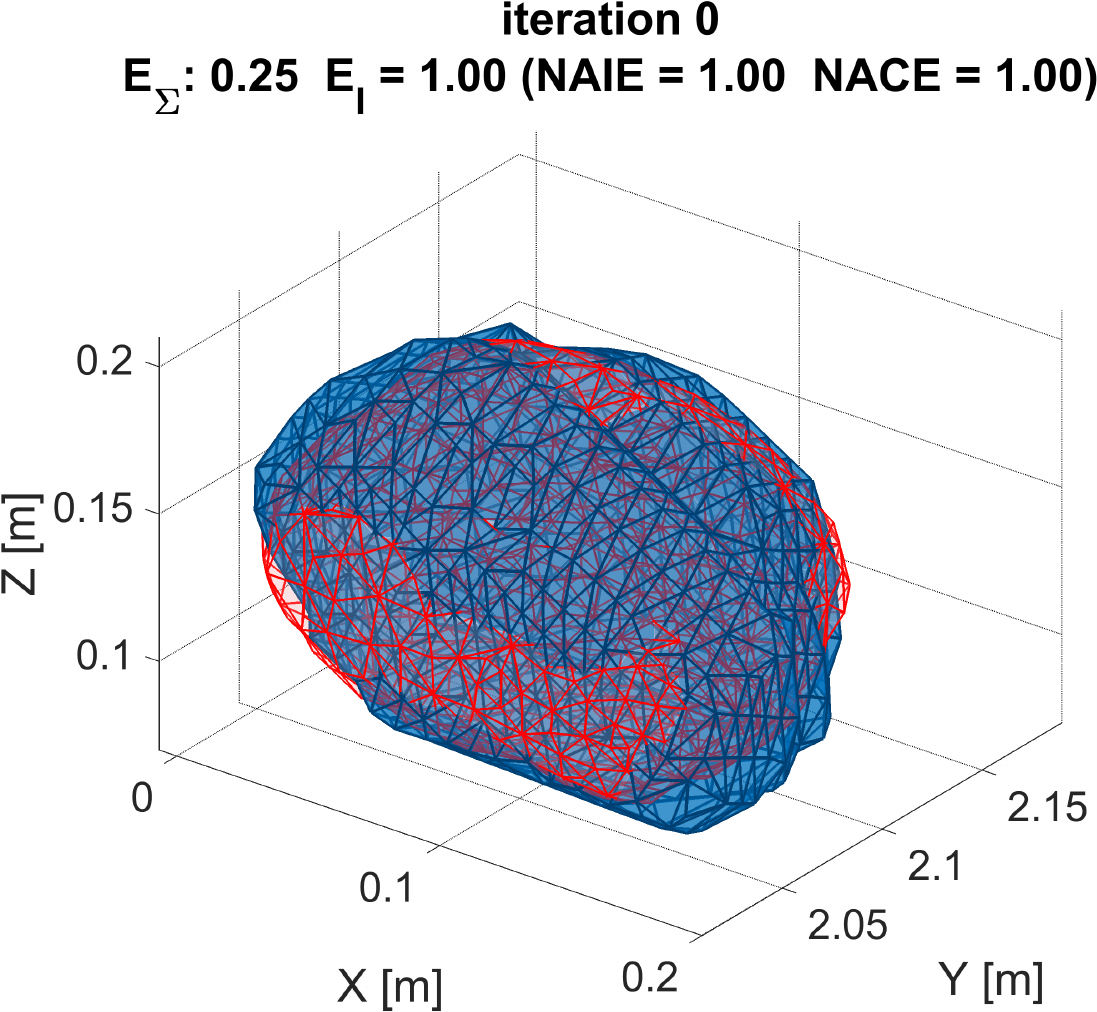} &
      \raisebox{18ex}{\hskip -3.25in(b)} &
          \hskip -.18in  \includegraphics[height=1.24in]    {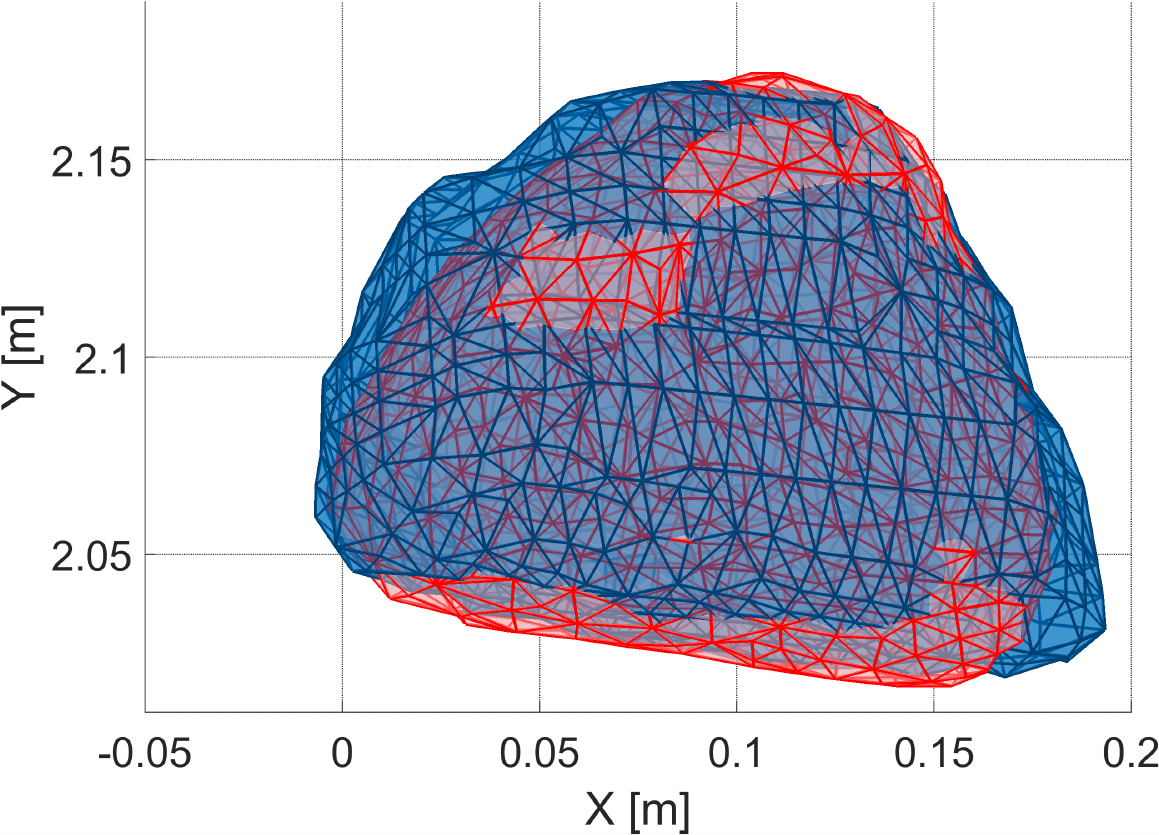} &
          \hskip -1.78in  \includegraphics[height=1.26in]  {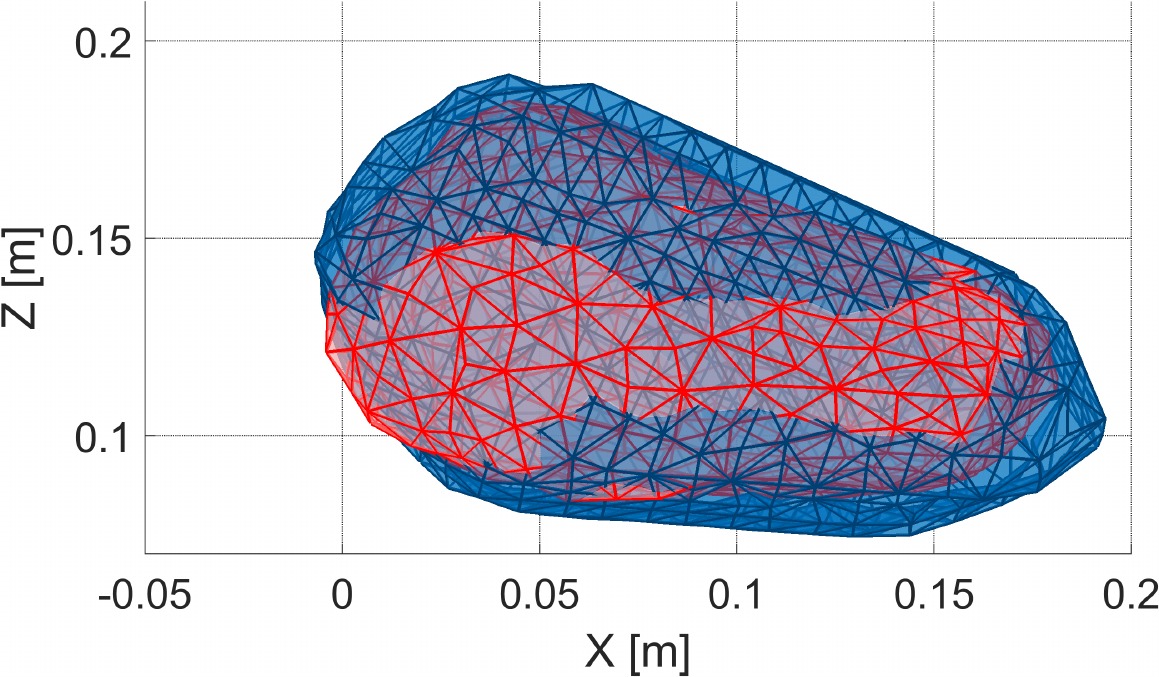} & 
          \hskip -1.78in  \includegraphics[height=1.24in]  {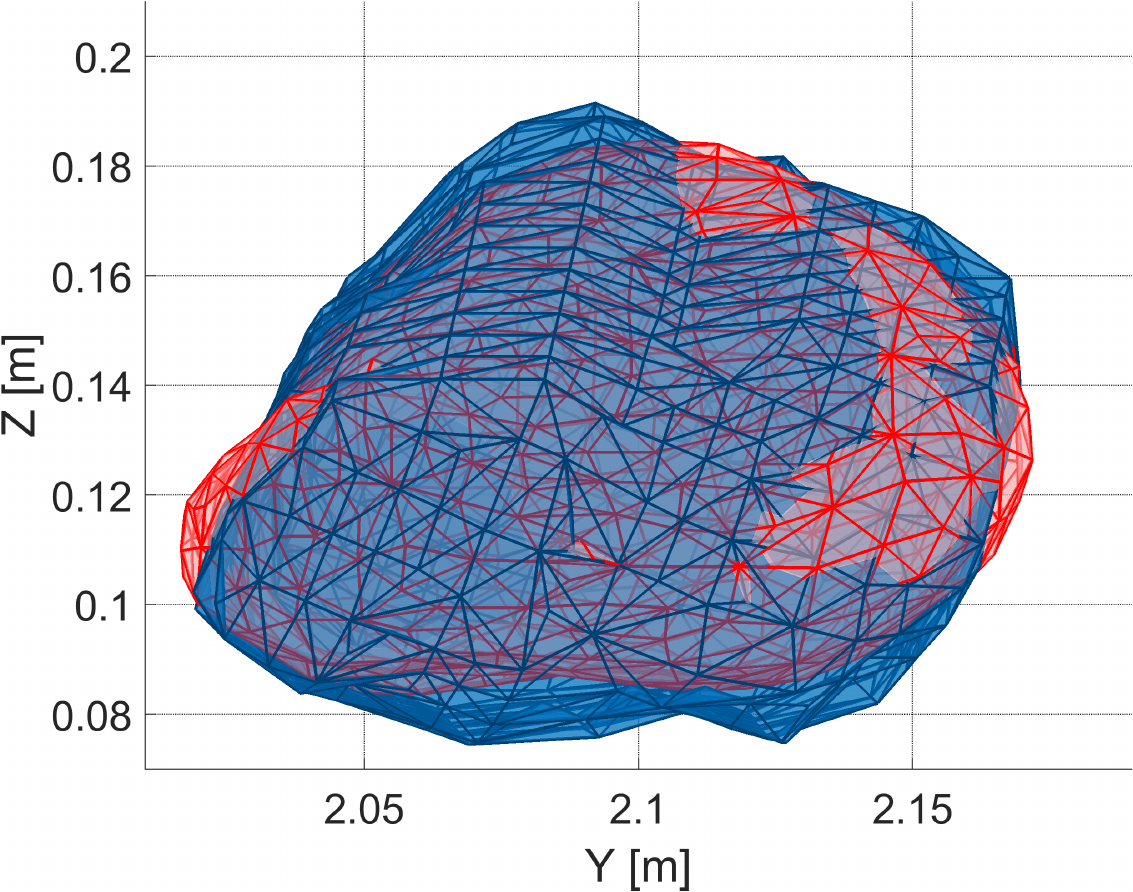} \\
         
          \hskip -.18in \includegraphics[height=1.5in]   {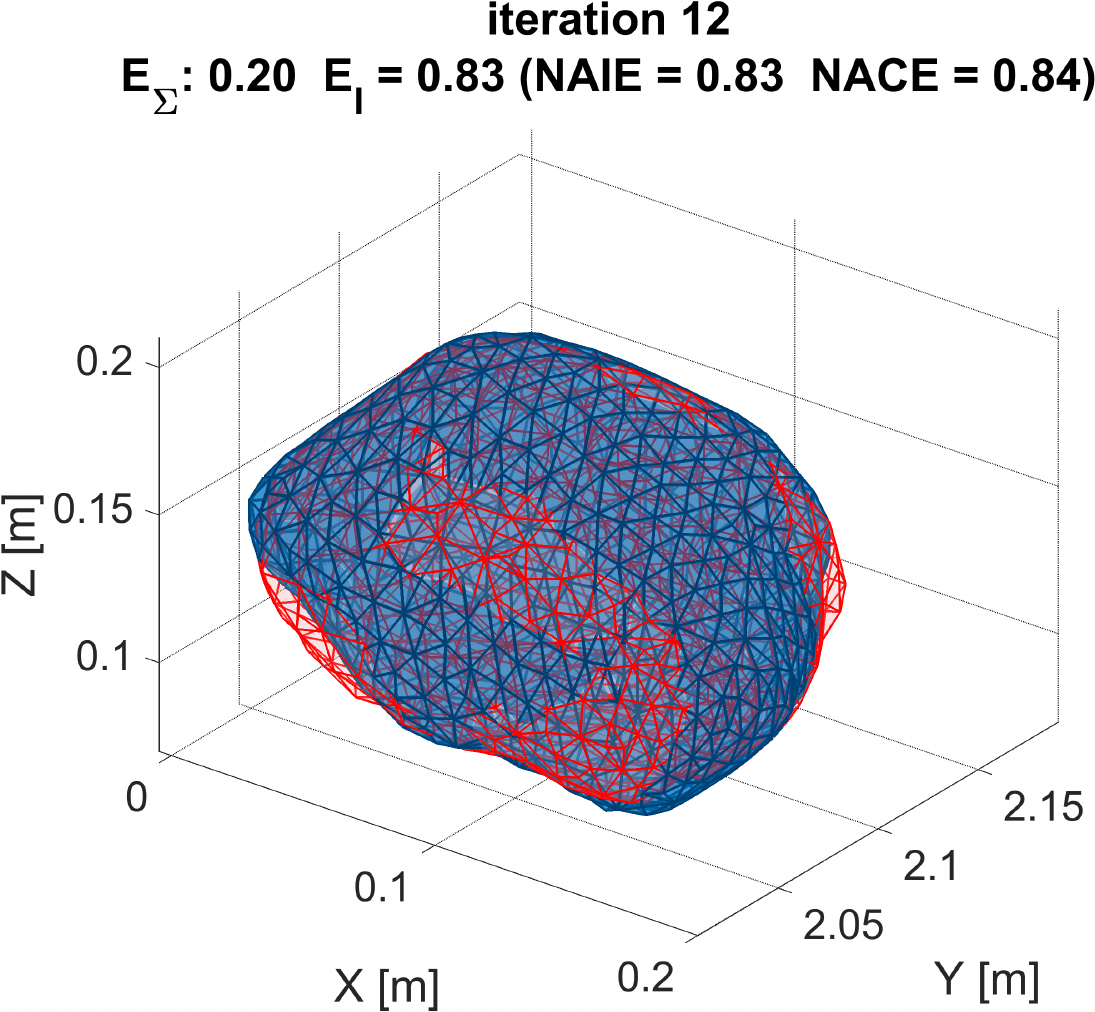} &        
      \raisebox{18ex}{\hskip -3.25in(c)} &
          \hskip -.18in  \includegraphics[height=1.24in]   {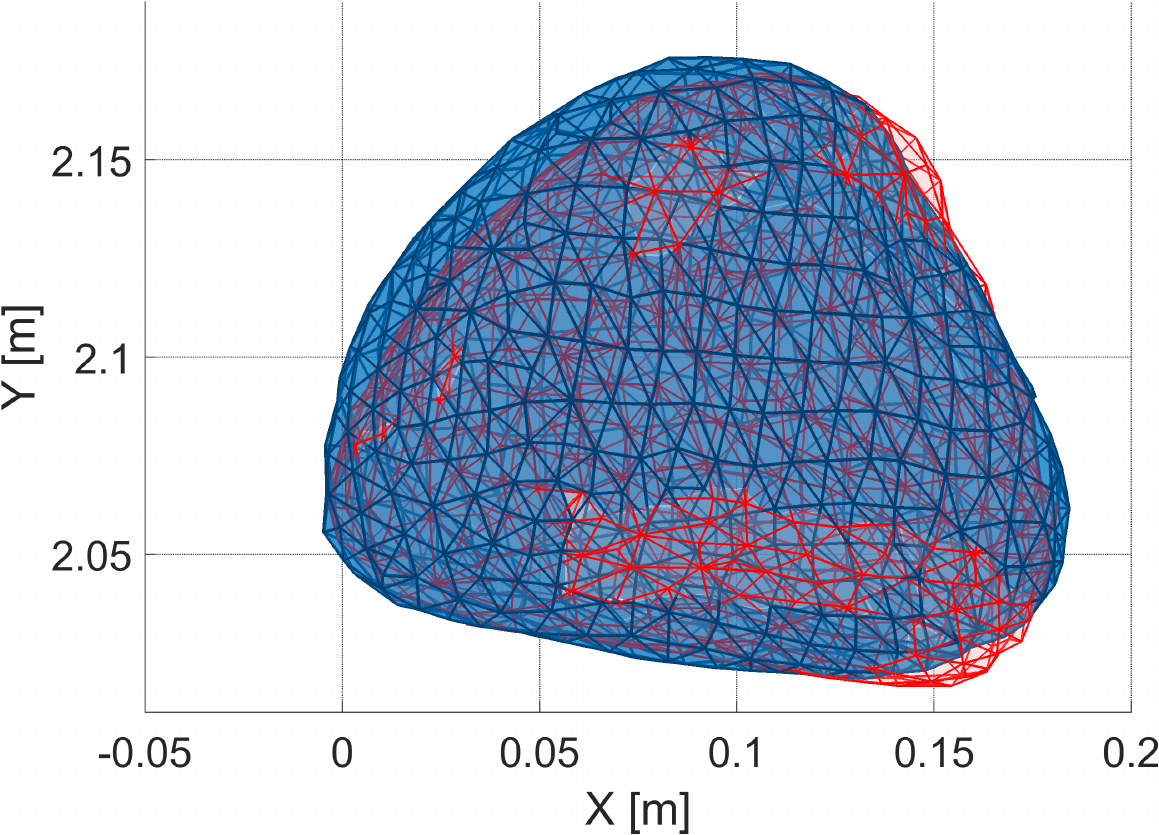}  &
          \hskip -1.68in  \includegraphics[height=1.26in]  {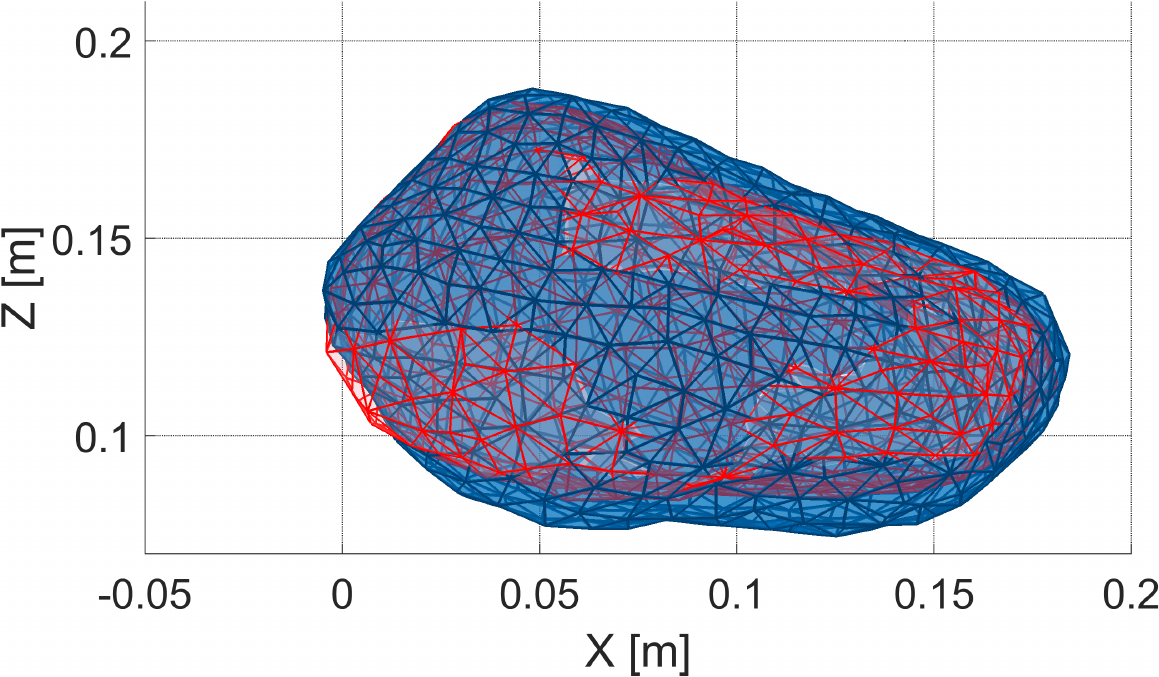} &     
          \hskip -1.78in  \includegraphics[height=1.24in]  {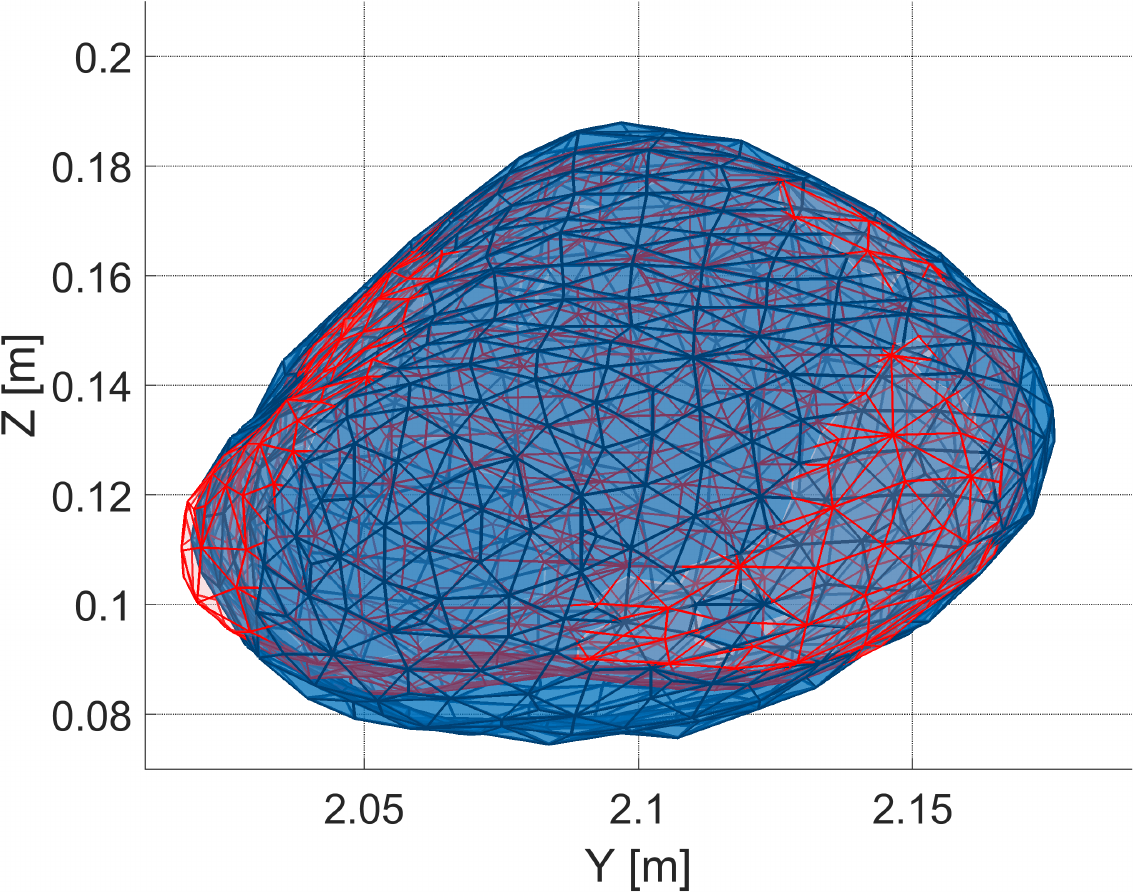} \\
         
          \hskip -.18in  \includegraphics[height=1.5in]  {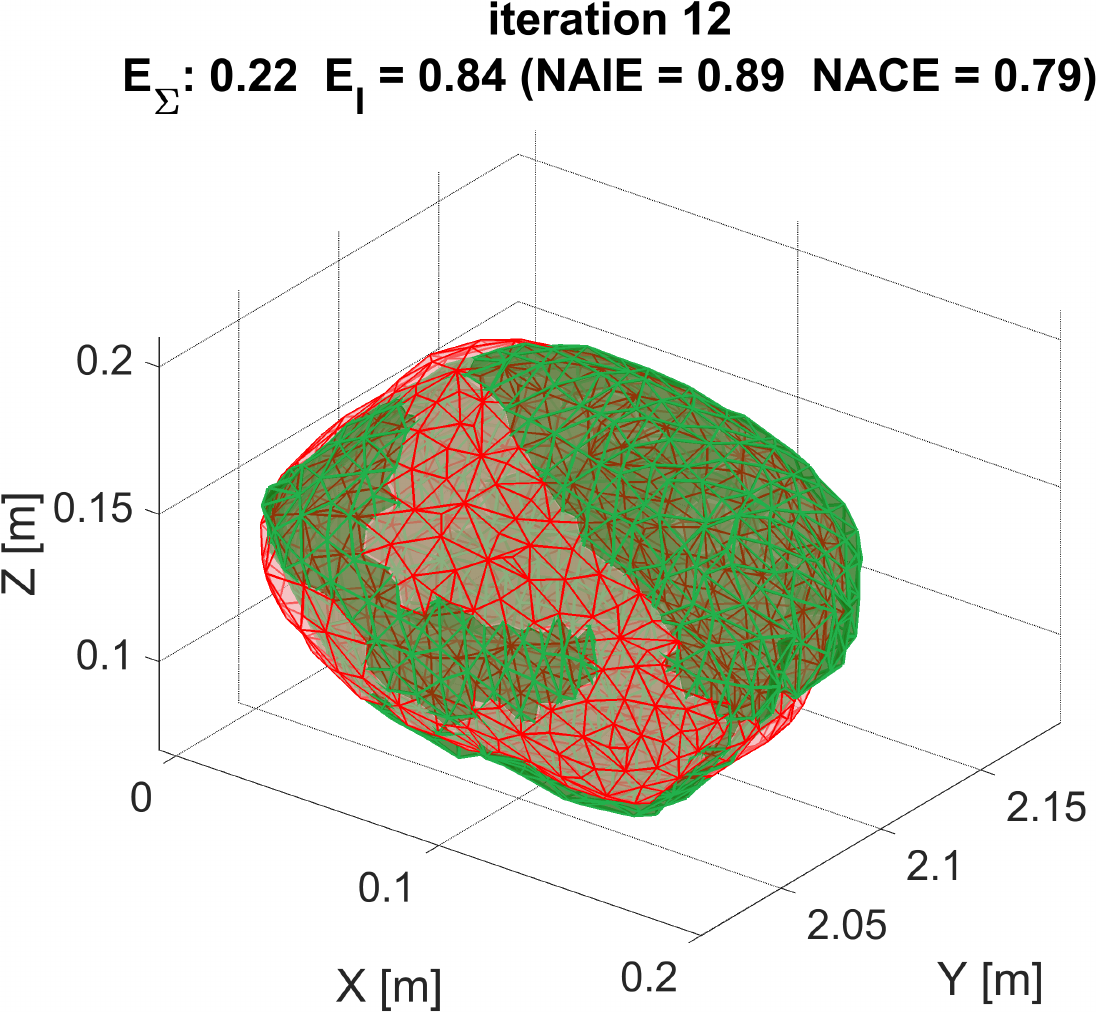} &       
      \raisebox{18ex}{\hskip -3.25in(d)} &
          \hskip -.18in  \includegraphics[height=1.24in]  {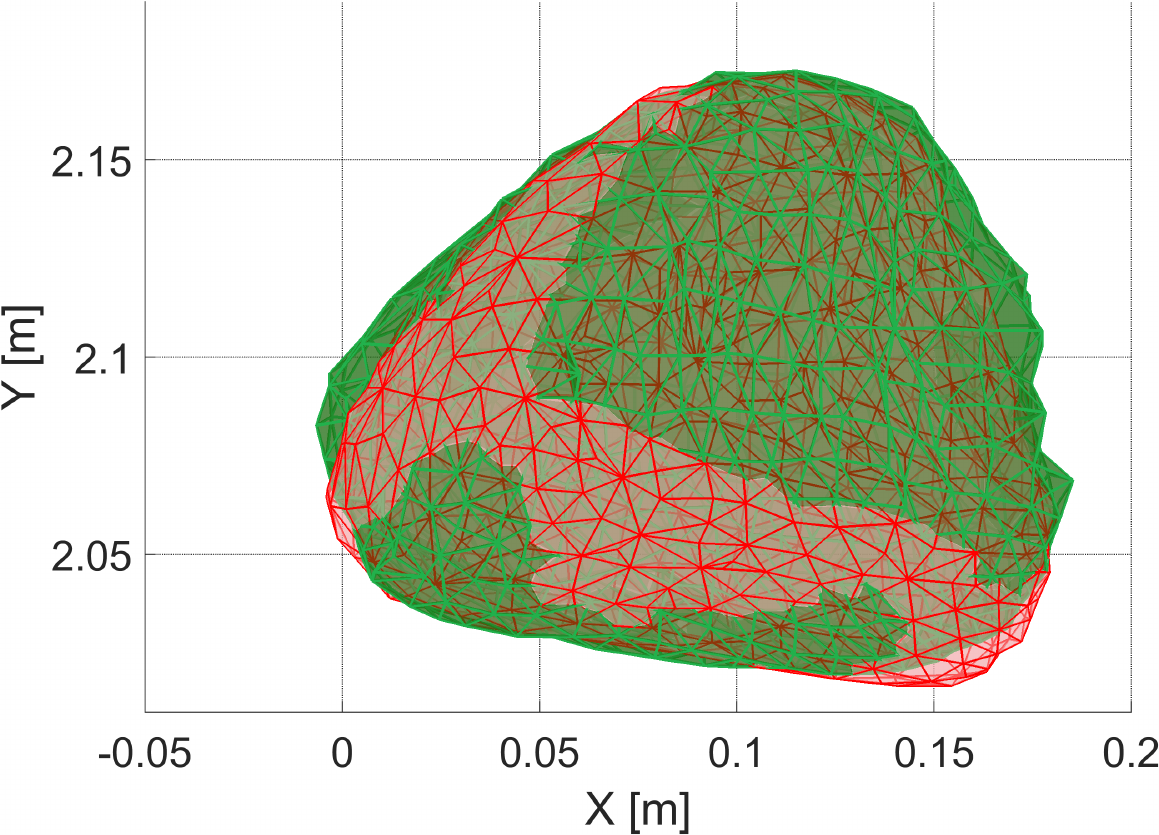} &             
          \hskip -.18in  \includegraphics[height=1.3in]  {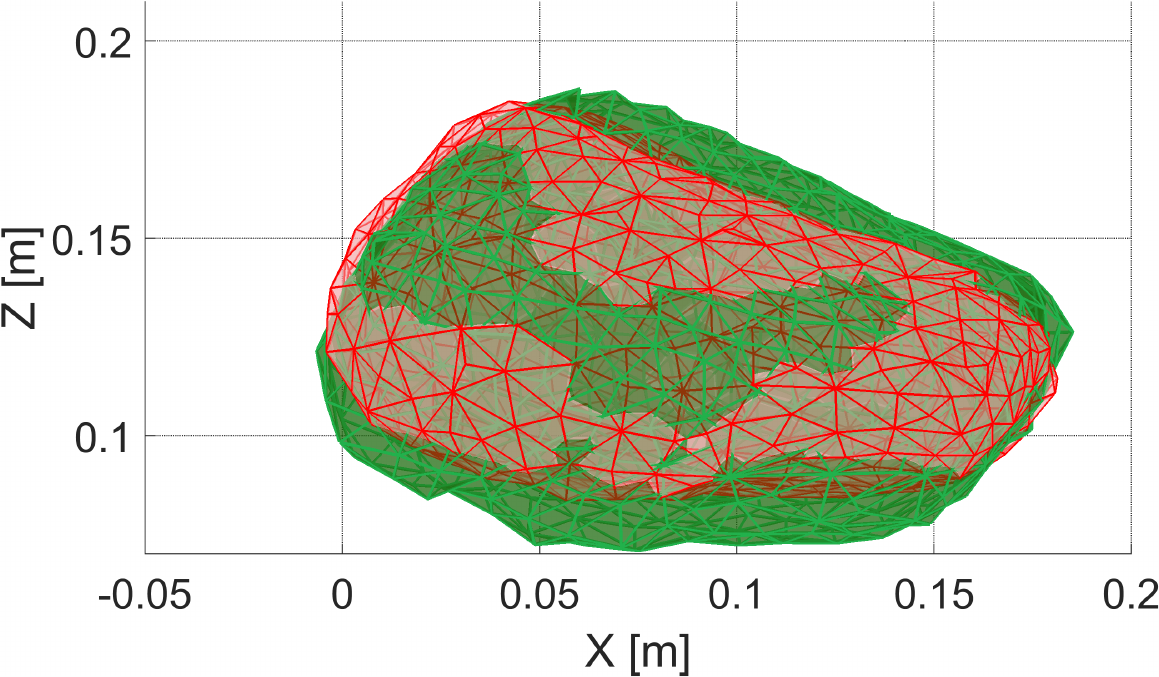} \&     
         \hskip -.18in  \includegraphics[height=1.24in]  {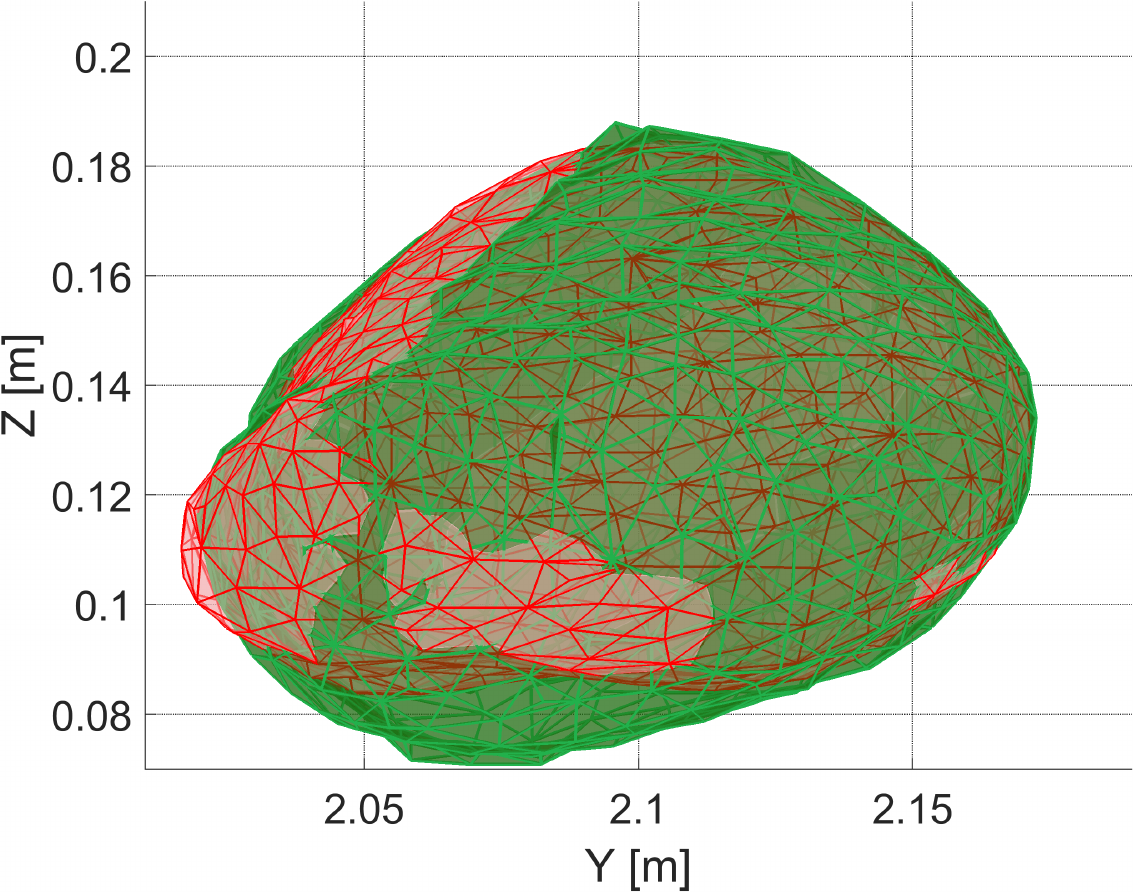} \\ 
     \end{tabular}

      \begin{tabular}{cccccc}
              \hskip -.3in
               \raisebox{12ex}{(e)}  & \hskip -.05in
     \includegraphics[height=1.45in]  {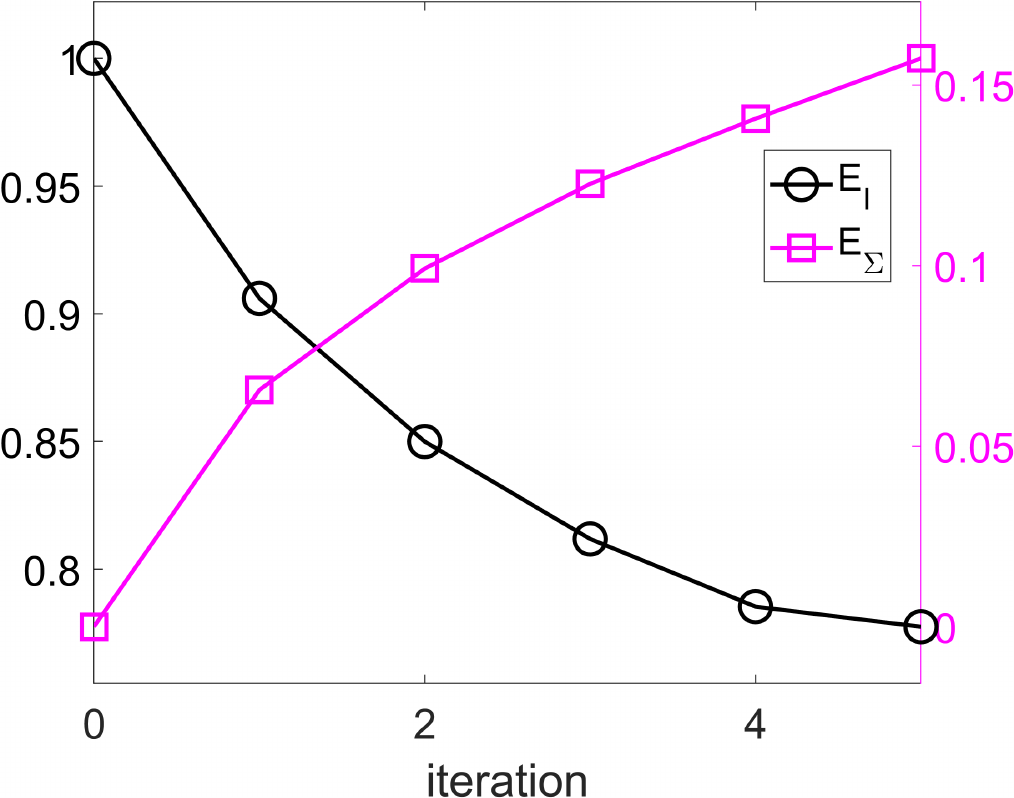} &
               \raisebox{12ex}{(f)}  & \hskip -.05in
     \includegraphics[height=1.45in] {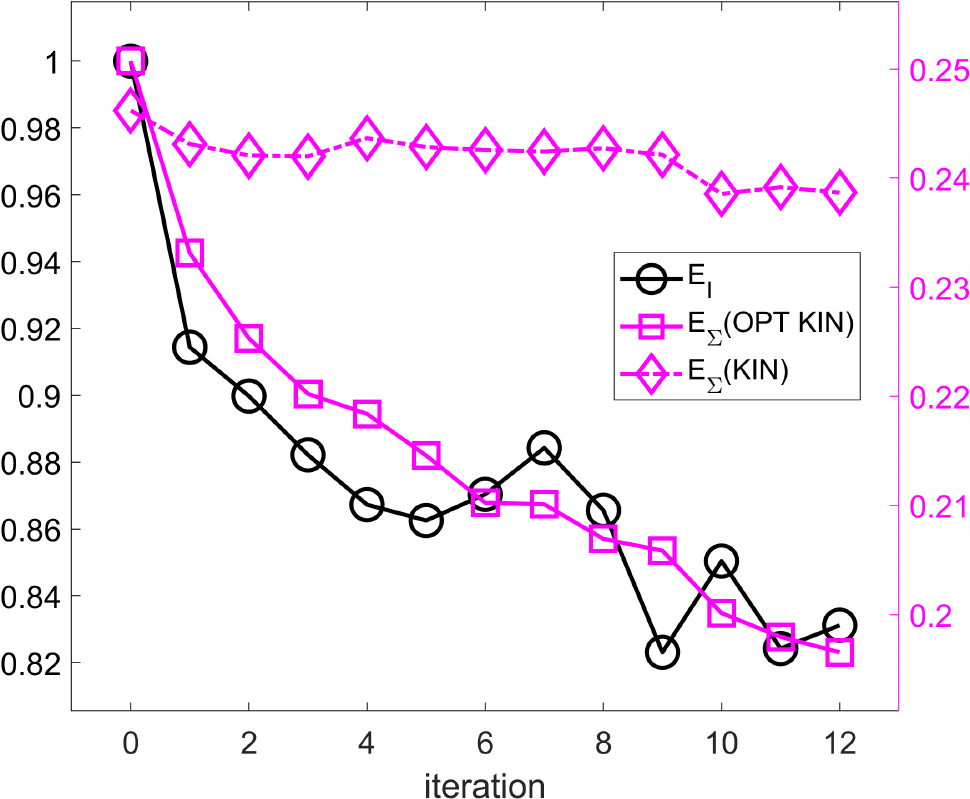}  &
               \raisebox{12ex}{(g)}  & \hskip -.05in
     \includegraphics[height=1.45in] {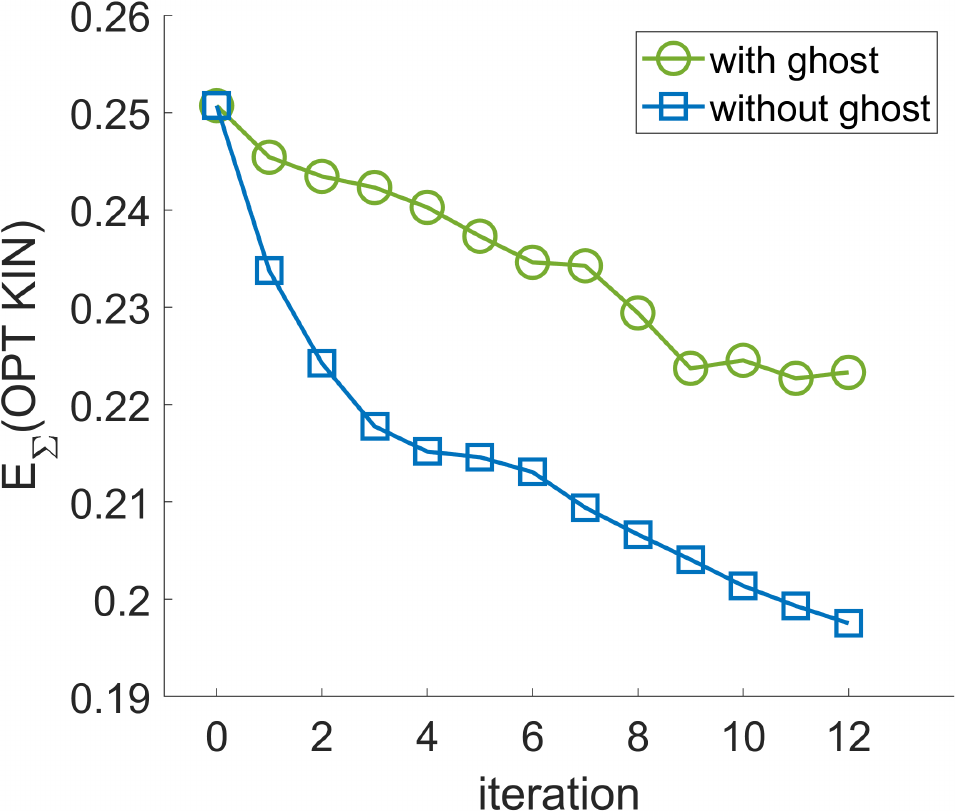}  \\
     \end{tabular}           
      \vskip -.1in
 \caption{Coral-one Experiment-- (a) Shape adjustment from original (black) to optimized (red) Kinect model. 
 (b) initial SC (blue) and optimized Kinect model (red), (c) optimized SC (blue) and  Kinect (red) models, and (d)  optimized Kinect model  (red) and optimized SC without removing ghost region from data (green). (e) Normalized image error $E_I$ of optimized Kinect model (red) decreases by roughly 20\% through 15\% volumetric variation w.r.t. original Kinect model 
 (f) Optimized SC solution improves image error by about 18\% (black circles) and volumetric error (magenta squares) from 25\% to less than 20\% (relative to optimized Kinect model), but adjusts negligibly from 25\% to 24\%, relative to original Kinect model (magenta diamonds). 
 (g) Volumetric error is higher when optimizing without ghost-region removal (green circles).   \label{fig:Coral1_SC_rd}}

\vskip -.15in
\end{figure*}

\begin{figure*}[th!]
  \centering
\begin{tabular}{cc}
\hskip -.2in
      {\begin{tabular}{ccccc}
         \raisebox{12ex} {\hskip -.2in(a)}  &
          \hskip -.2in    \includegraphics[height=1.785in]  {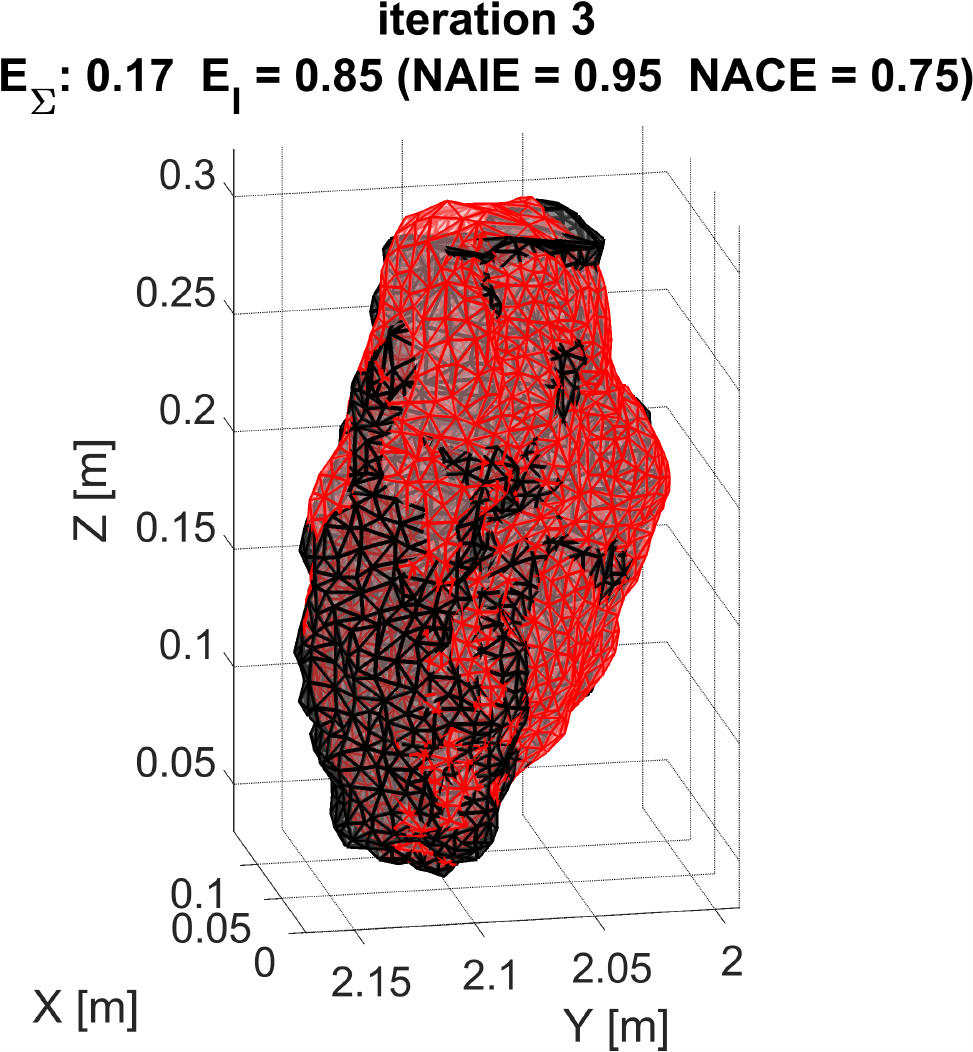}  & 
         \hskip -.6in.    \includegraphics[height=1.47in]  {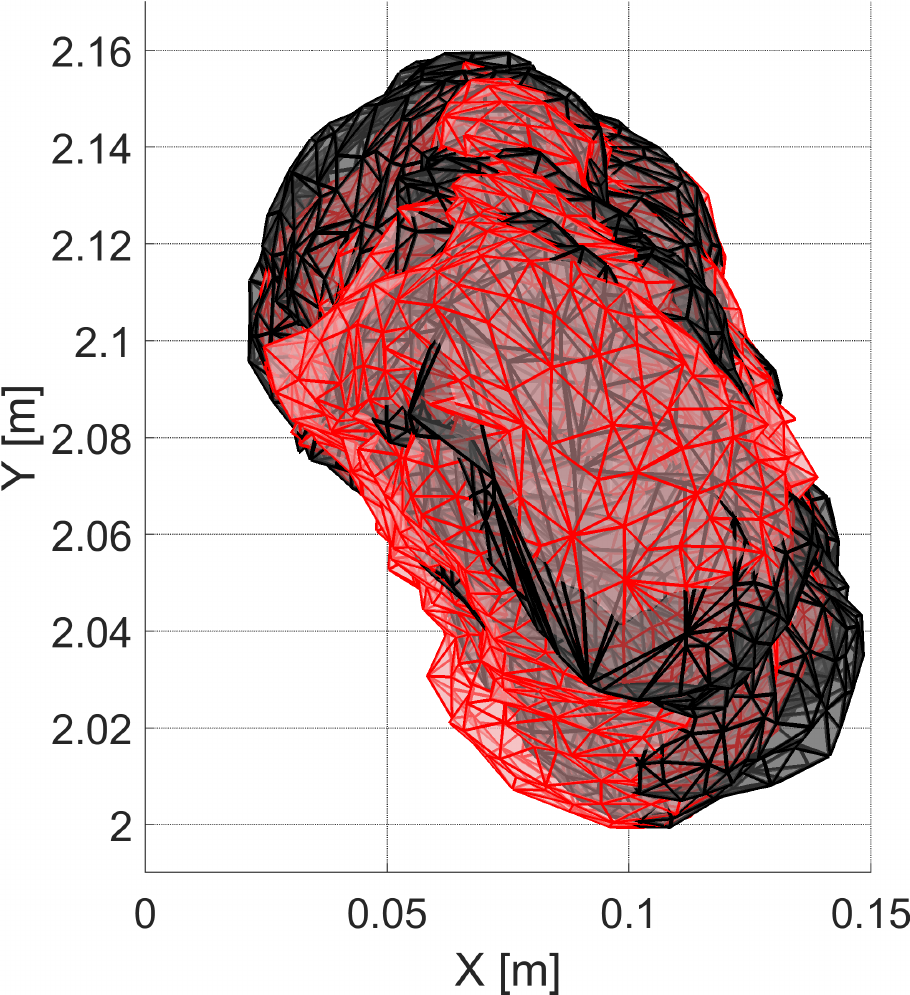} & 
         \hskip -.2in     \includegraphics[height=1.47in]  {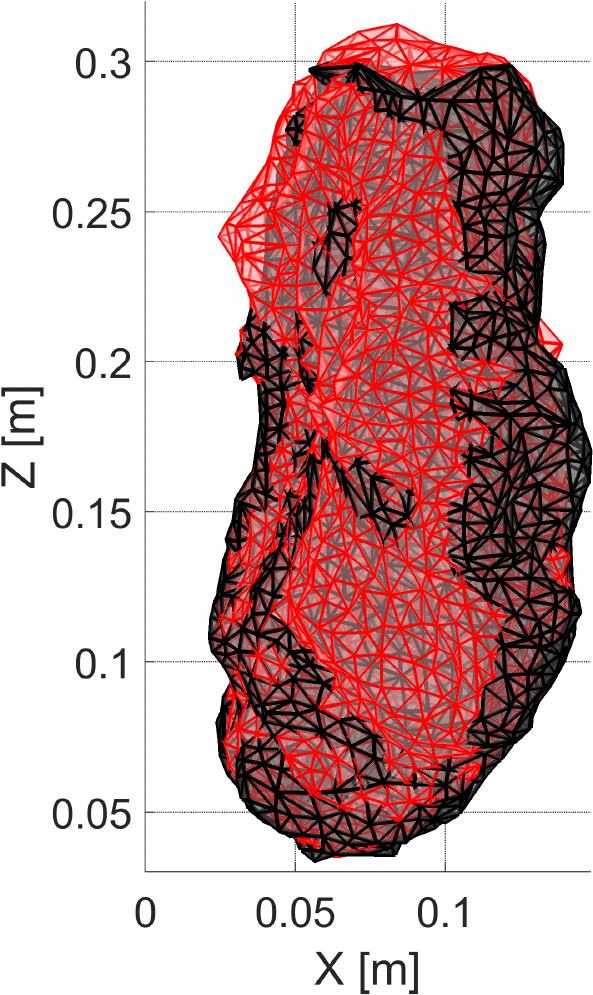} & 
         \hskip -.2in     \includegraphics[height=1.47in]  {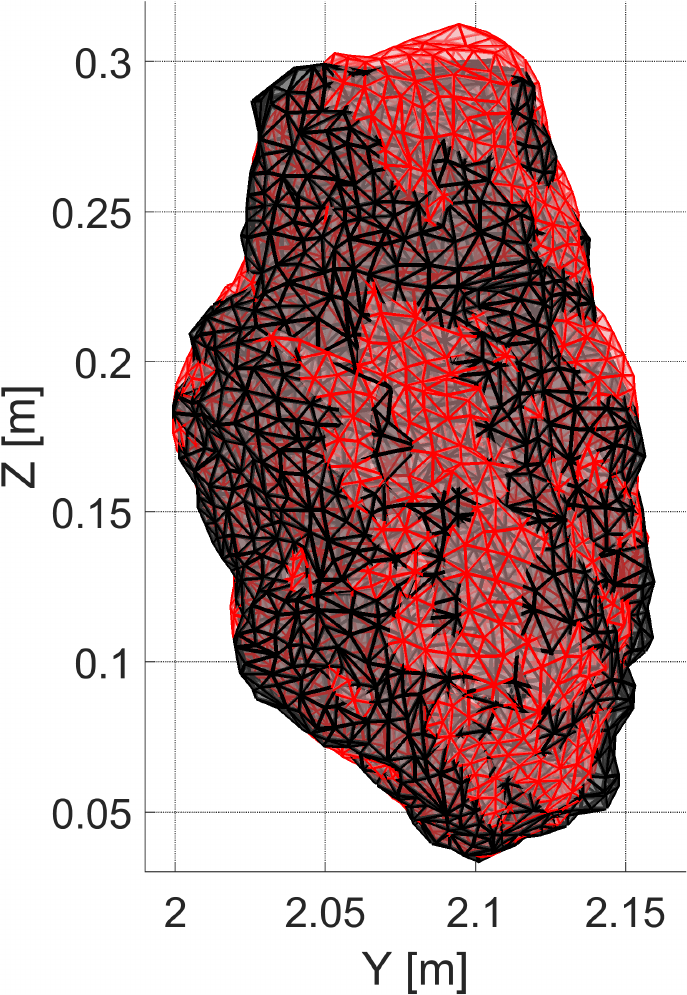} \\\ 
         \raisebox{12ex} {\hskip -.2in(b)} &
         \hskip -.2in    \includegraphics[height=1.785in]  {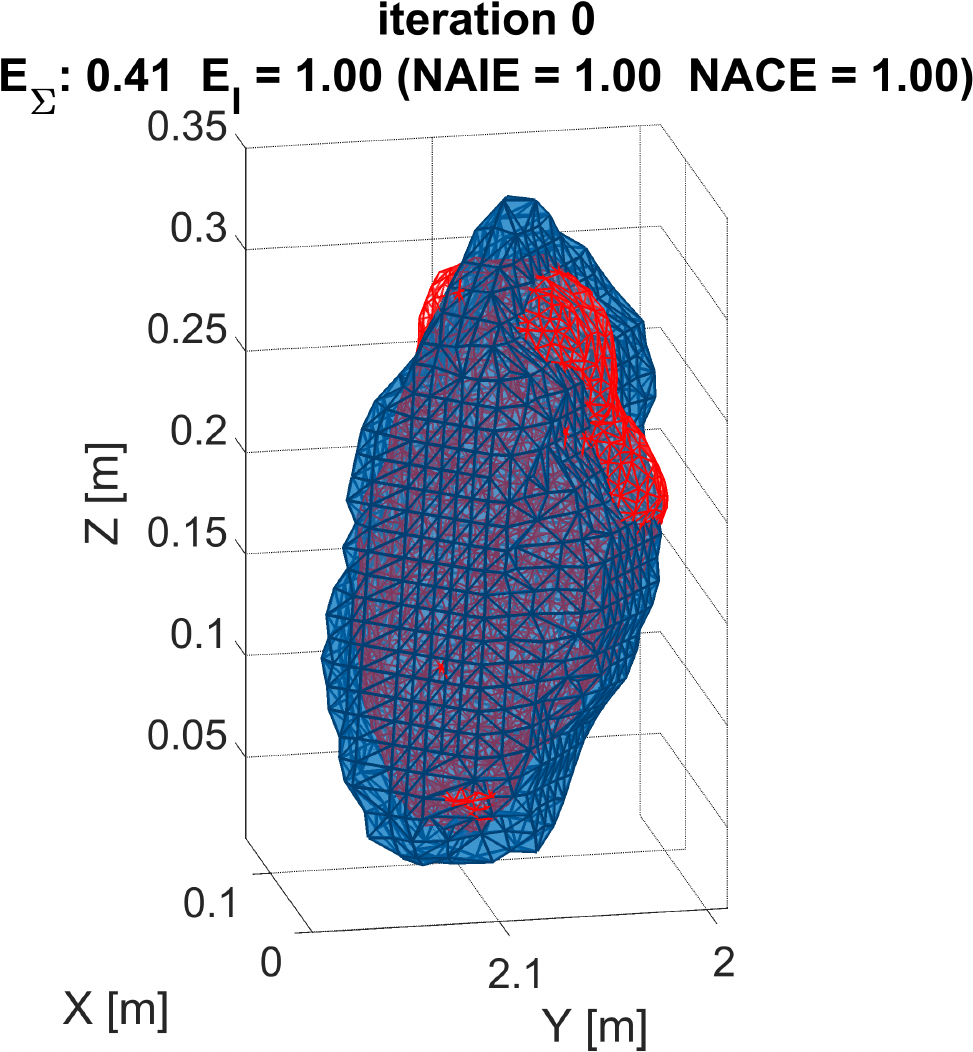}  & 
        \hskip -.5in.     \includegraphics[height=1.47in]  {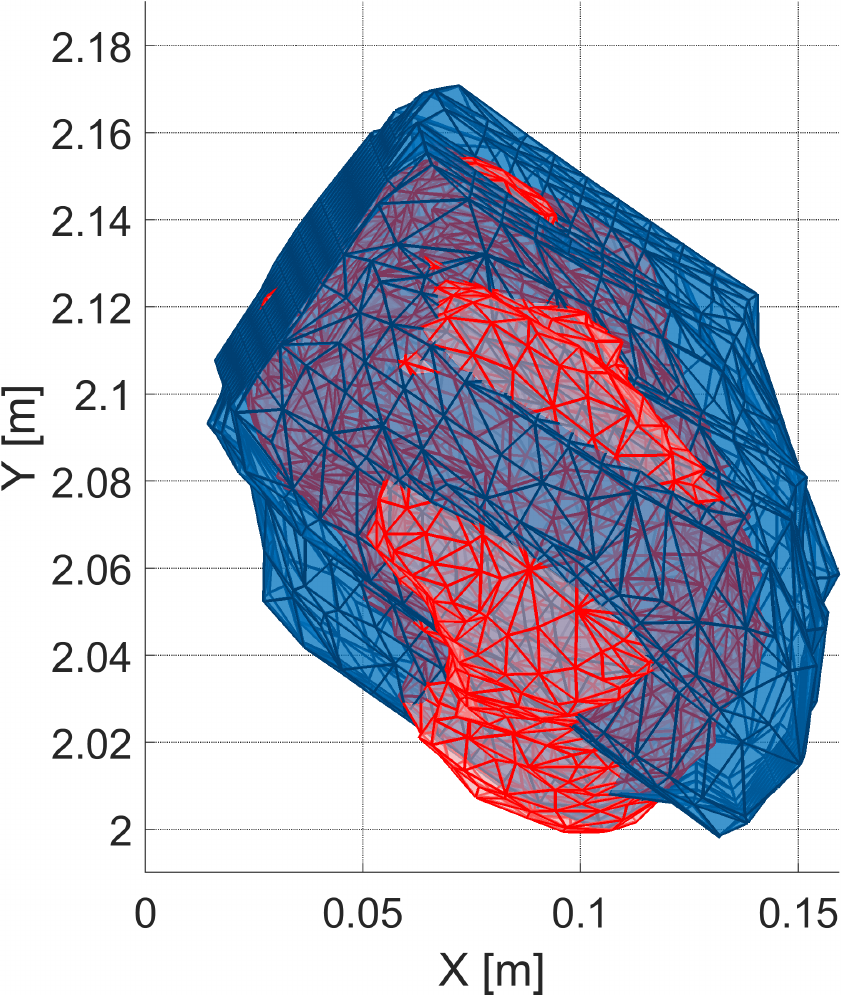} &
        \hskip -.1in      \includegraphics[height=1.47in]  {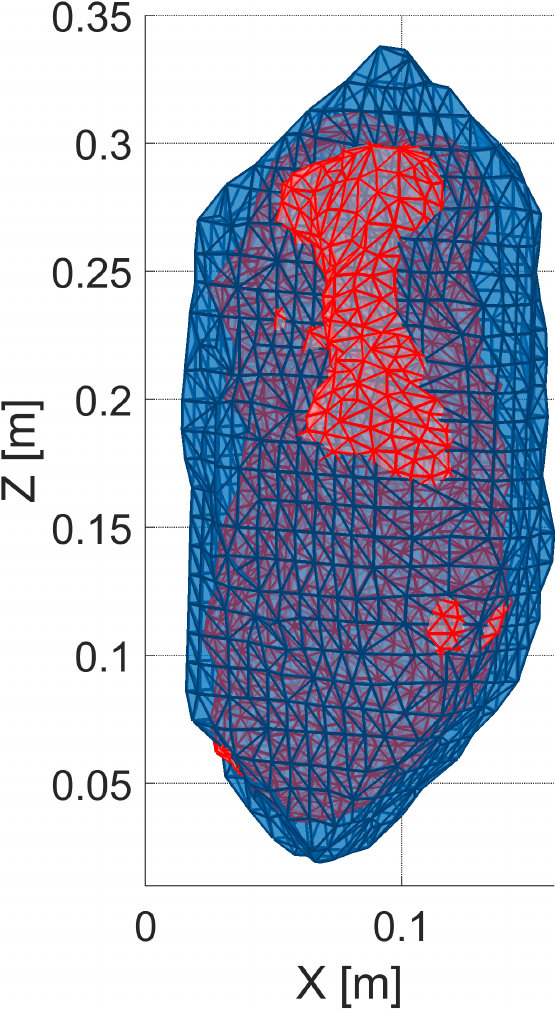} & 
        \hskip -.1in      \includegraphics[height=1.47in]  {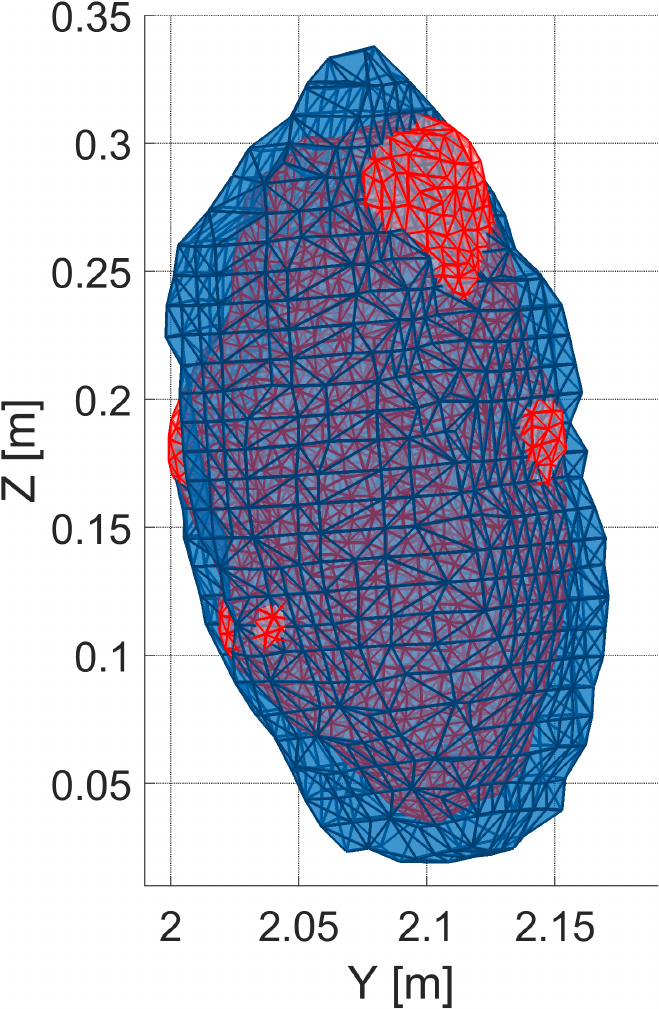} \\ 
         \raisebox{12ex} {\hskip -.2in(c)} &
        \hskip -.2in     \includegraphics[height=1.857in]  {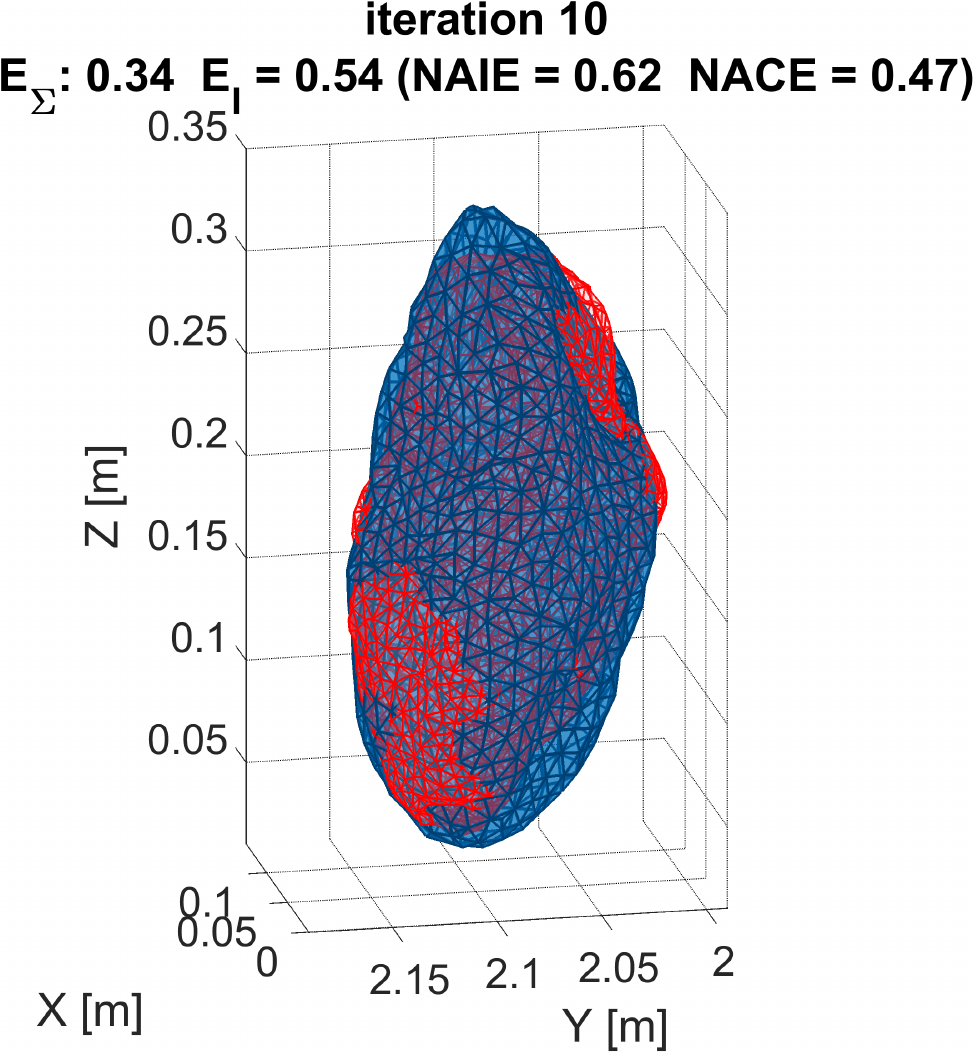} &
        \hskip -.3in      \includegraphics[height=1.47in]  {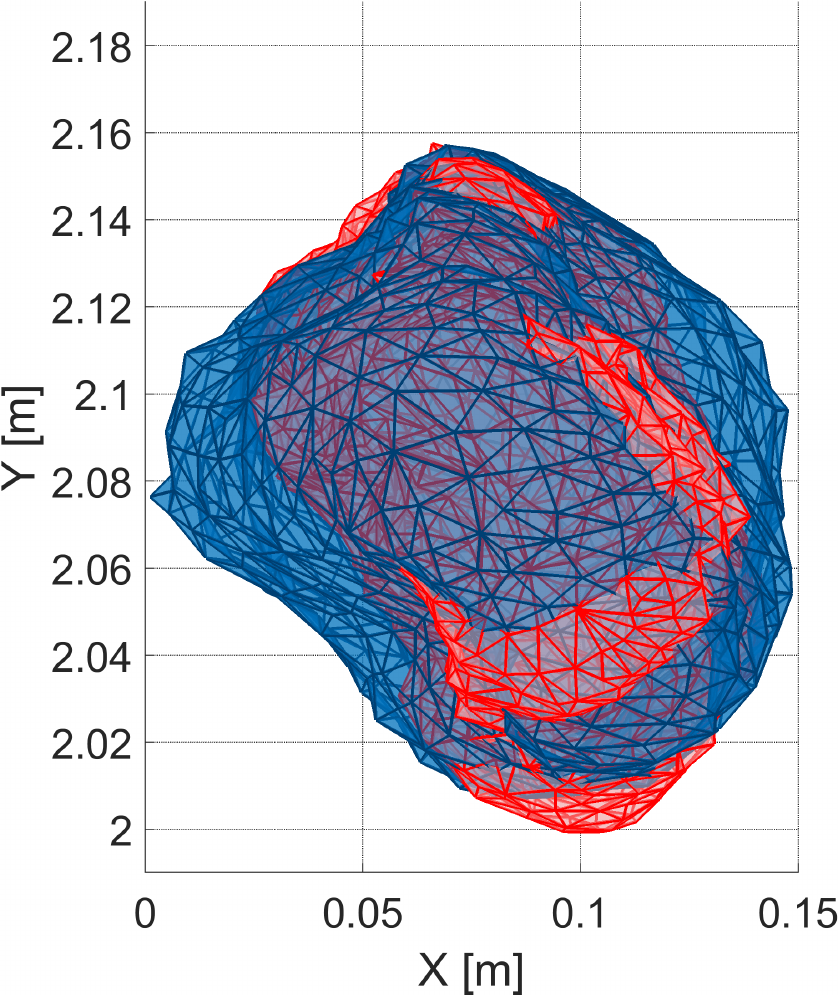} & 
        \hskip -.1in      \includegraphics[height=1.47in]  {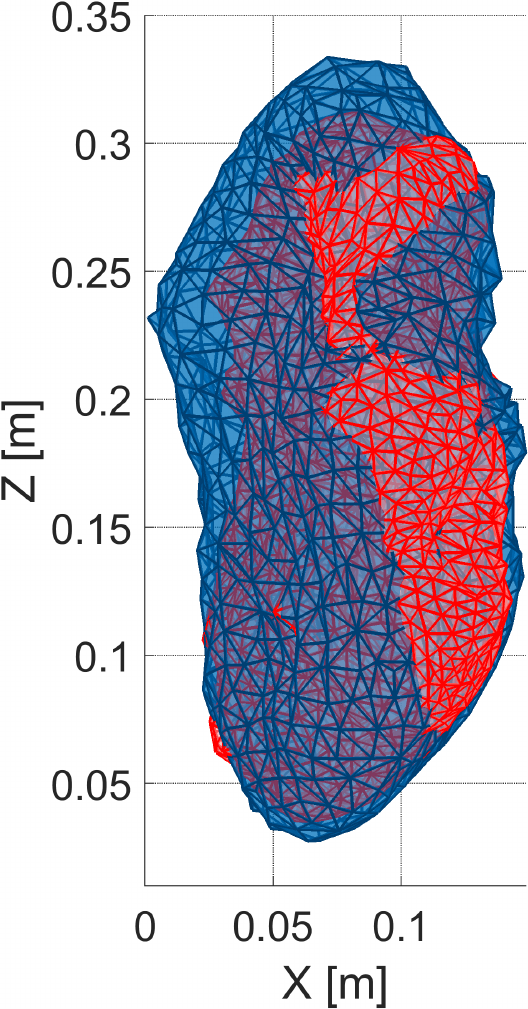} & 
        \hskip -.1in     \includegraphics[height=1.47in]  {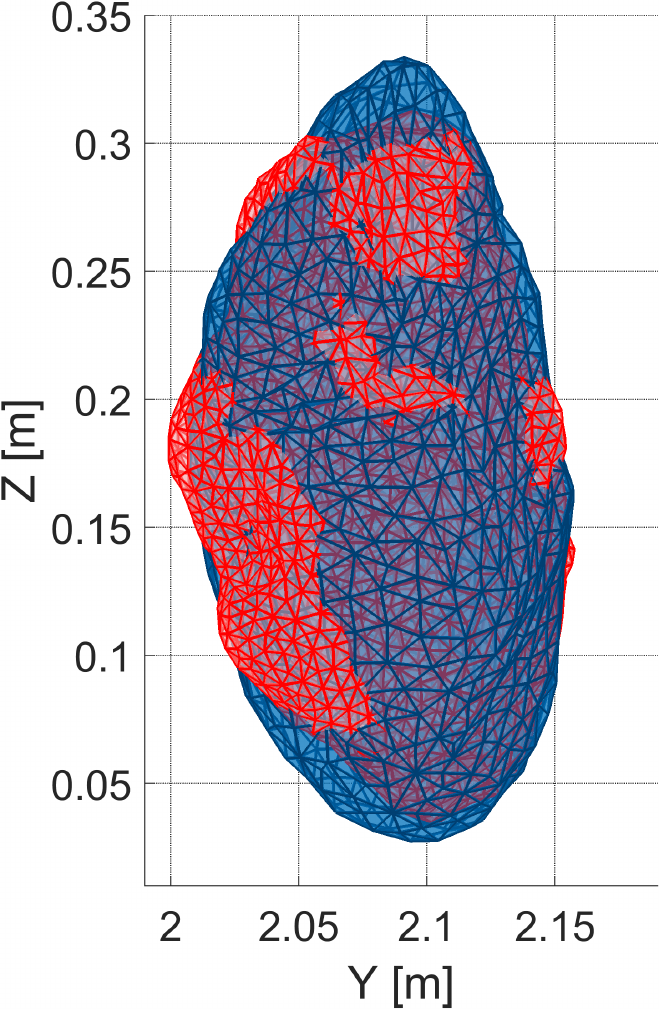} \\ 
      \end{tabular} }
 &
     {\begin{tabular}{c}
     \hskip -.2in
        \includegraphics[height=1.55in]  {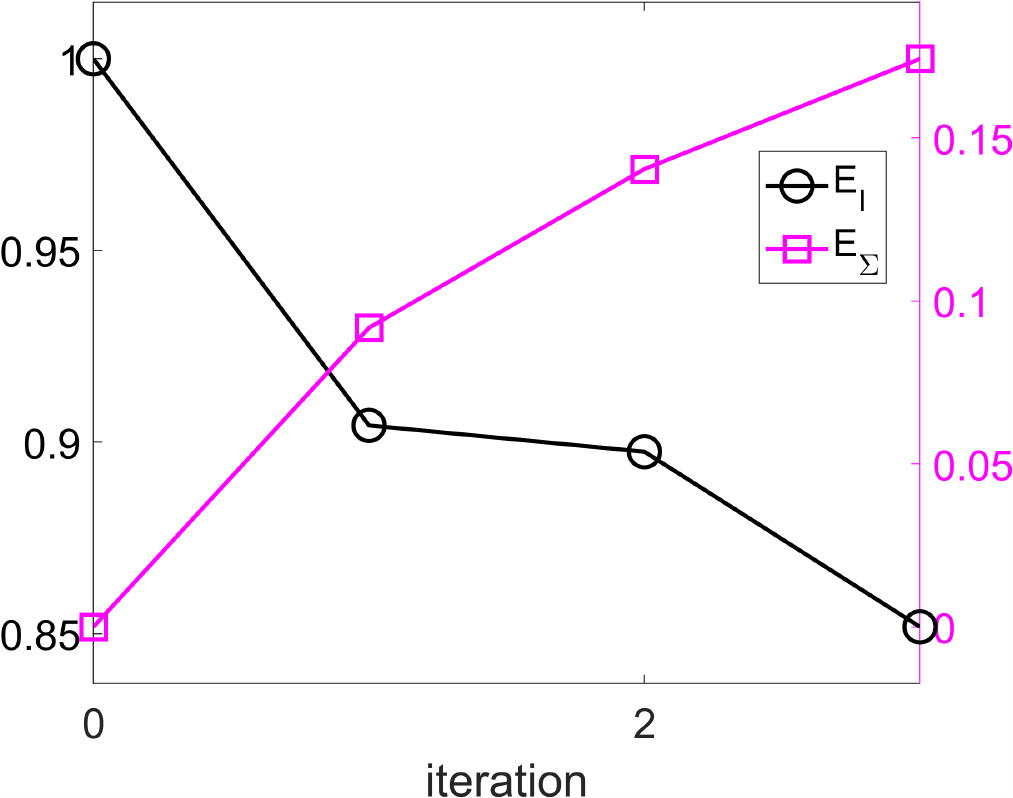} \\
               \raisebox{0ex}{\hskip -.15in(d)}  \\
     \hskip -.2in
        \includegraphics[height=1.55in]  {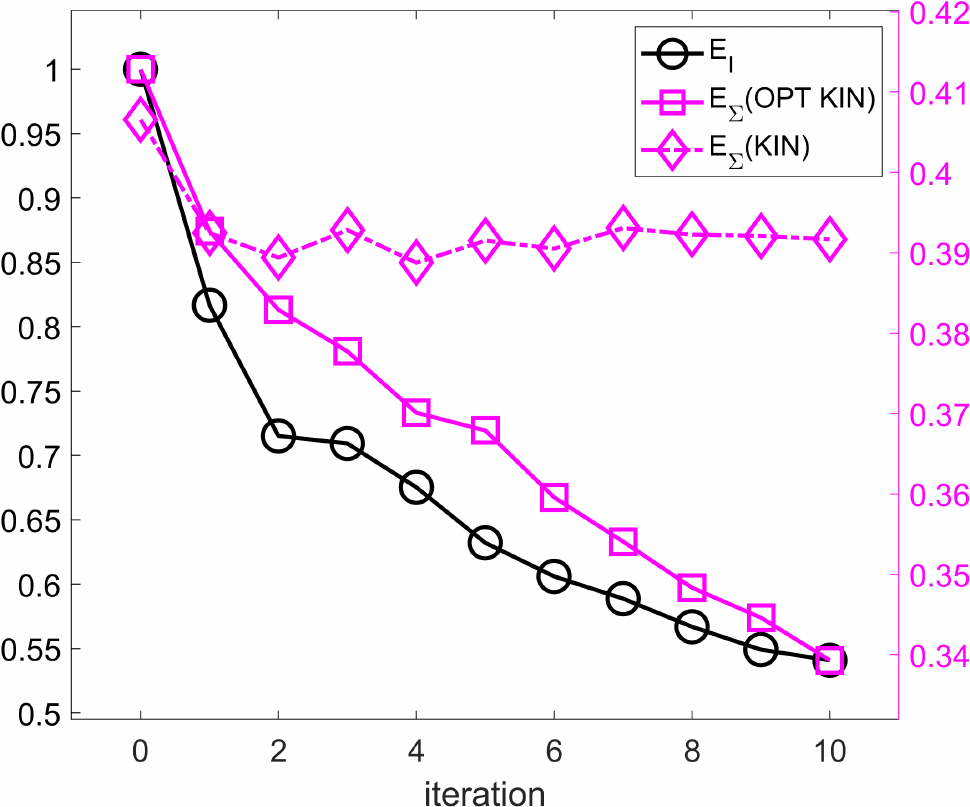} \\
               \raisebox{0ex}{\hskip -.15in(e)}  \\
      \end{tabular}} 
      \\     
 \end{tabular}      
      \caption{Same results as in Fig.~\ref{fig:Coral1_SC_rd} with ghost-corrupted region removed, for rock-two.}
      \label{fig:Coral2_SC_rd}
      \vskip -.2in
\end{figure*}

\subsection{Real Data}
Fig. 8a depicts sample 3-D views and projections onto $XY$, $XZ$ and $YZ$ planes of the original (black) and optimized (red) Kinect models, $S^K$ and $S^{\tilde{K}}$, respectively. For the most part, most local adjustments are relatively small. To quantify, we have given in (e) both image error $E_I$ (black) and normalized volumetric errors (magenta). The initial NVE of zero reflects the initialization by the Kinect model $S^K$.  The image error decreases by about 20\% for the optimized Kinect model $S^{\tilde{K}}$, corresponding to a deviation of 15\% in volumetric error.

Figs.~8b and 8c depict the initial and optimized SC solutions (blue) superimposed on the optimized Kinect model (red). After optimization, the improvement in various local regions can be confirmed. Also, the consistency between the improvement in image error $E_I$ (black) and NVE $E_{\Sigma}$ (magenta square) is noted in (f). {There is less agreement when the NVE of the optimized SC is determined relative to the original Kinect model (magenta diamond), decreasing insignificantly from 0.25 to 0.24.}  In (d), the optimized SC model  (green), without excluding the ghost-corrupted object regions from the sonar views, is less compatible with the optimized Kinect solution. Quantitatively, this is verified by the NVE $E_{\Sigma}$ (green circle), given in (g).

Similar conclusions are drawn from the results of an experiment for a larger coral rock with more complex shape; see Fig. 9. For example, we note a similar outcome from applying the optimization to the Kinect model  in (a,d). Here, the NVE variation of about 17\% due to small distortions in various local regions of the original Kinect model leads to 15\% improvement in image error $E_I$. 
 Next, the optimized SC becomes more consistent with the Kinect models in most regions; see (c). 
Yet, some discrepancies exist in the SW-W and NE-E regions, e.g., as noted in the $XY$ projection. This is dominantly due to lack of measurements from the E and W directions, to unmask the complex variations in the coral shape  [28]. 
Referring to (e), the NVE decreases negligibly again (from 0.41 to 0.39) when normalizing the optimized SC solution with the original Kinect model (magenta diamond); in contrast to much higher consistency through normalization with the optimized Kinect model (magenta square),

\begin{figure*}[th!]
  \centering
      {\begin{tabular}{cccccccc}
        \centering
         \hskip .05in
        \includegraphics[height=1.38in]  {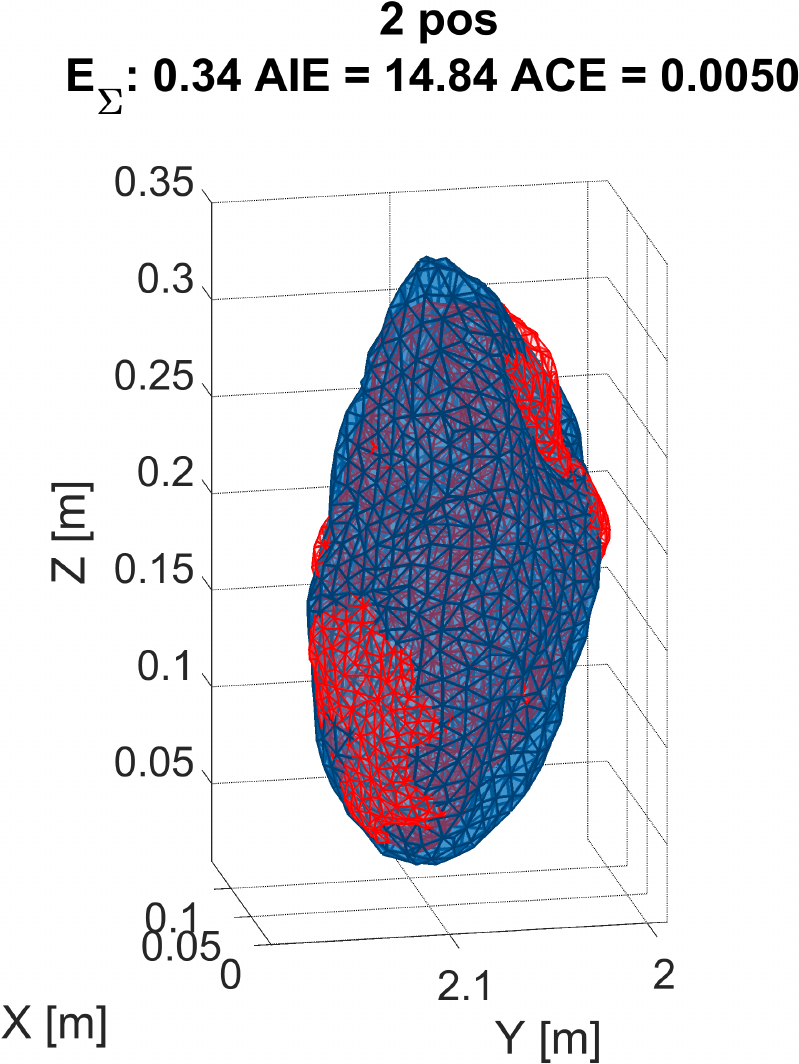} &
         \hskip -.2in
        \includegraphics[height=1.28in]  {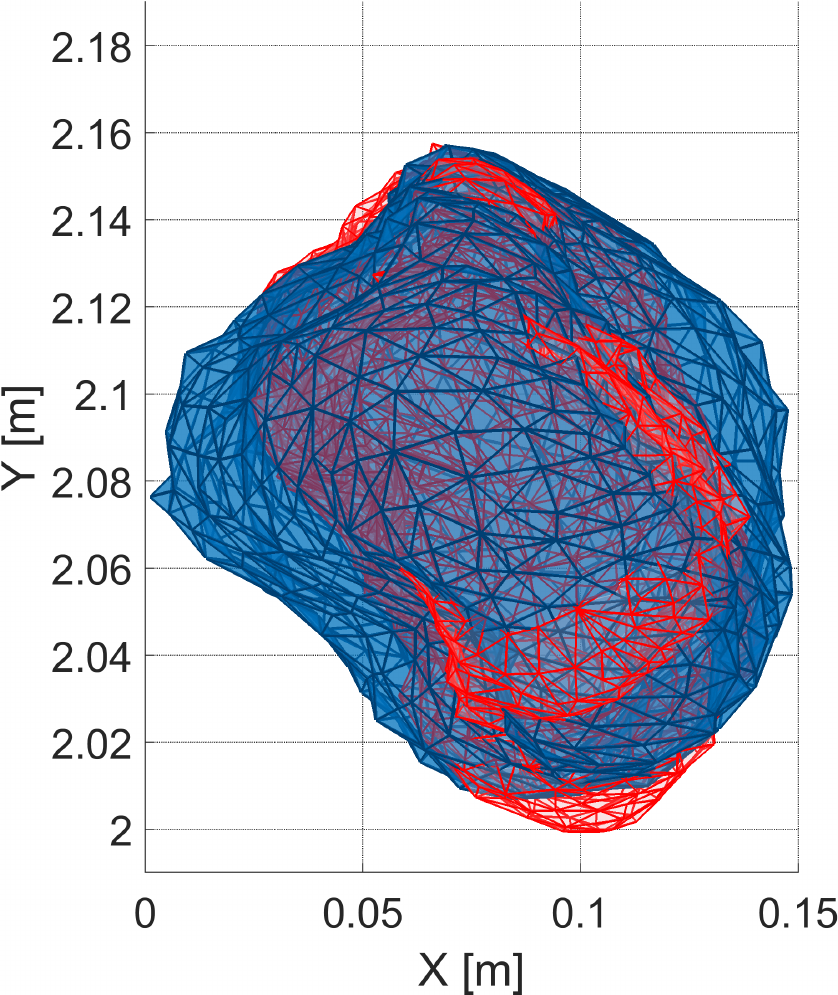} &
         \hskip -.2in
        \includegraphics[height=1.28in]  {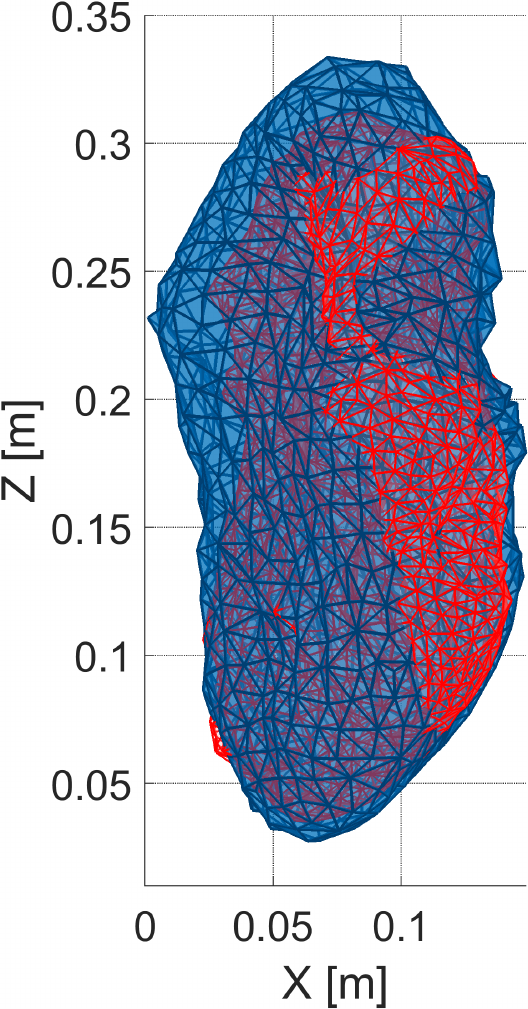} & 
         \hskip -.2in
        \includegraphics[height=1.28in]  {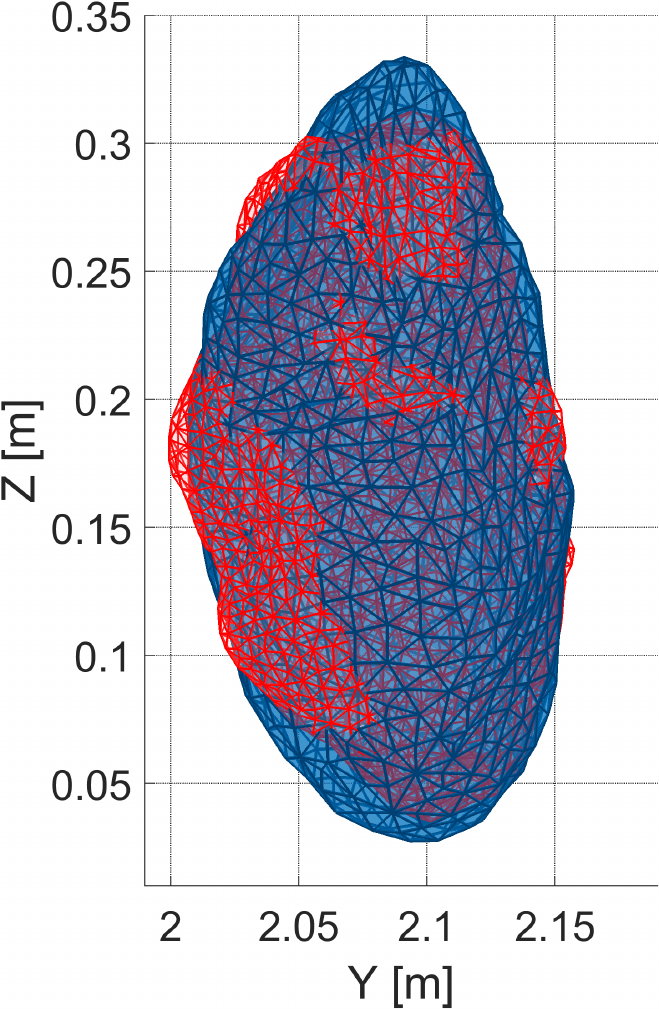} &
         \hskip -.15in
        \includegraphics[height=1.32in]  {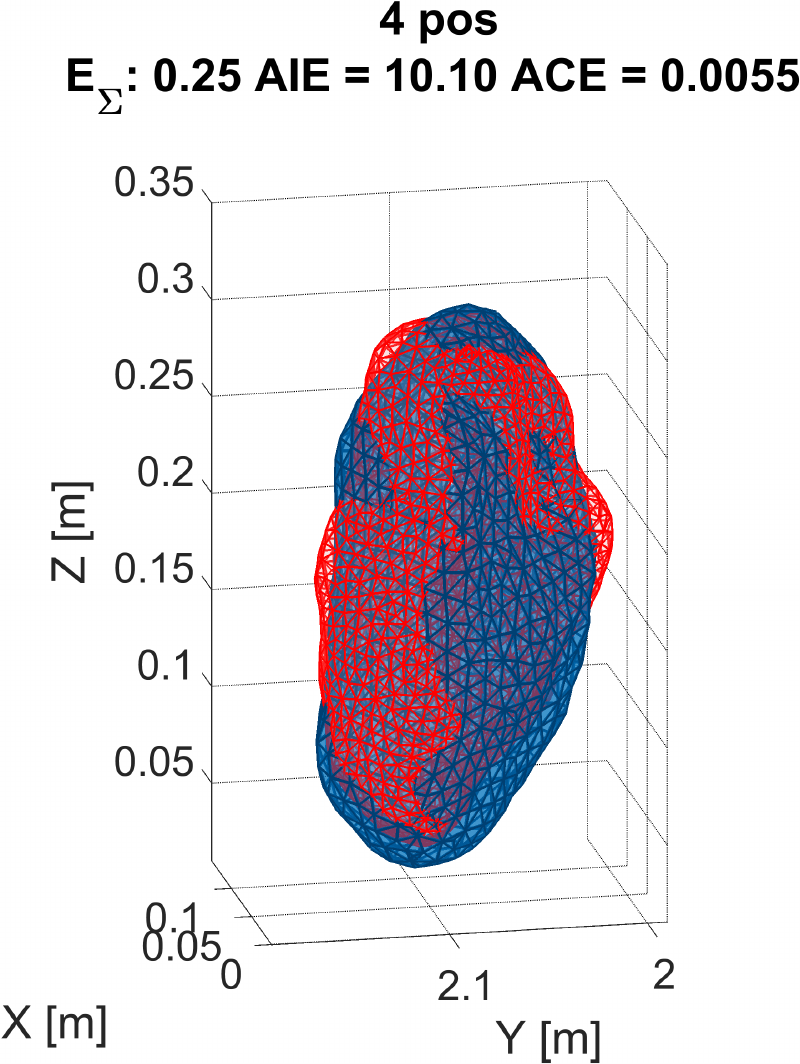} &
         \hskip -.2in
        \includegraphics[height=1.28in]  {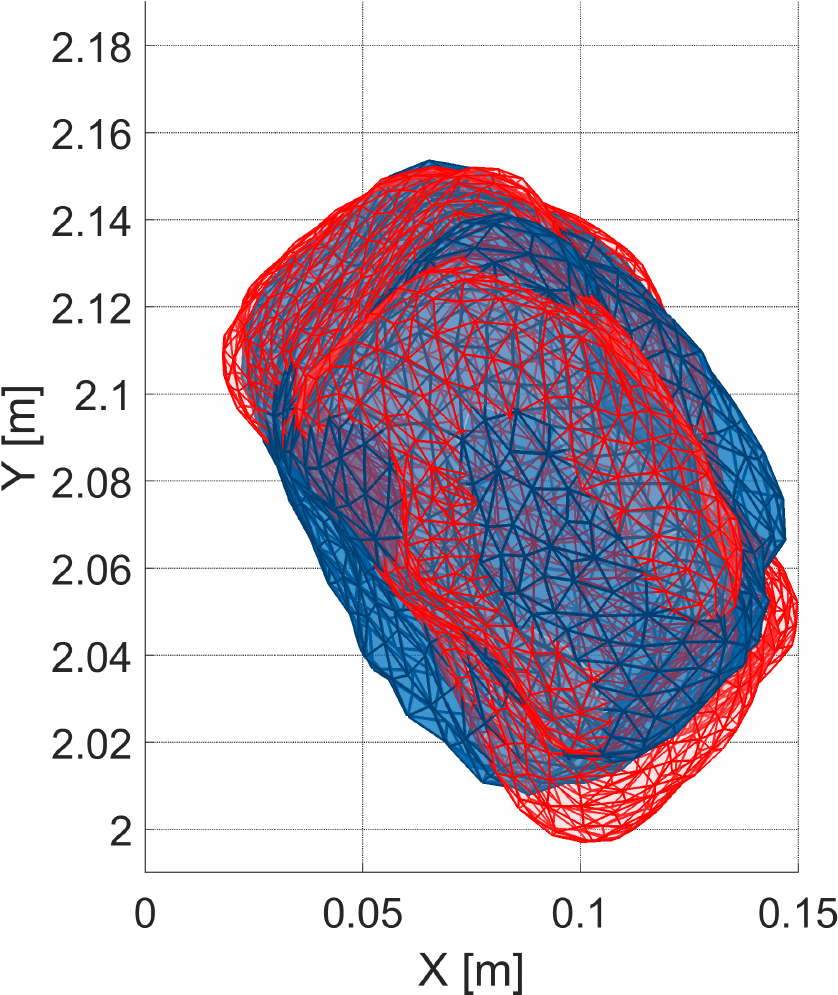} &
         \hskip -.2in
        \includegraphics[height=1.28in]  {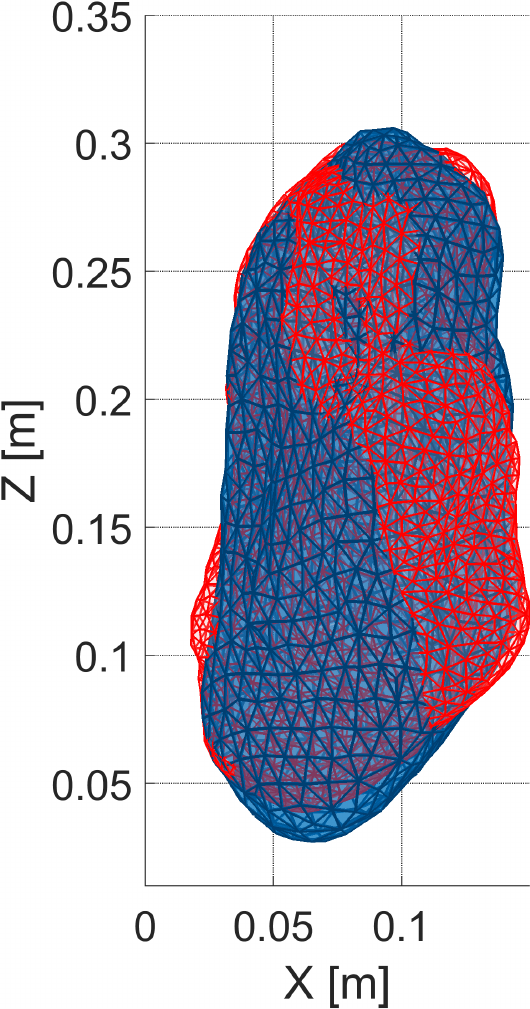} & 
         \hskip -.2in
        \includegraphics[height=1.28in]  {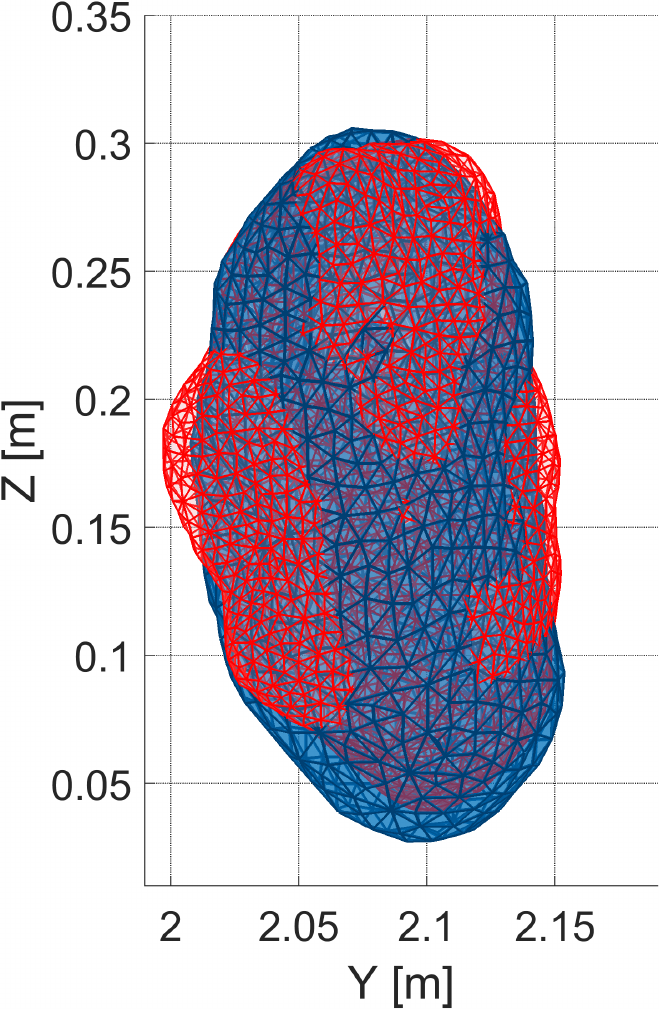} \\ 
        (a) & (b) & (c) & (d) & (a') & (b') & (c') & (d')\\
      \end{tabular}}
      \caption{Real-synthetic mixed data: (a-d) same mesh plots of optimized SC (blue) and optimized Kinect (red) models given in Fig.~\ref{fig:Coral2_SC_rd}c  (for 8 rotated views at each of N  and S sonar positions). (a'-d') improved accuracy of optimized SC model with addition of 8 more synthesized images  at each of E and W positions.}
      \label{fig:Coral2_SC_rd_4view}
\vskip -.1in
\end{figure*} 

\subsection{Real-Synthetic Mixed Data}
Achieving uniform reconstruction accuracy for objects with varying shape complexity in local regions requires data from key poses. With no knowledge of target shape, we may generally require a sonar position within each quadrant. For the second (elongated) coral reef, thickness variations are significant in the Y-direction, which are visible primarily in the missing views from the E and W directions. These become rather important to accurately reconstruct the shape variations. To assess the impact of these ''key'' views, we have utilized the optimized Kinect model 
to generate 8 rotated images at each of E and W sonar position. In Fig.~\ref{fig:Coral2_SC_rd_4view}, we compare the 2 reconstructions using real N-S data and N-S-E-W mixed real-synthetic data. The improvement is noted in various views, quantified by 26\% decrease in $E_{\Sigma}$ (from 0.34 to 0.25) and 32\% in AIE (from 14.84 to 10.10). The ACE remains at 0.005.

\subsection{Non-Flat Air-Water Interface}
For non-flat air-water interface, our results are expected to still apply where the reflection points on the water surface (i.e., $\mmp_W$ in Fig.~3a) for all target echos) lie on a relatively flat local region. However, adjustment may be necessary where the local patch 
varies due to water-surface dynamics, requiring an estimate of depth and sonar orientation relative to the water surface at each image acquisition time. Referring to Fig.~3a,  generalization of $\{d,\ \beta\}$ with $\{d_m,\beta_m=[\beta_x,\beta_y,\beta_z]_m\}$ allows for varying depth and arbitrary orientation relative to the {\it local planar patch} at each image, which may be estimated with the optimization scheme in \cite{thesis}, applied to each image.

Next, we require an independent investigation to extensively assess the impact of mirror and ghost image distortions, induced by arbitrary fluctuations of air-sea interface in variety of sea states, on the reconstruction accuracy. Here, we seek some initial insight into the reconstruction 
performance using ground-truth computer-generated data, by applying some general findings from an experiment with real data.

\begin{figure*}[th!]
  \centering
  \vskip -.65in
      \begin{tabular}{c}
        \includegraphics[width=5.8in]  {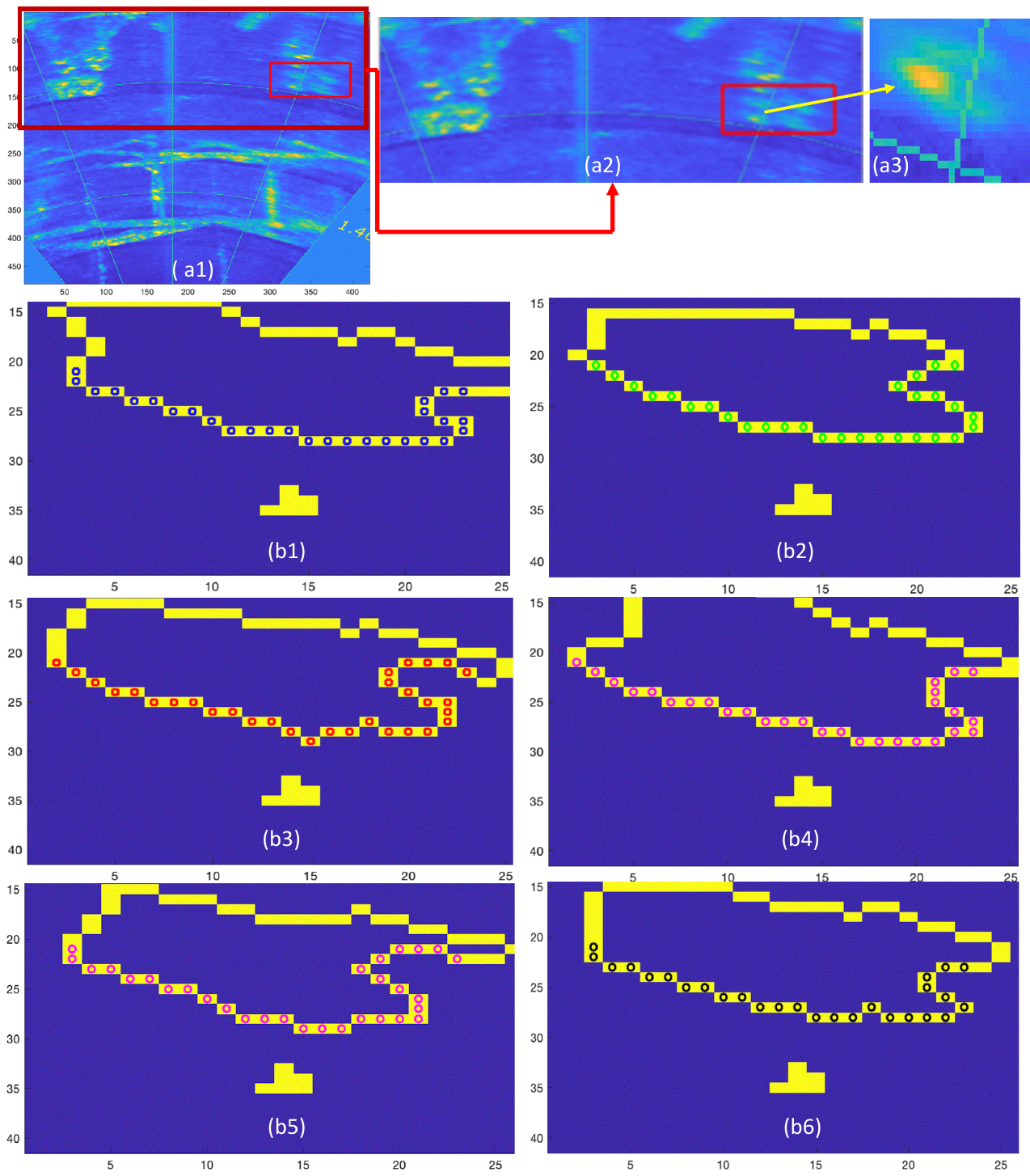}\\
         \end{tabular}
         \vskip -1.in
      \begin{tabular}{c}
         \includegraphics[width=5.8in]  {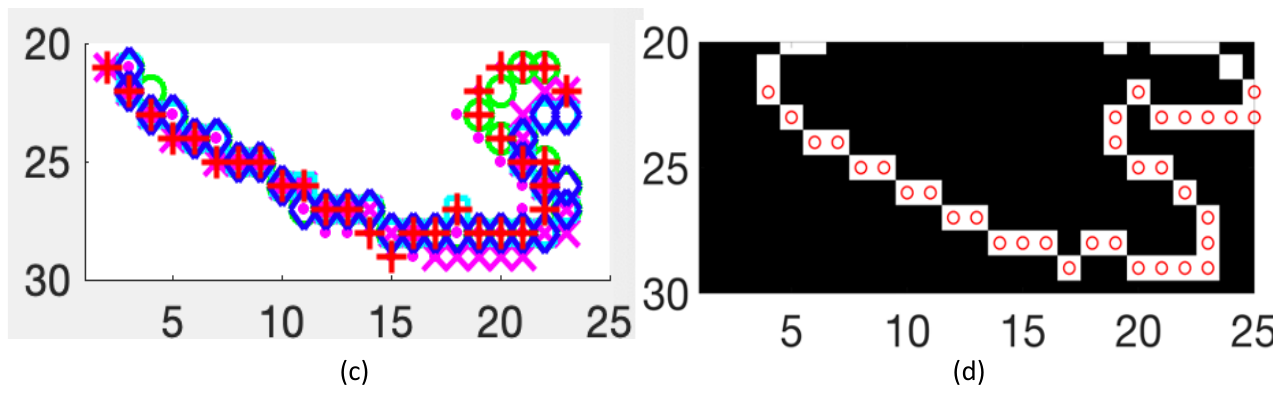}\\
         \end{tabular}
           \vskip -5.4in
         \caption{(a1-a2): Image of a rock on the bottom as main target of pool scene contains multiple mirror components from hard-bottom and surface mutipaths. (a3) Magnification of color-coded left mirror component (within red rectangle) induced by reflection of first surface echos from water surface. (b1-b6) Six sample lower contours, detected by (Canny) edge detector, highlight noticeable displacement with minor distortion of mirror image, due to perturbations in air-water interface by surface waves. (c) Lower contours from 6 samples are superimposed, and compared with (d) contour of the average image over the sequence (about 20 seconds). \label{surfwave}}

\end{figure*} 

\begin{figure*}[t!]
  \centering
    \begin{tabular}{ccc}
         \raisebox{0ex}{\hskip -.5in
         \includegraphics[height=1.75in]  {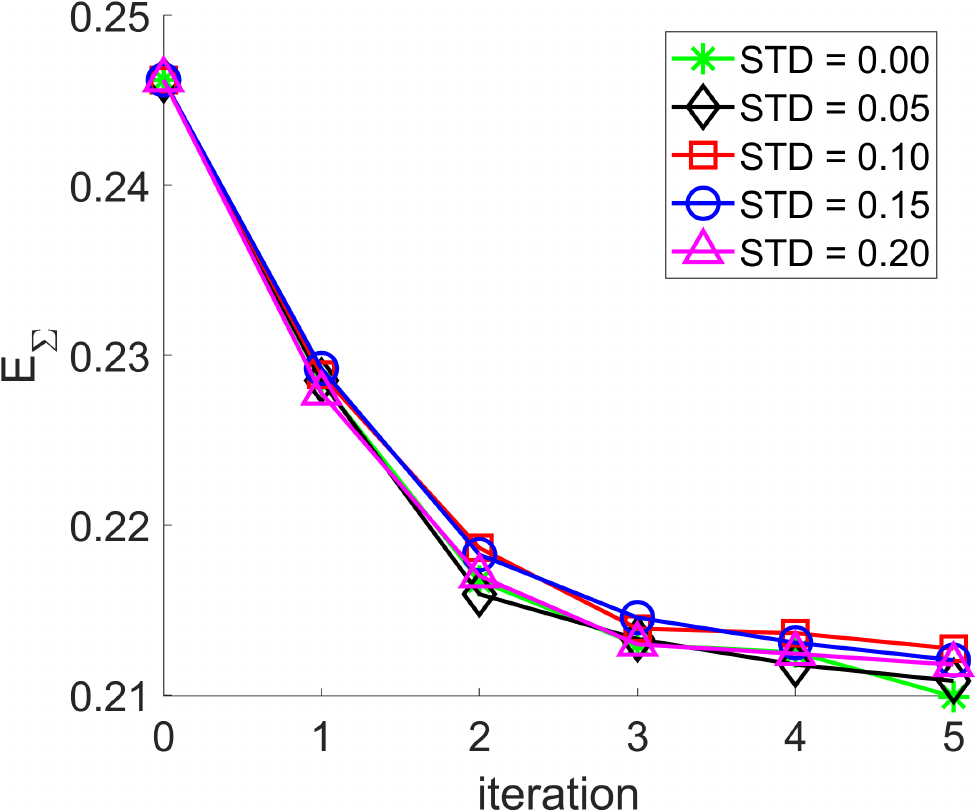} } & 
        \raisebox{0ex}{\hskip 0in
         \includegraphics[height=1.75in]  {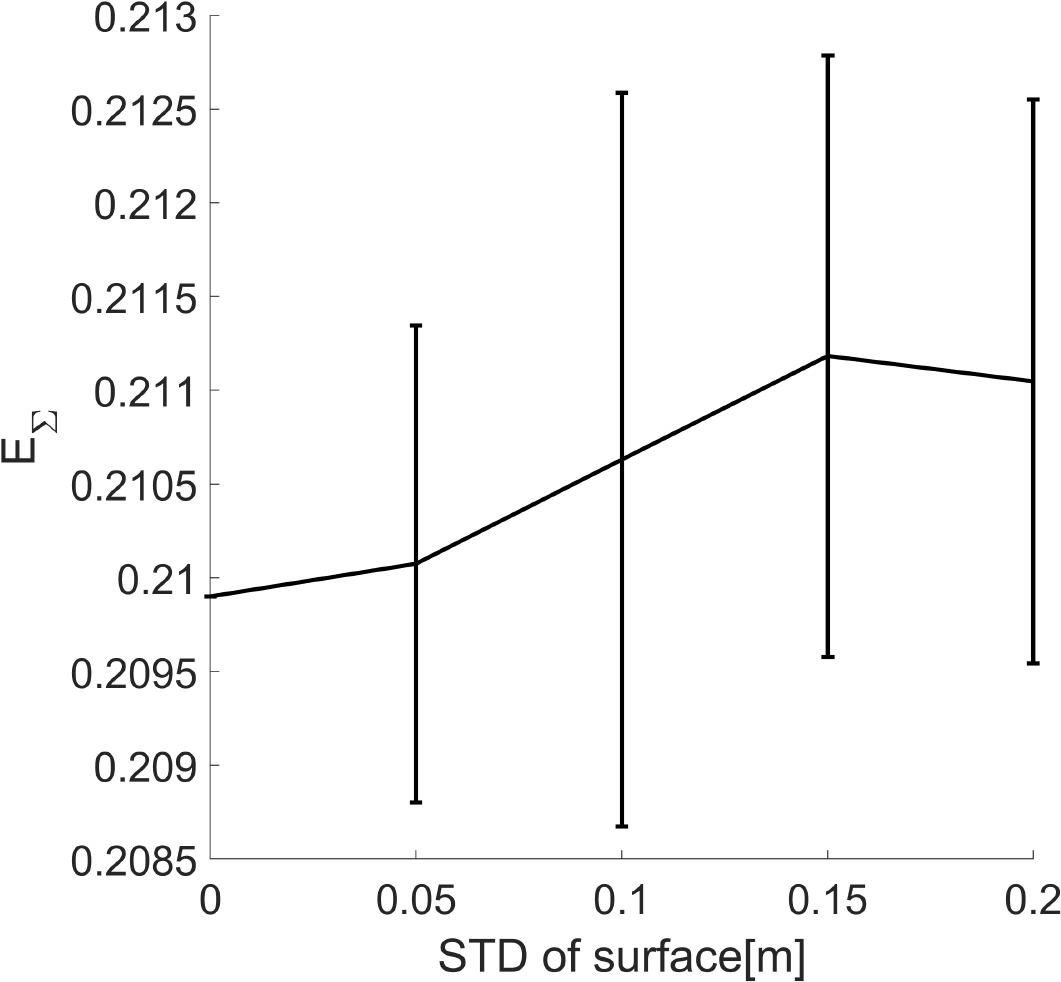}}  & 
         \raisebox{0ex}{\hskip 0.in
        \includegraphics[height=1.75in,width=2.35in]   {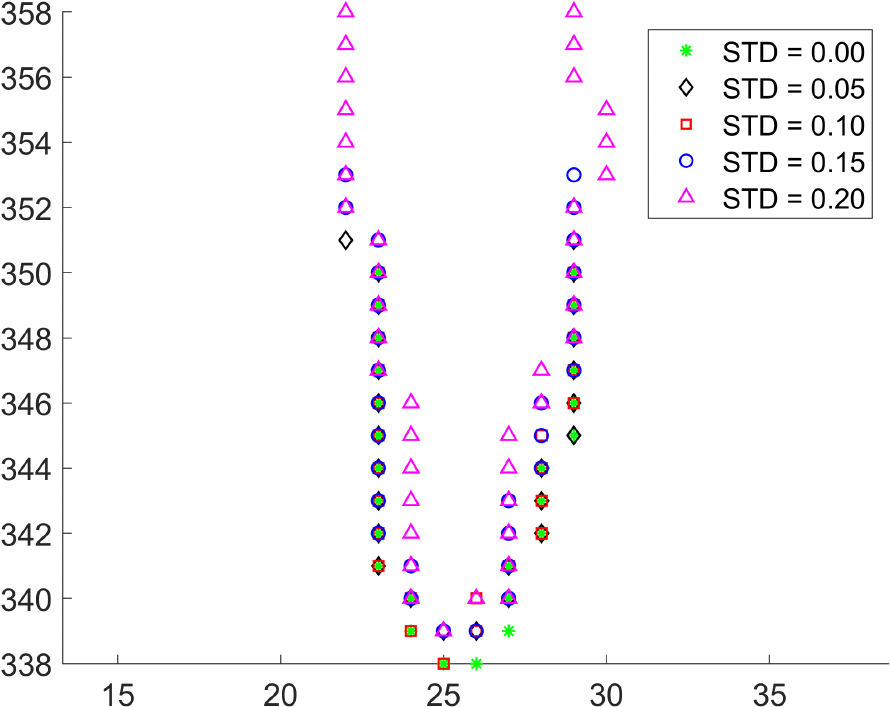}} \\
        \hskip -.5in (a) & (b) & (c)\\
 \end{tabular}  
\caption{(a) Normalized volumetric error -vs- iterative number for various fluctuations levels of air-water interface; (b) average and one standard deviation of normalized volumetric error in 30 simulations for various interface fluctuations levels; (c) sample distortion in (lower 30\% of) mirror contour for various fluctuation levels.}
      \label{nonflat}
\end{figure*}

The main target is a rock on the bottom of the water-tank scene of Fig.~11a1 in water depth of 2.5 ft; with 6-8 inches in various dimensions and similar shape to coral one.  Fig.~\ref{surfwave}a depicts a sample pseudo-colored image from an Oculus sonar in the same 90\degree\,rotated configuration. Water surface waves with magnitude of a few inches are introduced to disturb the mirror and ghost components (not visualized in a static image, the latter can be detected when viewing the video). Among several mirror images induced by the hard-bottom and surface multi-path, we track the left one within the red rectangle (see Fig.~\ref{surfwave}a2), formed by the water-surface reflection of target's first echos\footnote{Adjacent mirror objects arise by multipaths due to the delayed target echos from the first bottom refections, reaching the water surface and vice-versa, as well as bounces between bottom and water surface.}; see Fig.~3a.  The detected lower part of mirror contours from 6 sample frames at regular intervals are given in (b1) to b(6). These contours are superimposed in (c) to visualize their variations, also compared with the detected lower contour from the average mirror image over a 10-second duration. In these results, the time-varying surface waves primarily displace the mirror image, with shape distortion being less significant. Using these general observations, an experiment with computer-generated data is performed to quantitatively assess the 3-D reconstruction accuracy based on ground truth.
Here, we employ the 3-D Kinect model of coral-one object to generate synthetic images, with induced random water surface variations drawn from a Gaussian distribution. Aside from $\sigma=0$ defining a flat surface (used for comparison), the prescribed standard deviations $\sigma=\{0.05,\  0.1,\ 0.15, 0.2 \}$ generate maximum mirror contour variations (over lower 30\% part) at the same scales as in the rock experiment. Fig.~\ref{nonflat}(c) gives the mirror contours for one sample experiment out of 20 (as explained below). 

Starting with the SC solution, the optimizing for each water-surface fluctuation level is carried out by assuming a flat air-water interface. We compare the volumetric errors (normalized with respect to the 3-D optimized Kinect Model) after five optimization iterations for both the flat water surface ($\sigma=0$) and different fluctuation levels. Fig.~\ref{nonflat}(a) depicts one sample plot of the normalized volumetric error -vs- iteration number. Noting negligible impact from the interface fluctuations, the average and standard deviations of these errors, given in Fig.~\ref{nonflat}(b), have been calculated by repeating the experiment only 20 times (for each air-water interface fluctuation level). The negligible volumetric error variations is partly attributed to small deviations in the lower contour, and discarded outliers during the ICP-based contour matching process (in the first step of 3-D model optimization in Fig.~\ref{fig:optimization_overview}). 

\section{Conclusions}
In this work, we have proposed a novel solution to enhance the 3-D reconstruction of target models from a multitude of FSS
images at known poses, by minimizing the discrepancy between the data and synthesized images
generated from the 3-D model, introduced in [1]. Our iterative optimization scheme effectively
avoids the convergence to local minima, the key bottleneck of the proposed gradient-descent scheme in this earlier work. 
More importantly, we have addressed the treatment of ghost and mirror components formed by multipath propagations, when
images are recorded near the sea surface. The localization of ghost image component is critical
to avoid 3-D shape distortion arising from corrupted object regions. In contrast, the mirror image provides
complementary visual cues to regularize and improve the 3-D model reconstruction. 

A key extension is to incorporate the multipath from inter-reflection among (deep) object concavities. While these 
also correspond to longer paths travelled by the acoustic returns, they generally produce an exaggerated (thickened) 
object volume. In our current method, optimizing the 3-D shape model using generated synthetic images poses a 
chicken-and-egg problem: we need the sought after true 3-D model to eliminate the multi-path contributions that yield 
the thickened object views. A potential solution is to thin the initial space-carving solution, using which various 
multipath components are estimated. Here, the goal becomes to identify the optimum “thinned surface model” that again 
yields the least discrepancy between the data and synthesized images. Initial work has shown some merit in the approach,
although more efficient approaches may be devised.  Also, an extensive investigation of the non-flat air-water interface
can lead to a more general solution by incorporating the estimation of some relevant water surface parameters.

T. Guerneve, K. Subr, and Y. Petillot, “Three-dimensional reconstruction of underwater objects using wide-aperture imaging sonar,” Journal of Field Robotics, 2018.

E. Westman, I. Gkioulekas, and M. Kaess, “A volumetric albedo framework for 3d imaging sonar reconstruction,” in 2020 IEEE International Conference on Robotics and Automation (ICRA), pp. 9645–9651, IEEE, 2020.

Y. Wang, Y. Ji, D. Liu, H. Tsuchiya, A. Yamashita, and H. Asama, “Elevation angle estimation in 2d acoustic images using pseudo front view,” IEEE Robotics and Automation Letters, vol. 6(2), pp. 1535–1542, 2021.


\vskip -.75in
\begin{IEEEbiography}
[\raisebox{5ex}{\includegraphics[width=.8in,keepaspectratio] {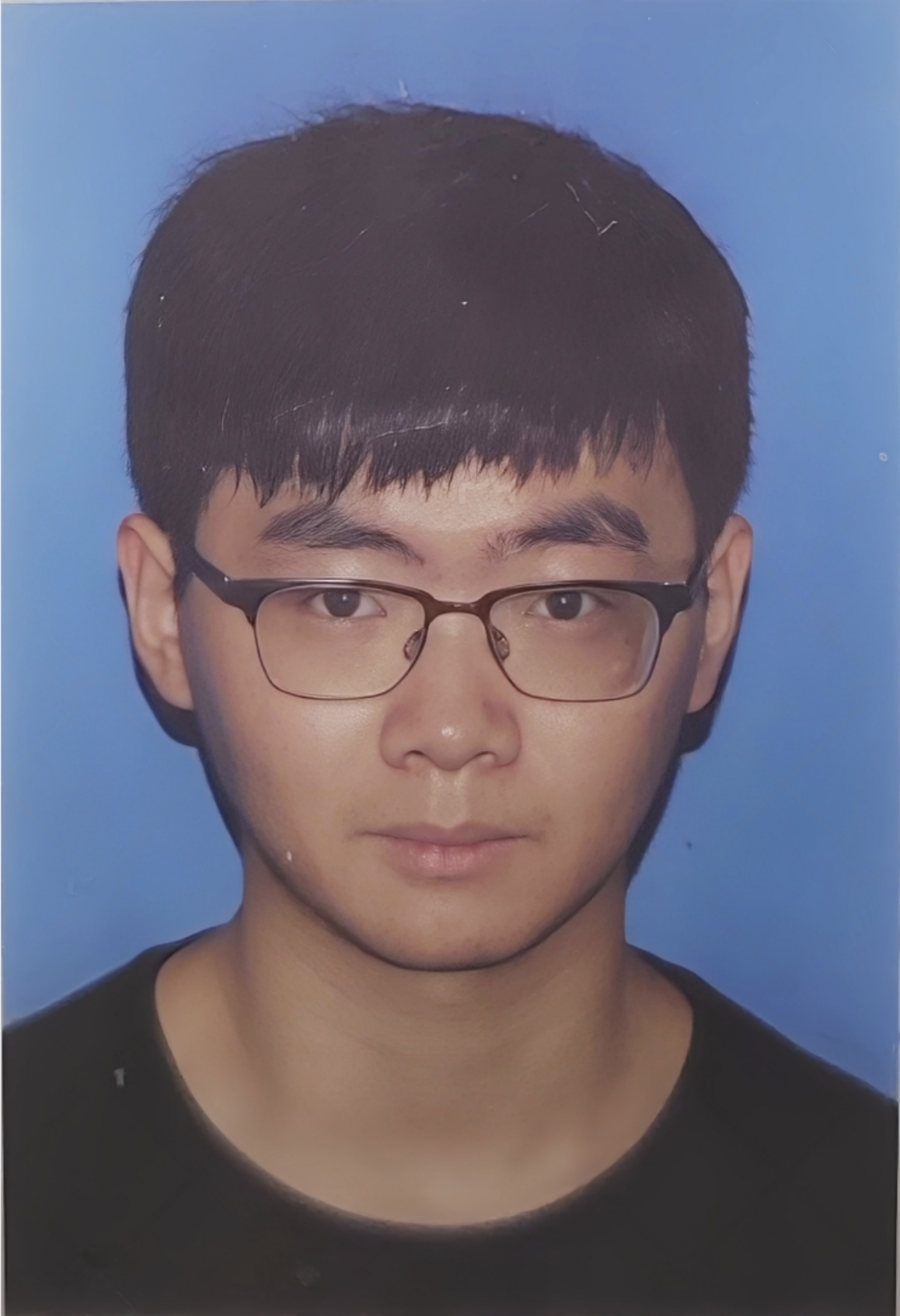}}]
{Yuhan Liu}  completed the B.S. degree in Electronic and Information Engineering from Beihang University (August 2019) and his M.S. degree in Electrical and Computer Engineering from the University of Miami (August 2022). His areas of research interest include image processing, computer vision, and machine learning. He is currently working for a high-tech company in Beijing, China, where he contributes to the technological advancement in these fields. 

\end{IEEEbiography}
\vskip -1.25in
\begin{IEEEbiography}
[\raisebox{-7ex}{\hskip 0in\includegraphics[width=0.8in,keepaspectratio] {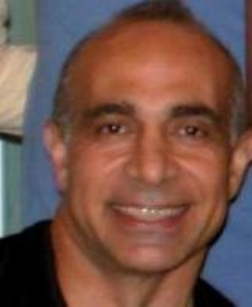}}]
{Shahriar Negahdaripour} received S.B., S.M., and Ph.D. degrees in Mechanical Engineering from MIT, Cambridge, MA (1979, 1980, and 1987). He received  the NSF Engineering Initiation Award (1989),  ``DoD SERDP Project of the Year (2009) for {\it Application of ROV-based video technology to complement coral reef resource mapping and monitoring}, Siemens Corp's Best Paper Award (2003) at {\it IEEE AVSS'03 Conference}, and Best Paper Award (2007 ) at {\it IEEE BMG'07 Workshop}, held in conj. with {\it CVPR'07}. He served as a General Co-Chair of {\it IEEE CVPR'91} and {\it IEEE ISCV'95}. He is an IEEE Fellow of OES. His research interests include underwater computer vision, and application to sonar imaging and opti-acoustic stereo.\\
\end{IEEEbiography}

\end{document}